\documentclass{article}

\usepackage{microtype}
\usepackage{graphicx}
\usepackage{subfigure}
\usepackage{booktabs} 

\usepackage{hyperref}

\usepackage[accepted]{icml2023}

\usepackage{amsmath}
\usepackage{amssymb}
\usepackage{mathtools}
\usepackage{amsthm}
\usepackage{paralist}

\usepackage[capitalize,noabbrev]{cleveref}

\theoremstyle{plain}

\theoremstyle{definition}

\theoremstyle{remark}

\usepackage{algorithmic}
\usepackage[algo2e]{algorithm2e}
\usepackage[utf8]{inputenc}
\usepackage[cjk]{kotex}
\usepackage{subfigure}
\usepackage{multirow}
\usepackage{pbox}
\usepackage{booktabs}
\usepackage{graphicx}
\usepackage{bm}
\usepackage{siunitx}
\usepackage[toc,page,header]{appendix}

\usepackage{minitoc}

\newcommand{\std}{\scriptsize{}}

\newcommand{\BEST}[1]{\textcolor{red}{#1}}
\newcommand{\SECOND}[1]{\textcolor{blue}{#1}}
\newcommand{\THIRD}[1]{\textcolor{violet}{#1}}

\icmltitlerunning{GREAD: Graph Neural Reaction-Diffusion Networks}

\begin{document}

\twocolumn[
\icmltitle{GREAD: Graph Neural Reaction-Diffusion Networks}



\icmlsetsymbol{equal}{*}

\begin{icmlauthorlist}
\icmlauthor{Jeongwhan Choi}{yonsei}
\icmlauthor{Seoyoung Hong}{yonsei}
\icmlauthor{Noseong Park}{yonsei}
\icmlauthor{Sung-Bae Cho}{yonsei}
\end{icmlauthorlist}

\icmlaffiliation{yonsei}{Yonsei University, Seoul, South Korea}

\icmlcorrespondingauthor{Noseong Park}{noseong@yonsei.ac.kr}

\icmlkeywords{Graph neural network, Reaction-diffusion equation}

\vskip 0.3in
]



\printAffiliationsAndNotice{}  

\begin{abstract}
Graph neural networks (GNNs) are one of the most popular research topics for deep learning. GNN methods typically have been designed on top of the graph signal processing theory. In particular, diffusion equations have been widely used for designing the core processing layer of GNNs, and therefore they are inevitably vulnerable to the notorious oversmoothing problem. Recently, a couple of papers paid attention to reaction equations in conjunctions with diffusion equations. However, they all consider limited forms of reaction equations. To this end, we present a reaction-diffusion equation-based GNN method that considers all popular types of reaction equations in addition to one special reaction equation designed by us. To our knowledge, our paper is one of the most comprehensive studies on reaction-diffusion equation-based GNNs. In our experiments with 9 datasets and 28 baselines, our method, called GREAD, outperforms them in a majority of cases. Further synthetic data experiments show that it mitigates the oversmoothing problem and works well for various homophily rates.
\end{abstract}
\section{Introduction}
Graphs are a useful data format that occurs frequently in real-world applications, e.g., computer vision and graphics~\cite{monti2017MoNet}, molecular chemistry inference~\cite{Gilmer2017chemi}, recommender systems~\cite{Rex2018pinsage,choi2023bspm},   drug discovery~\cite{gaudelet2021utilizing}, traffic forecasting~\cite{choi2022STGNCDE}, and so forth. With the rise of graph-based data, graph neural networks (GNNs) are attracting much attention these days. However, there have been fierce debates on the neural network architecture of GNNs~\cite{kipf2017GCN,velickovic2018GAT,defferrard2016chebnet,Wu2019SGC,chen2020gcnii,chien2021GPRGNN,zhu2020h2gcn,yan2021GGCN,lim2022LINKX,li2022GloGNN,luan2022ACMGCN,rusch2022gcon,chamberlain2021grand,chamberlain2021blend,bodnar2022Sheaf}.

\begin{table}[t]
\small
    \centering
    \setlength{\tabcolsep}{3pt}
    \caption{A comparison table of existing methods. `$\triangle$' means that it corresponds to a specific type of reaction only.}
    \begin{tabular}{cccc}\toprule
        Model     & Diffusion & Reaction    & Continuous-time\\\midrule
        FA-GCN    & O         & $\triangle$ & X \\
        GPR-GNN   & O         & $\triangle$ & X \\
        ACM-GCN   & O         & $\triangle$ & X \\
        CGNN      & O         & X           & O \\
        GRAND     & O         & X           & O \\
        BLEND     & O         & X           & O \\\midrule
        GREAD     & O         & O           & O \\
        \bottomrule
    \end{tabular}
    \label{tab:comp}
\end{table}

\begin{figure}[!t]
    \centering
    \subfigure[$t=0$]{\includegraphics[width=0.24\columnwidth]{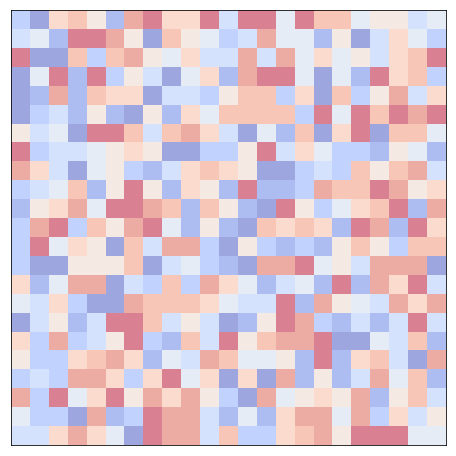}}
    \subfigure[$t=5$]{\includegraphics[width=0.24\columnwidth]{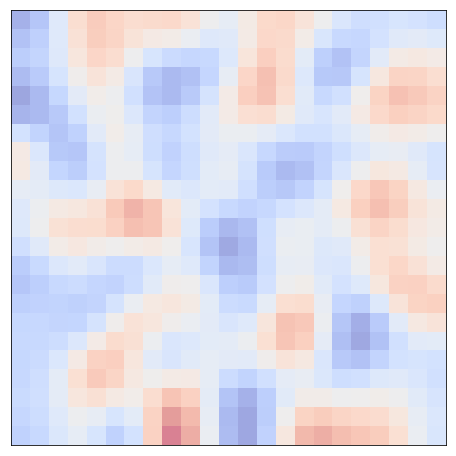}}
    \subfigure[$t=25$]{\includegraphics[width=0.24\columnwidth]{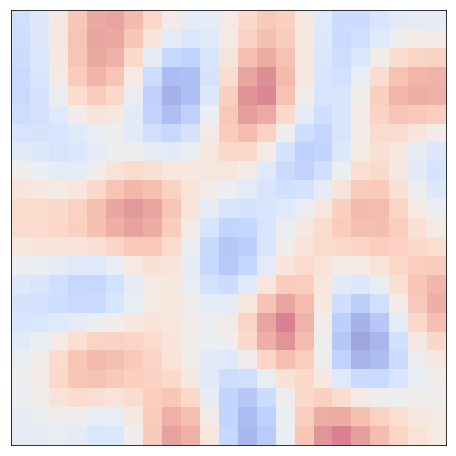}}
    \subfigure[$t=50$]{\includegraphics[width=0.24\columnwidth]{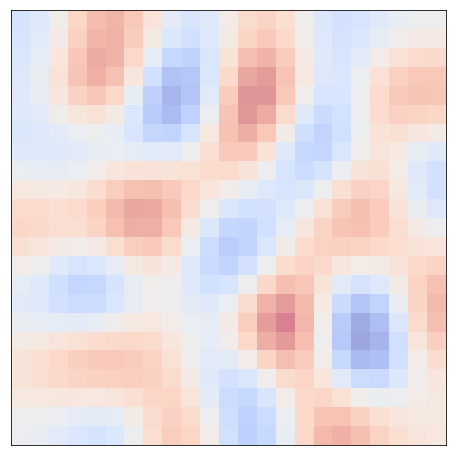}}
    \subfigure[$t=0$]{\includegraphics[width=0.24\columnwidth]{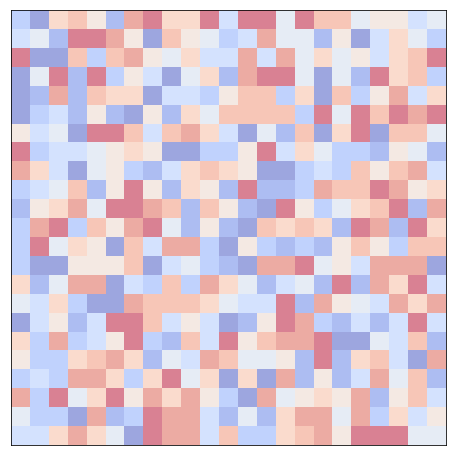}}
    \subfigure[$t=5$]{\includegraphics[width=0.24\columnwidth]{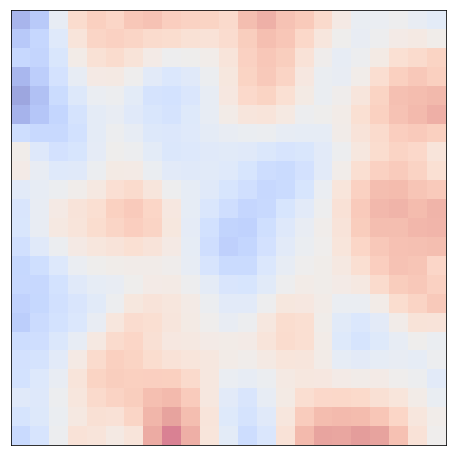}}
    \subfigure[$t=25$]{\includegraphics[width=0.24\columnwidth]{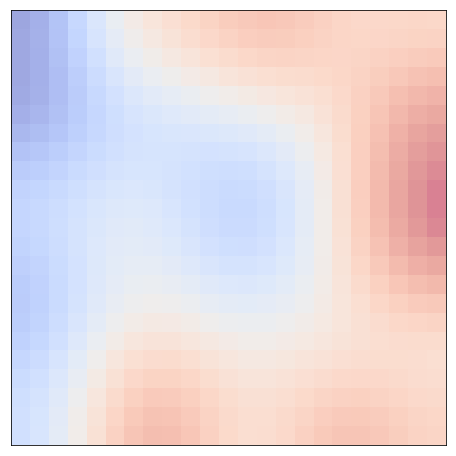}}
    \subfigure[$t=50$]{\includegraphics[width=0.24\columnwidth]{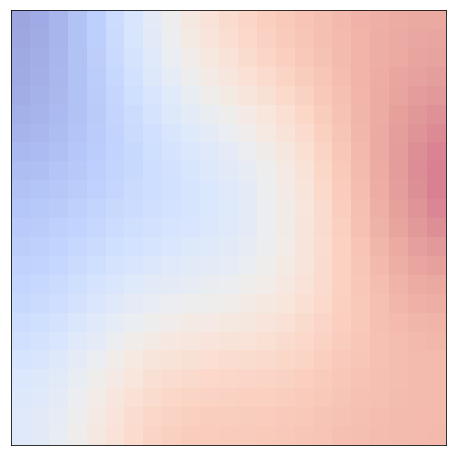}}
    \caption{An illustrative comparison between the diffusion equation in Eq.~\eqref{eq:blur} (bottom) and our proposed blurring-sharpening (reaction-diffusion) equation in Eq.~\eqref{eq:bsl} (top) on a grid graph with one-dimensional node features. The diffusion equation causes the problem of oversmoothing while the reaction-diffusion seeks a balance between smoothing and sharpening.}
    \label{fig:teaser}
\end{figure}

For the past couple of years, many proposed methods have been designed based on the diffusion concept. Many recent GNN methods that rely on low-pass filters fall into this category. Although they have shown non-trivial successes in many tasks, it is still unclear whether it is an optimal direction of designing GNNs.

\begin{table}[t]
\small
    \centering
    \setlength{\tabcolsep}{3pt}
    \caption{The average ranking/accuracy and the Olympic ranking of some selected high-performing models on 9 real-world datasets. `$\ast$' (resp. `$\dagger$') indicates that an improvement over GloGNN (resp. ACM-GCN) is statistically significant ($p<0.05$) under the Wilcoxon signed-rank test.}
    \begin{tabular}{c cc ccc} \toprule
        \multirow{2}{*}{Method} & \multicolumn{2}{c}{Average} 
                                                           & \multicolumn{3}{c}{Olympic Ranking}\\  \cmidrule(lr){2-3}\cmidrule(lr){4-6}
                 & Ranking & Accuracy                      & Gold & Silver & Bronze\\ \midrule
        GREAD-BS & 1.56    & \;\;\;76.64$^{\ast \dagger}$  & 5 & 4 & 0 \\
        GREAD-FB*& 6.72    & \;\;74.51$^{\dagger}$         & 0 & 0 & 3 \\
        GREAD-F  & 7.50    & 74.13                         & 1 & 1 & 1 \\ 
        GloGNN	 & 8.17    & 74.99                         & 0 & 0 & 1 \\
        GREAD-AC & 8.50    & 73.71                         & 1 & 0 & 0 \\
        ACM-GCN	 & 8.67    & 74.92                         & 0 & 1 & 0 \\
        GGCN	 & 9.50    & 75.05                         & 0 & 1 & 0 \\
        Sheaf	 & 10.33   & 75.06                         & 1 & 1 & 0 \\
        \bottomrule
    \end{tabular}
    \label{tab:summary}
\end{table}

In Table~\ref{tab:comp}, we compare recent methods. Most of them rely on diffusion processes, while some of them (i.e., FA-GCN~\cite{Bo2021fagcn}, GPR-GNN~\cite{chien2021GPRGNN}, and ACM-GCN~\cite{luan2022ACMGCN}) partially utilize reaction processes. Those three methods, however, utilize limited forms of the reaction processes. In this regard, there do not exist any methods that fully consider diverse forms of reaction processes --- we consider 7 reaction processes.

To this end, we propose the concept of graph neural reaction-diffusion network (GREAD), which is one of the most generalized architectures since we consider both the diffusion and the reaction processes. Reaction-diffusion equations are physical models that can be used when i) substances are diffused over space and time, and ii) they can sometimes react to each other. Whereas diffusion processes smooth node features on a graph out, reaction-diffusion processes lead to many local clusters that are also known as Turing patterns~\cite{alan1952morphogenesis} (see Fig.~\ref{fig:teaser}). Since it is natural that nodes on a graph also constitute local clusters (in terms of class labels), we conjecture that reaction-diffusion equations are suitable for GNNs.

Our proposed model, GREAD, consists of three parts: an encoder, a reaction-diffusion layer, and an output layer (cf. Eqs.~\eqref{eq:end} to~\eqref{eq:out}). The reaction-diffusion layer has seven different types in its core part as shown in Eq.~\eqref{eq:reac}: i) Fisher (F), ii) Allen-Cahn (AC), iii) Zeldovich (Z), iv) Blurring-sharpening (BS), v) Source (S), vi) Filter Bank (FB), and vii) Filter Bank* (FB*). The first three reaction-diffusion equations are widely used in many natural science domains, e.g., biology, combustion, and so on, and the last three are used by some recent GNN methods~\cite{xhonneux2019CGNN,thorpe2022grands,luan2022ACMGCN}. In particular, the blurring-sharpening (BS) equation is designed by us and marks the best accuracy in many cases (cf. Table~\ref{tab:summary}).

For our experiments, we consider 6 heterophilic and 3 homophilic datasets --- heterophilic (resp. homophilic) means that neighboring nodes tend to have different (resp. similar) classes. We also compare our method with a comprehensive set of 28 baselines, which covers early to recent GNNs. Our contributions can be summarized as follows:
\begin{compactenum}
    \item We design a reaction-diffusion layer that incorporates seven types of reaction equations, including one type, called BS, proposed by us.
    \item We carefully integrate the seven reaction equation types into our GNN method and customize its overall architecture for better accuracy. For instance, we use a soft adjacency matrix generating method, which shows a synergistic effect with the reaction-diffusion layer.
    \item We consider a comprehensive set of 9 datasets and 28 baselines. Our method marks the best accuracy in many cases. The ranking and accuracy averaged over all the datasets are summarized in Table~\ref{tab:summary}.
\end{compactenum}

\section{Preliminaries \& Related Work}
We first describe the meaning of the reaction-diffusion equation and various important GNN designs, followed by neural ordinary differential equations.

\subsection{Reaction-Diffusion Equations}
Reaction–diffusion equations are typically used to model the spatial and temporal change of the concentration of one or more chemical substances, i.e., substances are transformed into each other via local chemical reactions and spread out over a surface in space via diffusion. They are also frequently observed in other fields, such as biology, geology, physics (neutron diffusion theory), and ecology. In the field of graph machine learning, diffusion (resp. reaction) processes are typically carried out by applying low-pass (resp. high-pass) filters to graphs, which also corresponds to image blurring (resp. sharpening).

\subsection{Graph Neural Networks}
\paragraph{Notation}
Let $\mathcal{G}=\{\mathcal{V}, \mathcal{E}\}$ be a graph with node set $\mathcal{V}$ and edge set $\mathcal{E}$. The nodes are associated with a feature matrix $\mathbf{X}\in \mathbb{R}^{|\mathcal{V}| \times F}$, where $|\mathcal{V}|$ denotes the number of nodes and $F$ denotes the number of input features. $\mathbf{A}^{raw}\in\{0,1\}^{|\mathcal{V}| \times |\mathcal{V}|}$ is the adjacency matrix, where $\mathbf{A}^{raw}_{[i,j]}$ means the $(i,j)$-th element. The nodes are labelled by the index $i\in \mathcal{V}$, and one-hop neighborhood of each node is denoted as $\mathcal{N}_i$. 
The symmetric normalized Laplacian matrix, a commonly used feature aggregation matrix in GNNs, is defined as $\mathbf{L} := \mathbf{I}-\mathbf{D}^{-1/2}\mathbf{A}^{raw}\mathbf{D}^{-1/2}=\mathbf{I}-\mathbf{A}$, where the diagonal degree matrix of $\mathbf{A}^{raw}$ is $\mathbf{D}$, and $\mathbf{A} := \mathbf{D}^{-1/2}\mathbf{A}^{raw}\mathbf{D}^{-1/2}$ is the symmetric normalized adjacency matrix --- note that $\mathbf{A} \in [0,1]^{|\mathcal{V}| \times |\mathcal{V}|}$.

\paragraph{Graph Representation Learning}
GNNs~\cite{kipf2017GCN,velickovic2018GAT,hamilton2017graphSAGE,Wu2019SGC,zhu2020ssgc} have many variants and applications. We focus on a brief introduction to the representation learning for nodes in supervised or semi-supervised classification tasks. Most existing approaches follow the message-passing framework constructed by stacking layers that propagate and aggregate node features.

The neighbor aggregation used in many existing GNNs implicitly exploits homogeneity and often fails to generalize to non-homogeneous graphs. Many existing GNNs operate as low-pass graph filters~\cite{balcilar2021analyzing} that smooth features over the graph topology, which produces similar representations and as a result, similar predictions for neighboring nodes~\cite{tiezzi2021deep,oono2020oversmoothing,li2018deeper}. Various GNNs were proposed to improve performance in low-homophily settings~\cite{pei2020geomGCN,abu2019mixhop,zhu2020h2gcn,chien2021GPRGNN,he2021bernnet,lim2022LINKX,luan2022ACMGCN,bodnar2022Sheaf,giovanni2022GRAFF,li2022GloGNN} and alleviate the oversmoothing problem.~\cite{xu2018jknet,chen2020gcnii,zhao2020pairnorm,rusch2022gcon}.


\paragraph{Diffusion on Graphs and Continuous GNNs}

The diffusion on graphs has recently been actively used in various applications~\cite{freidlin1993diffusion,freidlin2000diffusion}, including data clustering~\cite{belkin2003laplacian,coifman2005geometric}, image processing~\cite{desquesnes2013eikonal,elmoataz2008nonlocal,gilboa2009nonlocal}, and so on. It is to apply the following diffusion process to the feature matrix $\mathbf{X}$ of a graph:
\begin{align}
\frac{d\mathbf{X}(t)}{dt} &:= div (\mathbf{A}(\mathbf{X}(t))) \nabla \mathbf{X}(t)) = -\mathbf{L}\mathbf{X}(t), \label{eq:heat_matrix}
\end{align}where $div$ and $\nabla$ are the divergence and the gradient operators, respectively. The initial features are evolved under the diffusion process to have the final representation. The diffusion equation and its unit-step Euler discretization can be defined as follows:
\begin{align}
    \mathbf{X}(t+1) = \mathbf{X}(t)-\mathbf{L}\mathbf{X}(t) = (\mathbf{I}-\mathbf{L})\mathbf{X}(t).\label{eq:euler_disc}
\end{align}

This is similar to GCN~\cite{kipf2017GCN} where the following augmented diffusion process with a weight matrix $\mathbf{W}$ and a nonlinear activation $\sigma$ is used:
\begin{align}
\sigma((\mathbf{I}-\mathbf{L})\mathbf{X}(t)\mathbf{W}).\label{eq:gcn-diffusion}
\end{align}

From this perspective, several papers have proposed continuous-depth GNNs~\cite{wang2021dgc,choi2021ltocf,hwang2021climate,thorpe2022grands,choi2023climate} inspired by the graph diffusion equation. One recent work is GRAND~\cite{chamberlain2021grand}, which parameterizes the diffusion equation on graphs with a neural network. BLEND~\cite{chamberlain2021blend} used a non-euclidean diffusion equation (known as Beltrami flow) to solve a joint positional feature space problem. These approaches contribute to non-trivial improvements in graph machine learning. We extend the diffusion to the reaction-diffusion equation in this work.


\subsection{Neural Ordinary Differential Equations (NODEs)}
Neural ordinary differential equations (NODEs)~\cite{chen2018NODE} solve the initial value problem (IVP), which involves a Riemann integral problem, to calculate $\mathbf{h}(t_{i+1})$ from $\mathbf{h}(t_i)$:
\begin{align}\label{eq:node}
\mathbf{h}(t_{i+1}) = \mathbf{h}(t_i) + \int_{t_i}^{t_{i+1}} f(\mathbf{h}(t_i), t;\mathbf{\theta}_f) dt,
\end{align}where the neural network parameterized by $\mathbf{\theta}_f$ approximates the time-derivative of $\mathbf{h}$, i.e., $\dot{\mathbf{h}} \stackrel{\text{def}}{=}\frac{d\mathbf{h}(t)}{dt}$. We rely on various ODE solvers to solve the integral problem, from the explicit Euler method to the 4th order Runge--Kutta (RK4) method and the Dormand--Prince (DOPRI) method ~\cite{dormand1980dopri}. For instance, the Euler method is as follows:
\begin{align}\label{eq:euler}
\mathbf{h}(t + h) = \mathbf{h}(t) + \tau \cdot f(\mathbf{h}(t)),
\end{align}where $\tau$, which is usually smaller than 1, is a pre-configured step size. Eq.~\eqref{eq:euler} is identical to a residual connection when $h=1$ and therefore, NODEs are a continuous generalization of residual networks.



\section{Proposed Method}
After describing an overview of our method, we describe its detailed designs, followed by its training algorithm. The theoretical and empirical complexity analyses are in Appendix~\ref{a:comp} and~\ref{a:time}, respectively.

\subsection{Overview of GREAD}
Given a graph $\mathcal{G}$ with its node feature matrix $\mathbf{X}$, its symmetric normalized Laplacian matrix $\mathbf{L}$, and its symmetric normalized adjacency $\mathbf{A}$, the overall architecture of GREAD can be written as follows --- instead of $\mathbf{A}$, we can also use a generated soft adjacency matrix $\tilde{\mathbf{A}}$, which will be described in the next subsection:
\begingroup\makeatletter\def\f@size{9}\check@mathfonts
\def\maketag@@@#1{\hbox{\m@th\small\normalfont#1}}%
\begin{align}
    \mathbf{H}(0) &= \mathbf{e}(\mathbf{X})\, &\textrm{(Encoding layer),}\label{eq:end}\\
    \mathbf{H}(T) &= \mathbf{H}(0) + \int_0^T \mathbf{f}(\mathbf{H}(t))dt\, &\textrm{(Reac.-diff. layer),}\label{eq:rd}\\
    \hat{\mathbf{y}} &= \mathbf{o}(\mathbf{H}(T))\, &\textrm{(Output layer)},\label{eq:out}
\end{align}\endgroup where $\mathbf{f}(\mathbf{H}(t)) := \frac{d \mathbf{H}(t)}{dt} = -\alpha\mathbf{L}\mathbf{H}(t) + \beta\mathbf{r}(\mathbf{H}(t))$ is in the reaction-diffusion form. $\mathbf{r}(\mathbf{H})$ is a reaction term, and $\alpha$ and $\beta$ are trainable parameters to (de-)emphasize each term. $\mathbf{e}$ is an encoder embeds the node feature matrix $\mathbf{X}$ into an initial hidden state $\mathbf{H}(0)$. We then evolve the initial hidden state to $\mathbf{H}(T)$ via the reaction-diffusion equation of $\mathbf{f}$. The function $\mathbf{o}$ is an output layer for a downstream task, e.g., node classification. In particular, $\beta$ can be either a scalar or a vector parameter, where the scalar setting means that we apply the same reaction process to all nodes and in the vector setting, we apply different reaction processes with different coefficients to nodes.

The encoder $\mathbf{e}$ has a couple of fully-connected layers with rectified linear unit (ReLU) activations. The output layer $\mathbf{o}$ is typically a fully-connected layer, followed by a softmax activation for classification in our experiments.

In particular, we consider almost all existing reaction terms for $\mathbf{r}$, which is different from existing works that do not consider them in a thorough manner. In this perspective, our work is the most comprehensive study on reaction-diffusion GNNs to our knowledge. In the following subsection, we also show that some choices of the reaction term correspond to other famous models --- in other words, some other famous models are special cases of GREAD.

\subsection{Soft Adjacency Matrix Generation}
Given a graph $\mathcal{G}$, one can use its original symmetric normalized adjacency matrix $\mathbf{A} \in [0,1]^{|\mathcal{V}| \times |\mathcal{V}|}$ for $\mathbf{f}(\mathbf{H}(t))$. However, we also provide the method to generate a soft adjacency matrix, denoted $\tilde{\mathbf{A}} \in [0,1]^{|\mathcal{V}| \times |\mathcal{V}|}$ --- we use $\tilde{\mathbf{L}}$ to denote the Laplacian counterpart of $\tilde{\mathbf{A}}$. Our soft adjacency matrix plays a crucial role in learning diffusivity. Our reaction-diffusion layer uses the soft adjacency matrix to learn the diffusivity.

In order to generate such soft adjacency matrices, we use the scaled dot product method~\cite{vaswani2017attention}:
\begin{align}
    \tilde{\mathbf{A}}_{[i,j]} := softmax\Big(\frac{(\mathbf{W}_K\mathbf{H}_i)^T\mathbf{W}_Q\mathbf{H}_j}{d_K}\Big)\label{eq:soft},
\end{align}where $\tilde{\mathbf{A}}_{[i,j]}$ means the $(i,j)$-th element of $\tilde{\mathbf{A}}$, $\mathbf{W}_K$ and $\mathbf{W}_Q$ are trainable parameters, and $d_K$ is the scale factor. $\mathbf{H}_i, \mathbf{H}_j$ are trainable embedding vectors of nodes $i, j$.

\subsection{Reaction-diffusion Layer}
Eq.~\eqref{eq:rd} is our method's main processing layer, called the reaction-diffusion layer. Given the definition of $\mathbf{f}$, `$-\mathbf{L}\mathbf{H}(t)$' is a diffusion term that corresponds to the heat equation describing the spread of heat over $\mathcal{G}$ and has been used widely by various GNNs ~\cite{wang2021dgc,choi2021ltocf,chamberlain2021grand}. It is known that the diffusion term causes the problem of oversmoothing, which means that the last hidden states of nodes become too similar when applying only the diffusion processing too much. To this end, many models prefer shallow architectures that do not cause the oversmoothing problem~\cite{Wu2019SGC,kipf2017GCN} or use heuristic methods to prevent it~\cite{zhao2020pairnorm,Chen2018FastGCN,chen2020gcnii,li2019deepgcns,liu2020towards,huang2018adaptivesample,chen2018stochastictraining}.

In our case, we prevent the oversmoothing problem by adding the reaction term $\mathbf{r}$ and solving Eq.~\eqref{eq:rd} with ODE solvers~\cite{dormand1980dopri}. In other words, our reaction-diffusion layer is continuous, which is yet another distinguishing point of our method since many other models are based on discrete layers~\cite{kipf2017GCN,Bo2021fagcn,chien2021GPRGNN,zhu2020h2gcn,hamilton2017graphSAGE}. We consider the following options for $\mathbf{r}$:
\begingroup\makeatletter\def\f@size{9}\check@mathfonts
\def\maketag@@@#1{\hbox{\m@th\small\normalfont#1}}%
\begin{align}\label{eq:reac}
    \mathbf{r}(\mathbf{H}(t)) := \begin{cases} \mathbf{H}(t)\odot(1-\mathbf{H}(t)),\textrm{ if Fisher (F)}\\
    \mathbf{H}(t)\odot(1-\mathbf{H}(t)^{\circ 2}),\textrm{ if Allen-Cahn (AC)}\\
    \mathbf{H}(t)\odot(\mathbf{H}(t)-\mathbf{H}(t)^{\circ 2}),\textrm{ if Zeldovich (Z)}\\
    (\tilde{\mathbf{A}}-\tilde{\mathbf{A}}^2)\mathbf{H}(t),\textrm{ if Blurring-Sharpening (BS)}\\
    \mathbf{H}(0), \textrm{ if Source Term (ST)} \\
    \mathbf{L}\mathbf{H}(t), \textrm{ if Filter Bank (FB)} \\
    \mathbf{L}\mathbf{H}(t)+\mathbf{H}(t), \textrm{ if Filter Bank* (FB*)}
    \end{cases}
\end{align}\endgroup where `$\odot$' means the Hadamard product, and `$\circ 2$' means the Hadamard power.

The first three reaction terms, i.e., F, AC, and Z, are widely used in various domains. For instance, F is used to describe the spreading of biological populations~\cite{fisher1937wave}, and AC is used for describing the phase separation process in multi-component alloy systems, which includes order-disorder transitions~\cite{allen1979microscopic}. Z is a generalized equation that describes the phenomena that occur in combustion theory~\cite{gilding2004travelling}. The last BS is specially designed by us for GNNs, which we will describe shortly. ST is a case where the initial hidden state is added as a reaction term~\cite{xhonneux2019CGNN}. ST is not theoretically a reaction process, but we consider it as part of our model since their goals are the same, i.e., alleviating the notorious oversmoothing problem. FB means high-pass filters that correspond to reaction processes. By adding a high-pass filter, our reaction-diffusion layer acts like a filter bank holding the low and high-pass filters. FB* is a reaction term that also considers the identity channel $\mathbf{H}(t)$. 

\paragraph{Blurring-Sharpening (BS)} Given the reaction-diffusion layer in Eq.~\eqref{eq:rd}, the proposed blurring-sharpening (BS) process, whose time-derivative of $\mathbf{H}(t)$ will be defined in Eq.~\eqref{eq:bsl}, is to perform the blurring (diffusion) and the sharpening (reaction) operations alternately in the layer. 
We show that our proposed blurring-sharpening process reduces to a certain form of the reaction-diffusion process. Many GNNs can be generalized to the following blurring (or diffusion) process, i.e., the low-pass graph convolutional filtering for blurring. We also use the same blurring operation at first:
\begin{align}\begin{split}\label{eq:blur}
    \mathbf{B}(t+h) &= \mathbf{H}(t) -\tilde{\mathbf{L}}\mathbf{H}(t),\\
            &\Rightarrow \mathbf{H}(t) + (\tilde{\mathbf{A}}-\mathbf{I})\mathbf{H}(t),\\
            &\Rightarrow \tilde{\mathbf{A}}\mathbf{H}(t).
\end{split}\end{align}

We then propose to apply the following high-pass graph convolutional filtering or sharpening process to $\mathbf{B}(t+h)$. In other words, there is a sharpening process following the above blurring process in a layer as follows --- the full derivation is in Appendix~\ref{a:bs}:
\begin{align}\begin{split}\label{eq:bs}
    \mathbf{H}(t+h) &= \mathbf{B}(t+h) + \tilde{\mathbf{L}}(\mathbf{B}(t+h)),\\
            &\Rightarrow \mathbf{H}(t) - \tilde{\mathbf{L}}\mathbf{H}(t) + (\tilde{\mathbf{A}}-\tilde{\mathbf{A}}^2)\mathbf{H}(t).
\end{split}\end{align}

Therefore, we can derive the following difference equation:
\begin{align}
    \mathbf{H}(t+h) - \mathbf{H}(t) = - \tilde{\mathbf{L}}\mathbf{H}(t) + (\tilde{\mathbf{A}}-\tilde{\mathbf{A}}^2)\mathbf{H}(t).
\end{align}

After taking the limit of $h \rightarrow 0$ and adding the coefficients $\alpha, \beta$,
\begin{align}\label{eq:bsl}
   \mathbf{f}(\mathbf{H}(t)) :=  \frac{d \mathbf{H}(t)}{dt} = -\alpha \tilde{\mathbf{L}}\mathbf{H}(t) + \beta(\tilde{\mathbf{A}}-\tilde{\mathbf{A}}^2)\mathbf{H}(t),
\end{align}which is a reaction-diffusion equation where $\mathbf{r}(\mathbf{H}(t)) := (\tilde{\mathbf{A}}-\tilde{\mathbf{A}}^2)\mathbf{H}(t)$. Therefore, our proposed method, BS, uses the reaction-diffusion layer in Eq.~\eqref{eq:rd} with the specific time-derivative definition of Eq.~\eqref{eq:bsl}.


\begin{table*}[ht!]
    \small
    \centering
    \setlength{\tabcolsep}{2pt}
    \caption{Benchmark dataset properties and statistics}
    \begin{tabular}{c ccccccccc}\toprule
        Dataset         & Texas & Wisconsin & Cornell & Film   & Squirrel & Chameleon & Cora  & Citeseer & PubMed \\ \midrule
        Classes         & 5     & 5         & 5       & 5      & 5        & 5         & 6     & 7        & 3      \\
        Features        & 1,703 & 1,703     & 1,703   & 932    & 2,089    & 235       & 1,433 & 3,703    & 500    \\
        Nodes           & 183   & 251       & 183     & 7,600  & 5,201    & 2,277     & 2,708 & 3,327    & 19,717 \\
        Edges           & 279   & 466       & 277     & 26,752 & 198,353  & 31,371    & 5,278 & 4,552    & 44,324 \\
        Hom. ratio      & 0.11  & 0.21      & 0.30    & 0.22   & 0.22     & 0.23      & 0.81  & 0.74     & 0.80  \\
        \bottomrule
    \end{tabular}
    \label{tab:data}
\end{table*}

\subsection{Training Algorithm}
We use Alg.~\eqref{alg:train} to train our proposed model. The full training process minimizes the cross-entropy loss:
\begin{align}\label{eq:loss}
    \mathcal{L} := \sum_{i}^{n}{\mathbf{y}^T_i\log{\hat{\mathbf{y}_i}}},
\end{align} where $\mathbf{y}_i$ is the one-hot ground truth vector of $i$-th training sample, and $\hat{\mathbf{y}}_i$ is its inference outcome by our model.

\begin{algorithm}[t]
\small
\SetAlgoLined
\caption{How to train our proposed GREAD}\label{alg:train}
\KwIn{Training data $D_{train}$, Validating data $D_{val}$, Maximum iteration number $max\_iter$}
Initialize model parameters $\mathbf{\theta}$;

$k \gets 0$;

\While {$k < max\_iter$}{
    Construct a mini-batch $B$ from $D_{train}$\;
    
    Train $\mathbf{\theta}$\ with Eq.~\eqref{eq:loss} and $B$;\label{alg:train1}
    
    Validate and update the best parameters $\mathbf{\theta}^*$ with $D_{val}$\;
    
    $k \gets k + 1$;
}
\Return $\mathbf{\theta}^*$;
\end{algorithm}

\subsection{Comparison with GNNs}

When ST is used, GREAD is analogous to GCNII in the perspective that both methods inject the initial hidden state. GREAD, FA-GCN, and GPR-GNN differ in how to utilize low and high-pass filters. FA-GCN learns edge-level aggregation weights as in GAT but allows negative weights. GPR-GNN uses learnable weights that can be both positive and negative for feature propagation. Those enable FA-GCN and GPR-GNN to adapt to heterophilic graphs and to handle both high and low-frequency parts of graph signals. However, GREAD-BS sharpens low-pass filtered signals following our developed reaction-diffusion system. GREAD-BS also adaptively adjusts each term.

We also compare with some continuous-time GNN models. CGNN can be derived from the reaction-diffusion layer in Eq.~\eqref{eq:rd} with $\mathbf{L}$ by setting $\mathbf{f}$ with $\mathbf{r}(\mathbf{H}(t)) := \mathbf{H}(0)$ and using a weight parameter $\mathbf{W}$:
\begin{align}
\mathbf{f}(\mathbf{H}(t))_{\text{CGNN}} := -\mathbf{L}\mathbf{H}(t) + \mathbf{H}(t)\mathbf{W} + \mathbf{H}(0).
\end{align}
The linear GRAND model corresponds to using only our diffusion process:
\begin{align}
\mathbf{f}(\mathbf{H}(t))_{\text{GRAND}} := -\tilde{\mathbf{L}}\mathbf{H}(t) = -(\mathbf{I}-\tilde{\mathbf{A}})\mathbf{H}(t).
\end{align}
We note that two continuous models can not capture high frequency parts. In particular, GRAND does not use any reaction term.

\section{Experiments}
We first compare our method with other baselines for node classification tasks. We then discuss the ability of mitigating oversmoothing on a synthetic graph and show the experiment with different heterophily levels on other synthetic graphs. Our code is available at \url{https://github.com/jeongwhanchoi/GREAD}.

\begin{table*}[ht]
    \small
    \centering
    \setlength{\tabcolsep}{2pt}
    \caption{Results on real-world datasets: mean $\pm$ std. dev. accuracy for 10 different data splits. We show the best three methods in \BEST{red} (first), \SECOND{blue} (second), and \THIRD{purple} (third). Other missing 16 baselines are in Appendix~\ref{a:full}.}
    \begin{tabular}{c ccccccccc c}\toprule
        Dataset     & Texas      & Wisconsin  & Cornell    & Film       & Squirrel   & Chameleon  & Cora       & Citeseer   & PubMed & Avg.\\ \midrule
        Geom-GCN	& 66.76\std{±2.72} & 64.51\std{±3.66} & 60.54\std{±3.67} & 31.59\std{±1.15} & 38.15\std{±0.92} & 60.00\std{±2.81} & 85.35\std{±1.57} & \BEST{78.02\std{±1.15}} & 89.95\std{±0.47} & 63.87\\
        H2GCN	    & 84.86\std{±7.23} & 87.65\std{±4.98} & 82.70\std{±5.28} & 35.70\std{±1.00} & 36.48\std{±1.86} & 60.11\std{±2.15} & 87.87\std{±1.20} & 77.11\std{±1.57} & 89.49\std{±0.38} & 71.33\\
        GGCN        & 84.86\std{±4.55} & 86.86\std{±3.29} & 85.68\std{±6.63} & 37.54\std{±1.56} & 55.17\std{±1.58} & \SECOND{71.14\std{±1.84}} & 87.95\std{±1.05} & 77.14\std{±1.45} & 89.15\std{±0.37} & 75.05\\
        LINKX       & 74.60\std{±8.37} & 75.49\std{±5.72} & 77.84\std{±5.81} & 36.10\std{±1.55} & \BEST{61.81\std{±1.80}} & 68.42\std{±1.38} & 84.64\std{±1.13} & 73.19\std{±0.99} & 87.86\std{±0.77} & 71.11\\
        GloGNN      & 84.32\std{±4.15} & 87.06\std{±3.53} & 83.51\std{±4.26} & 37.35\std{±1.30} & \THIRD{57.54\std{±1.39}} & 69.78\std{±2.42} & 88.31\std{±1.13} & 77.41\std{±1.65} & 89.62\std{±0.35} & 74.99\\
        ACM-GCN	    & 87.84\std{±4.40} & \SECOND{88.43\std{±3.22}} & 85.14\std{±6.07} & 36.28\std{±1.09} & 54.40\std{±1.88} & 66.93\std{±1.85} & 87.91\std{±0.95} & 77.32\std{±1.70} & 90.00\std{±0.52} & 74.92\\
        \midrule
        GCNII	    & 77.57\std{±3.83} & 80.39\std{±3.40} & 77.86\std{±3.79} & 37.44\std{±1.30} & 38.47\std{±1.58} & 63.86\std{±3.04} & \THIRD{88.37\std{±1.25}} & 77.33\std{±1.48} & \SECOND{90.15\std{±0.43}} & 70.16\\
        \midrule
        CGNN	    & 71.35\std{±4.05} & 74.31\std{±7.26} & 66.22\std{±7.69} & 35.95\std{±0.86} & 29.24\std{±1.09} & 46.89\std{±1.66} & 87.10\std{±1.35} & 76.91\std{±1.81} & 87.70\std{±0.49} & 63.96\\
        GRAND       & 75.68\std{±7.25} & 79.41\std{±3.64} & 82.16\std{±7.09} & 35.62\std{±1.01} & 40.05\std{±1.50} & 54.67\std{±2.54} & 87.36\std{±0.96} & 76.46\std{±1.77} & 89.02\std{±0.51} & 68.94\\
        BLEND       & 83.24\std{±4.65} & 84.12\std{±3.56} & 85.95\std{±6.82} & 35.63\std{±1.01} & 43.06\std{±1.39} & 60.11\std{±2.09} & 88.09\std{±1.22} & 76.63\std{±1.60} & 89.24\std{±0.42} & 71.79\\
        Sheaf       & 85.05\std{±5.51} & \BEST{89.41\std{±4.74}} & 84.86\std{±4.71} & \SECOND{37.81\std{±1.15}} & 56.34\std{±1.32} & 68.04\std{±1.58} & 86.90\std{±1.13} & 76.70±\std{1.57} & 89.49\std{±0.40} & 75.06\\
        GRAFF       & \THIRD{88.38\std{±4.53}} & 87.45\std{±2.94} & 83.24\std{±6.49} & 36.09\std{±0.81} & 54.52\std{±1.37} & \THIRD{71.08\std{±1.75}} & 87.61\std{±0.97} & 76.92±\std{1.70} & 88.95\std{±0.52} & 74.92\\
        \midrule
        \textbf{GREAD-BS}
                    & \SECOND{88.92\std{±3.72}} & \BEST{89.41\std{±3.30}} & \SECOND{86.49\std{±7.15}} 
                    & \BEST{37.90\std{±1.17}} & \SECOND{59.22\std{±1.44}} & \BEST{71.38\std{±1.31}} 
                    & \BEST{88.57\std{±0.66}} & \SECOND{77.60\std{±1.81}} & \BEST{90.23\std{±0.55}} & 76.64\\
        \textbf{GREAD-F} 
                    & \BEST{89.73\std{±4.49}} & 86.47\std{±4.84} & \SECOND{86.49\std{±5.13}}
                    & 36.72\std{±0.66}	& 46.16\std{±1.44} & 65.20\std{±1.40} 
                    & 88.39\std{±0.91} & 77.40\std{±1.54}	& 90.09\std{±0.31} & 74.13\\
        \textbf{GREAD-AC}
                    & 85.95\std{±2.65} & 86.08\std{±3.56} & \BEST{87.03\std{±4.95}} 
                    & 37.21\std{±1.10} & 45.10\std{±2.11} & 65.09\std{±1.08} 
                    & 88.29\std{±0.67} & 77.38\std{±1.53} & 90.10\std{±0.36} & 73.71\\
        \textbf{GREAD-Z} 
                    & 87.30\std{±5.68} & 86.29\std{±4.32} & 85.68\std{±5.41} 
                    & 37.01\std{±1.11} & 46.25\std{±1.72} & 62.70\std{±2.30} 
                    & 88.31\std{±1.10} & 77.39\std{±1.90} & 90.11\std{±0.27} & 73.45\\
        \textbf{GREAD-ST} 
                        & 81.08\std{±5.67} & 86.67\std{±3.01} & \THIRD{86.22\std{±5.98}}
                        & 37.66\std{±0.90} & 45.83\std{±1.40} & 63.03\std{±1.32} 
                        & \SECOND{88.47\std{±1.19}} & 77.25\std{±1.47} & \THIRD{90.13\std{±0.36}} & 72.93\\
        \textbf{GREAD-FB}   
                        & 86.76\std{±5.05} & 87.65\std{±3.17} & \THIRD{86.22\std{±5.85}} 
                        & 37.40\std{±0.55} & 50.83\std{±2.27} & 66.05\std{±1.21} 
                        & 88.03\std{±0.78} & 77.28\std{±1.73} & 90.07\std{±0.45} & 74.48\\
        \textbf{GREAD-FB*}   
                        & 87.03\std{±3.97} & \THIRD{88.04\std{±1.63}} & 85.95\std{±5.64} 
                        & \THIRD{37.70\std{±0.51}} & 50.57\std{±1.52} & 65.83\std{±1.10} 
                        & 88.01\std{±0.80} & \THIRD{77.42\std{±1.93}} &  90.08\std{±0.46} & 74.51\\
        \bottomrule
    \end{tabular}
    \label{tab:result}
\end{table*}

\subsection{Node Classification on Real-world Datasets}
\paragraph{Real-world Datasets}
We now evaluate the performance of GREAD and existing GNNs on a variety of real-world datasets. We consider 6 heterophilic datasets with low homophily ratios used in~\cite{pei2020geomGCN}: i,ii) Chameleon, Squirrel~\cite{benedek2021musae}, iii) Film~\cite{tang2009social}, iv, v, vi) Texas, Wisconsin and Cornell from WebKB. We also test on 3 homophilic graphs with high homophily ratios: i) Cora~\cite{mccallum2000automating}, ii) CiteSeer~\cite{sen2008collective}, iii) PubMed~\cite{yang2016revisiting}. Table~\ref{tab:data} summarizes the number/size of nodes, edges, classes, features, and the homophily ratio. We use the dataset splits taken from~\cite{pei2020geomGCN}. We report the mean and standard deviation accuracy after running each experiment with 10 fixed train/val/test splits.

\begin{figure}[t]
    \centering
    \subfigure[$t=0$]{\includegraphics[width=0.32\columnwidth]{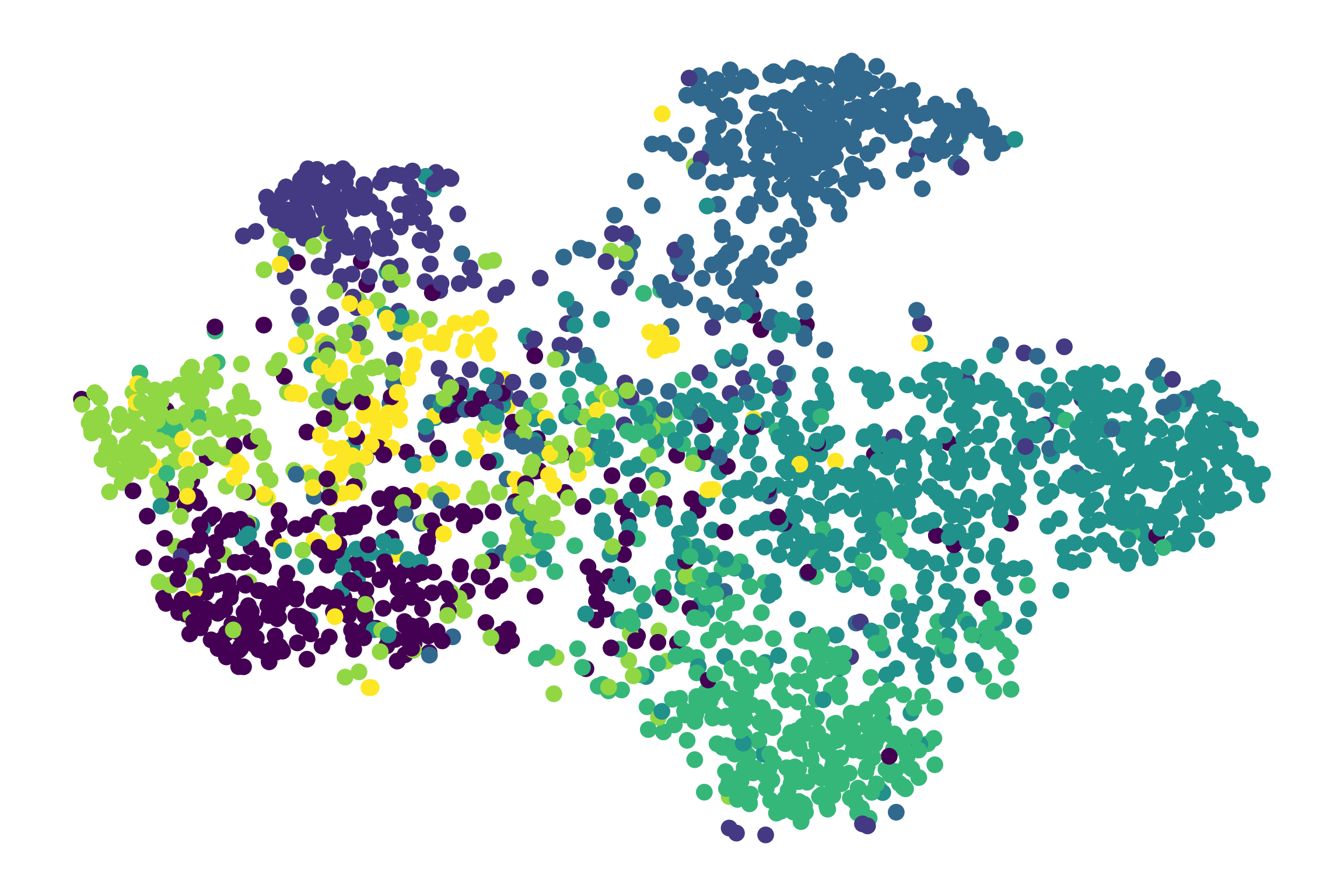}}
    \subfigure[$t=1$]{\includegraphics[width=0.32\columnwidth]{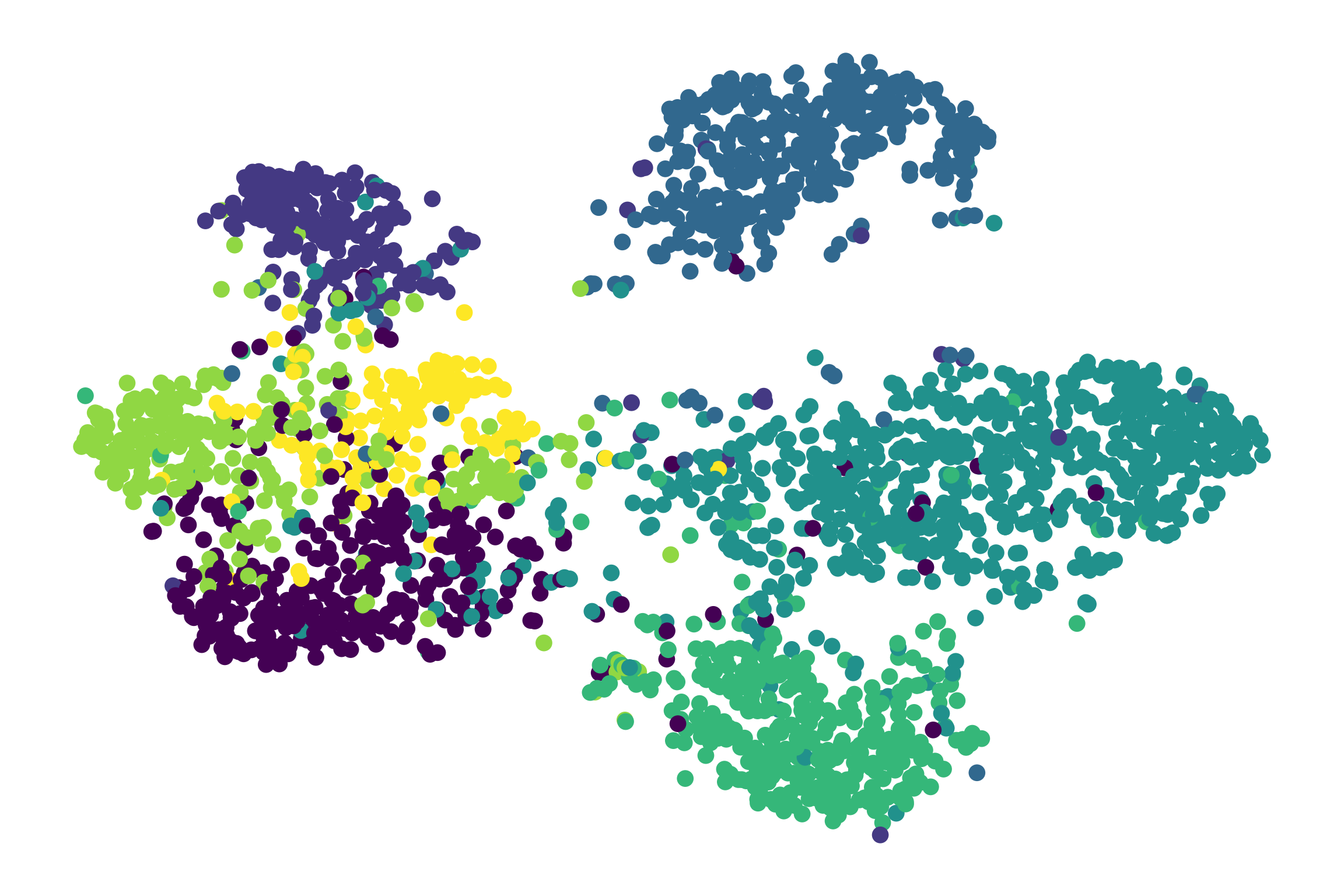}}
    \subfigure[$t=3.5$]{\includegraphics[width=0.32\columnwidth]{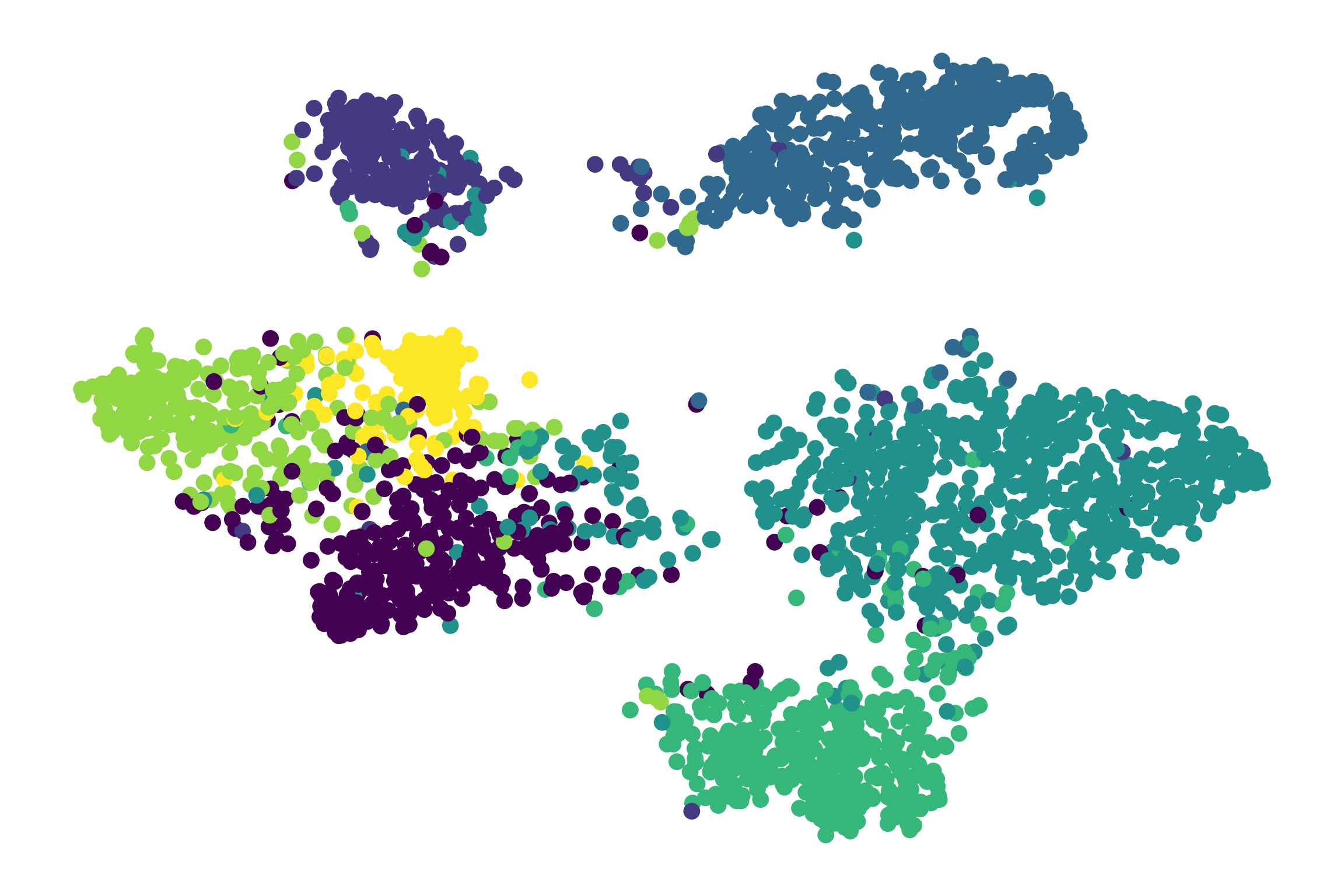}}
    \caption{The snapshots of the evolution process of the node feature at various ODE time points in GREAD for Cora. Different colors correspond to different ground truth classes. More visualizations in other datasets are in Appendix~\ref{a:visualization}.}
    \label{fig:cora_tsne}
\end{figure}

\paragraph{Baselines} We use a comprehensive set of baselines classified into the following four groups:
\begin{compactenum}
    \item In the first group of baselines, we consider classical GNN methods: ChebNet~\cite{defferrard2016chebnet}, GCN~\cite{kipf2017GCN}, GAT~\cite{velickovic2018GAT}, GraphSAGE~\cite{hamilton2017graphSAGE}, and SGC~\cite{Wu2019SGC}.
    \item The next group includes the GNN methods designed for heterophilic settings: MixHop~\cite{abu2019mixhop}, Geom-GCN~\cite{pei2020geomGCN}, H2GCN~\cite{zhu2020h2gcn}, FA-GCN~\cite{Bo2021fagcn}, GPR-GNN~\cite{chien2021GPRGNN}, WRGAT~\cite{suresh2021WRGAT}, GGCN~\cite{yan2021GGCN}, LINKX~\cite{lim2022LINKX}, GloGNN~\cite{li2022GloGNN} and ACM-GCN~\cite{luan2022ACMGCN}.
    \item The third group has GNN methods tackling the oversmoothing problem: PairNorm~\cite{zhao2020pairnorm}, JKNet~\cite{xu2018jknet}, GCNII~\cite{chen2020gcnii}, and GCON~\cite{rusch2022gcon}.
    \item The last group contains continuous-time GNN methods: GDE~\cite{poli2019gde}, CGNN~\cite{xhonneux2019CGNN}, GRAND~\cite{chamberlain2021grand}, BLEND~\cite{chamberlain2021blend}, ACMP~\cite{wang2023acmp}, Sheaf~\cite{bodnar2022Sheaf}, and GRAFF~\cite{giovanni2022GRAFF}.
\end{compactenum}

\paragraph{Hyperparameters} For our method, we test with the following hyperparameter configurations: we train for 200 epochs using the Adam optimizer. The detailed search space and other hyperparameters are in Appendix~\ref{a:setting}. We also list the best hyperparameter configuration for each data in Appendix~\ref{a:setting}. If a baseline's accuracy is known and its experimental environments are the same as ours, we use the officially announced accuracy. If not, we execute a baseline using its official codes and the hyperparameter search procedures based on their suggested hyperparameter ranges.

\begin{table}[t]
     \small
     \centering
     \setlength{\tabcolsep}{2pt}
     \caption{Ablation study on soft adjacency matrix. More results in other datasets are in Appendix~\ref{a:ablation}.}
     \begin{tabular}{cc cccc ccc}\toprule
        Dataset & $\mathbf{A}$ &  BS & F & AC & Z & ST & FB & FB*\\ \midrule
        \multirow{2}{*}{Cornell}
        & OA   &  85.14 & 85.41 & 83.51 & 83.78 & 85.41 & 85.19 & 83.90 \\
        & SA   &  \textbf{86.49} & \textbf{86.49} & \textbf{87.03} & \textbf{85.68} & \textbf{86.22} & \textbf{86.22} & \textbf{85.95}\\
        \midrule
        \multirow{2}{*}{Film}
        & OA   &  37.24 & 36.68 & 35.93 & 36.04 & 37.43 & 37.18 & 37.13\\
        & SA   &  \textbf{37.90} & \textbf{37.20 }& \textbf{37.21} & \textbf{37.01} & \textbf{37.66} & \textbf{37.40} & \textbf{37.70}\\
        \bottomrule
     \end{tabular}
     \label{tab:soft}
 \end{table}

 \begin{table}[t]
     \small
     \centering
     \setlength{\tabcolsep}{2pt}
     \caption{Ablation study on $\beta$. More results in other datasets are in Appendix~\ref{a:ablation}.}
     \begin{tabular}{cc cccc ccc}\toprule
        Dataset & $\mathbf{A}$ &  BS & F & AC & Z & ST & FB & FB*\\ \midrule
        \multirow{2}{*}{Texas}
        & SC    &  81.35 & 84.05 & 85.41 & \textbf{87.30} & 79.73 & 76.95 & 77.08\\
        & VC    &  \textbf{88.92} & \textbf{89.73} & 85.95 & 86.49 & \textbf{81.08} & \textbf{86.76} & \textbf{87.03}\\
        \midrule
        \multirow{2}{*}{Wisconsin}
        & SC    &  84.71 & 85.69 & \textbf{86.08} & 84.51 & 83.52 & 84.24 & 86.21\\
        & VC    &  \textbf{89.41} & \textbf{86.47} & 85.69 & \textbf{86.29} & \textbf{86.67} & \textbf{87.65} & \textbf{88.04}\\
        \bottomrule
     \end{tabular}
     \label{tab:beta}
 \end{table}
 
\paragraph{Experimental Results} Table~\ref{tab:summary} shows the average ranking and accuracy in all the real-world datasets. GREAD-BS is ranked at the top with the average ranking of 1.56. GREAD-BS shows a clearly higher ranking in comparison with GloGNN and others. In Fig.~\ref{fig:cora_tsne}, we visualize the hidden node features at each ODE time step of Eq.~\eqref{eq:rd}, and the reaction-diffusion processes of GREAD lead to local clusters after several steps.

Table~\ref{tab:result} presents the detailed classification performance. As reported, our method marks the best accuracy in all cases except for Squirrel and Citeseer. GloGNN and Sheaf show comparable accuracy values from time to time. However, there are no existing methods that are as stable as GREAD-BS. For example, GCNII shows reasonably high accuracy in homophilic datasets, but not heterophilic ones. Sheaf shows the best or the second-best place in merely two cases. While GREAD-BS is the best method overall, GREAD-F is the best method for Texas and is the second-best for Cornell. GREAD-AC marks the best accuracy on Cornell.

\paragraph{Ablation Studies} We conduct ablation studies about the soft adjacency matrix generation. GREAD can use both the original symmetric normalized adjacency matrix, denoted as OA, and the soft adjacency matrix denoted as SA. We compare both options. As reported in Table~\ref{tab:soft}, SA increases the model accuracy in Cornell and Film.

Next, we also perform the ablation study on $\beta$. $\beta$ can be either a scalar parameter (denoted as SC) or a learnable vector parameter (denoted as VC). We compare them in Table~\ref{tab:beta}. VC shows effectiveness for most cases. The VC setting creates a reaction-diffusion process rich enough to classify nodes, as shown in Fig.~\ref{fig:cora_tsne}.

 \begin{figure}[t]
    \centering
    \includegraphics[width=\columnwidth]{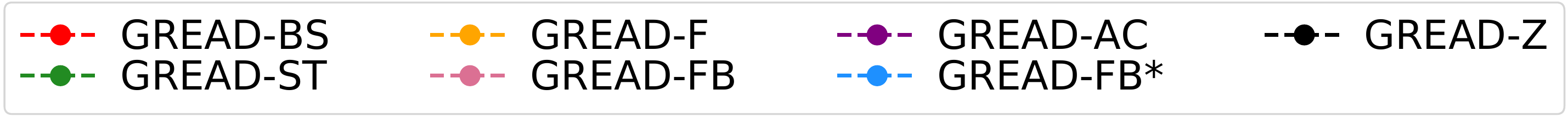}
    \subfigure[Chameleon]{\includegraphics[width=0.48\columnwidth]{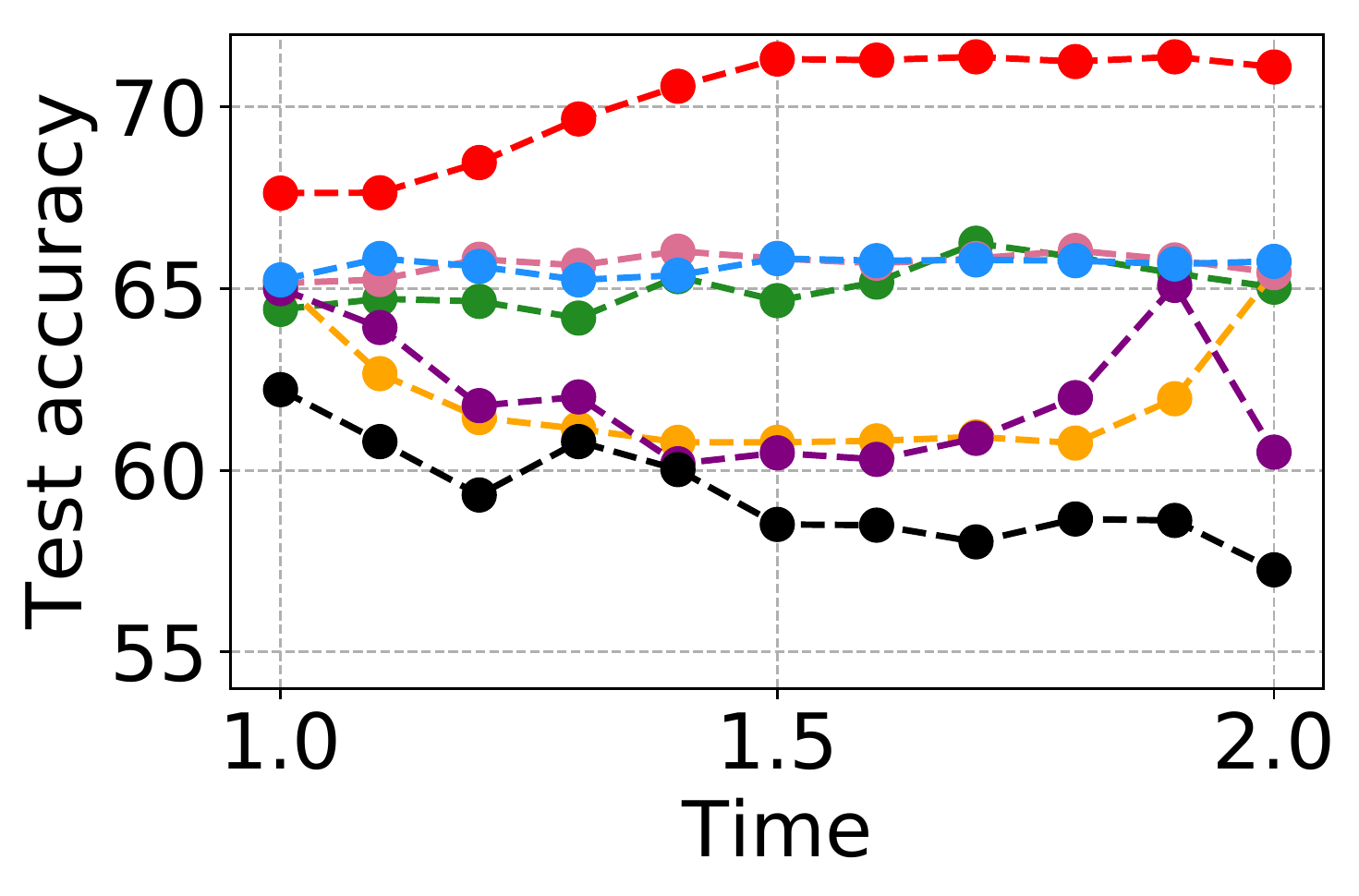}}
    \subfigure[Cora]{\includegraphics[width=0.48\columnwidth]{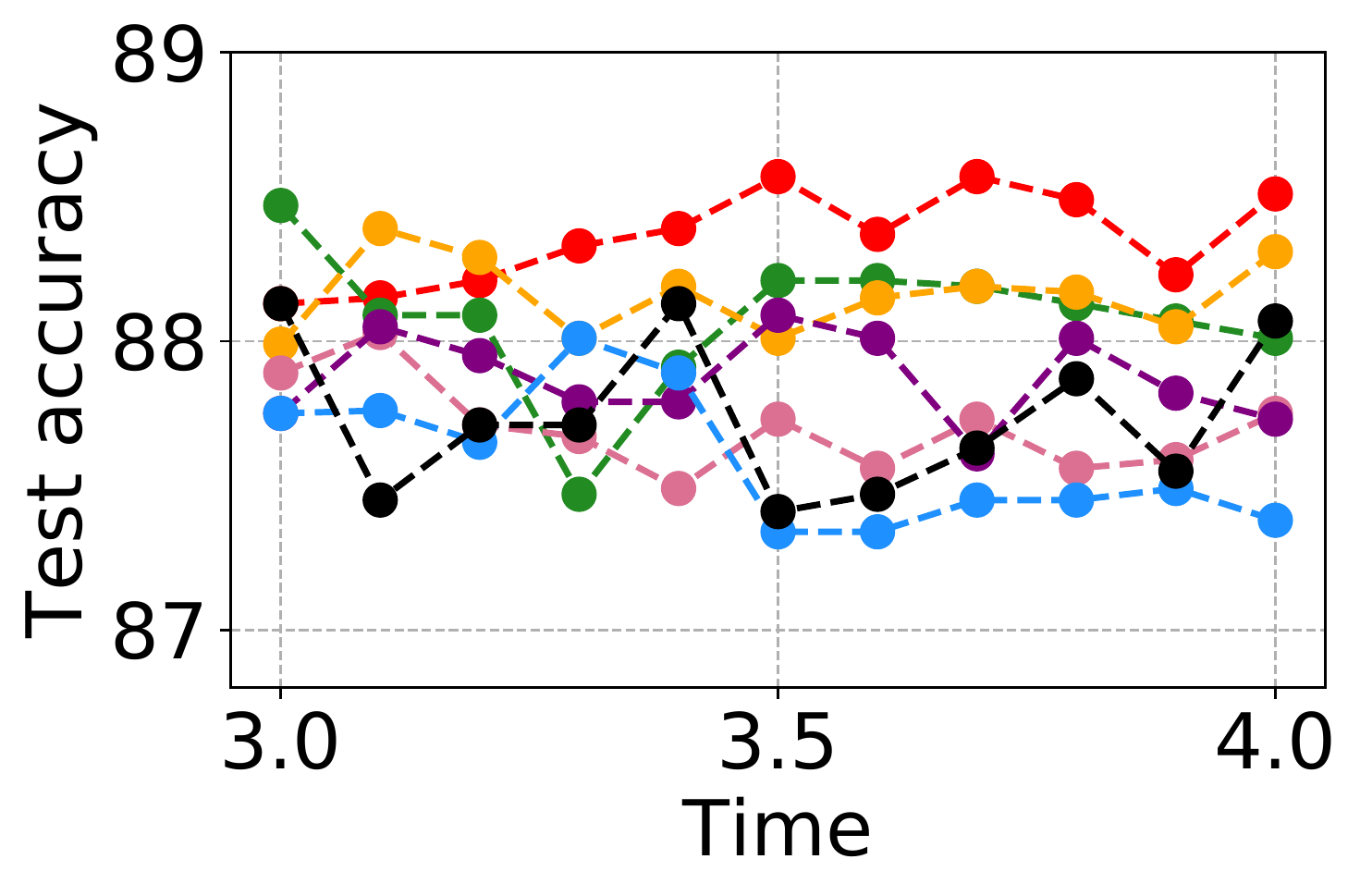}}
    \caption{Sensitivity to $T$. More results in other datasets are in Appendix~\ref{a:sensitivity}.}
    \label{fig:sens_T}
\end{figure}
\begin{figure}[t]
    \centering
    \includegraphics[width=\columnwidth]{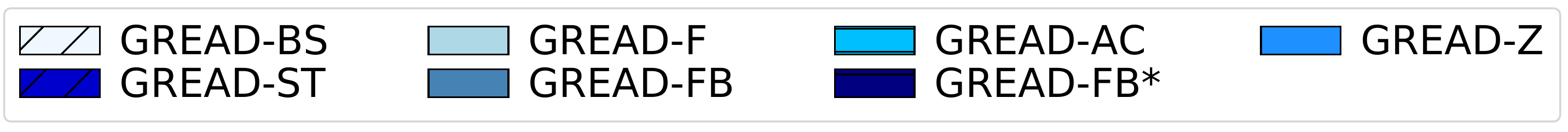}
    \subfigure[Chameleon]{\includegraphics[width=0.48\columnwidth]{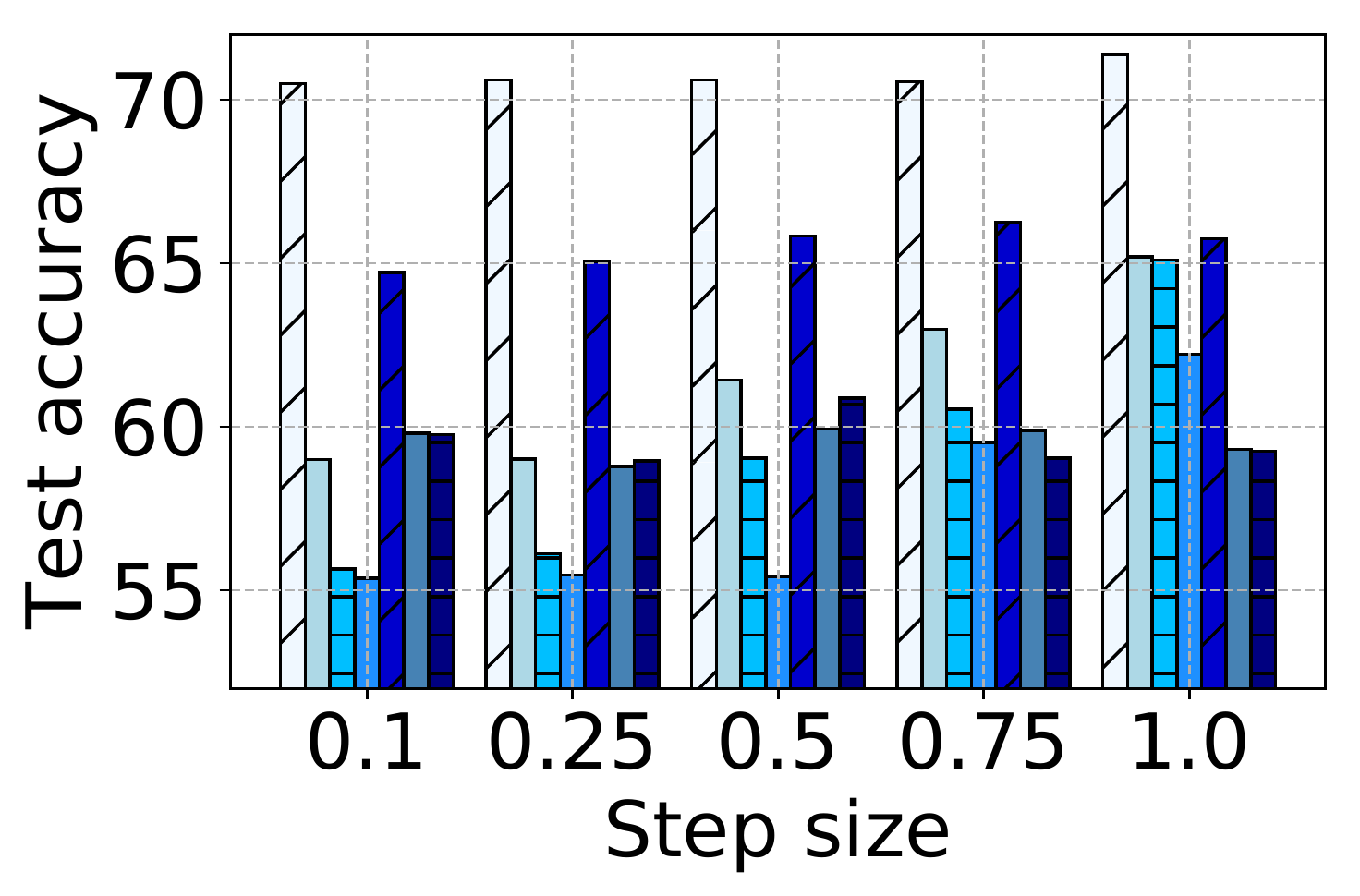}}
    \subfigure[Cora]{\includegraphics[width=0.48\columnwidth]{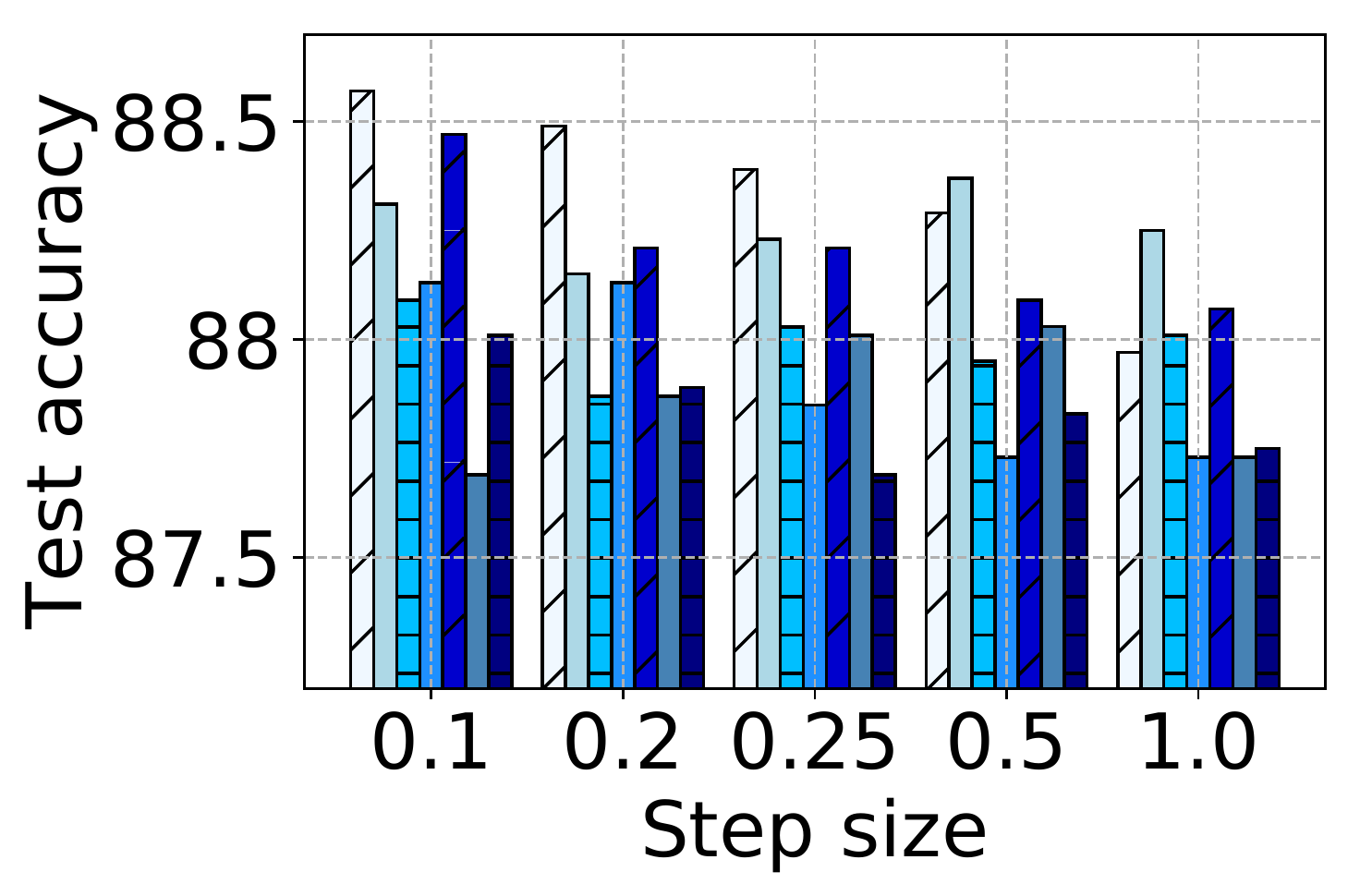}}
    \caption{Sensitivity to the ODE step size $\tau$. More results in other datasets are in Appendix~\ref{a:sensitivity}.}
    \label{fig:sens_step}
\end{figure}

\paragraph{Sensitivity w.r.t. the Terminal Integration Time $T$}
By varying $T$ in Eq.~\eqref{eq:rd}, we investigate how the model accuracy changes. The detailed results are in Fig.~\ref{fig:sens_T}. In Chameleon, GREAD-BS achieves the highest mean test accuracy at $T=1.9$. 

\paragraph{Sensitivity w.r.t. the ODE Step Size $\tau$}
Fig.~\ref{fig:sens_step} shows the mean test accuracy by varying the step size $\tau$ of RK4.
In Chameleon, GREAD-BS shows stable test accuracy at all the step sizes, whereas in Cora, GREAD-BS tends to show higher accuracy with larger step sizes.

\subsection{Oversmoothing and Dirichlet Energy}
\paragraph{The Dirichlet Energy} We can analyze the degree of oversmoothing~\cite{nt2019revisiting,oono2020oversmoothing} from the perspective of the Dirichlet energy~\cite{rusch2022gcon,rusch2023survey}. The Dirichlet energy $E(\mathbf{H}, \mathbf{A}^{raw})$ on the node hidden feature $\mathbf{H}$ of an undirected graph $\mathcal{G}$ is defined as follows:
\begin{align}\begin{small}
    \label{eq:energy}
    E(\mathbf{H}, \mathbf{A}^{raw})=\frac{1}{N}\sum_{i \in \mathcal{V}}\sum_{j \in \mathcal{N}_i} \mathbf{A}^{raw}_{[i,j]}||\mathbf{H}_i - \mathbf{H}_j ||^2,
\end{small}\end{align} where $\mathbf{H}_i, \mathbf{H}_j$ mean $i$-th and $j$-th rows, respectively.

The oversmoothing phenomenon means that as the depth increases, all node features converge to constants. Thus, $E(\mathbf{H}, \mathbf{A})$ decays to zero asymptotically in time. We will show, via the evolution of the Dirichlet energy, that our proposed method mitigates the oversmoothing problem.

\begin{figure}[t]
    \centering
    \includegraphics[width=0.88\columnwidth]{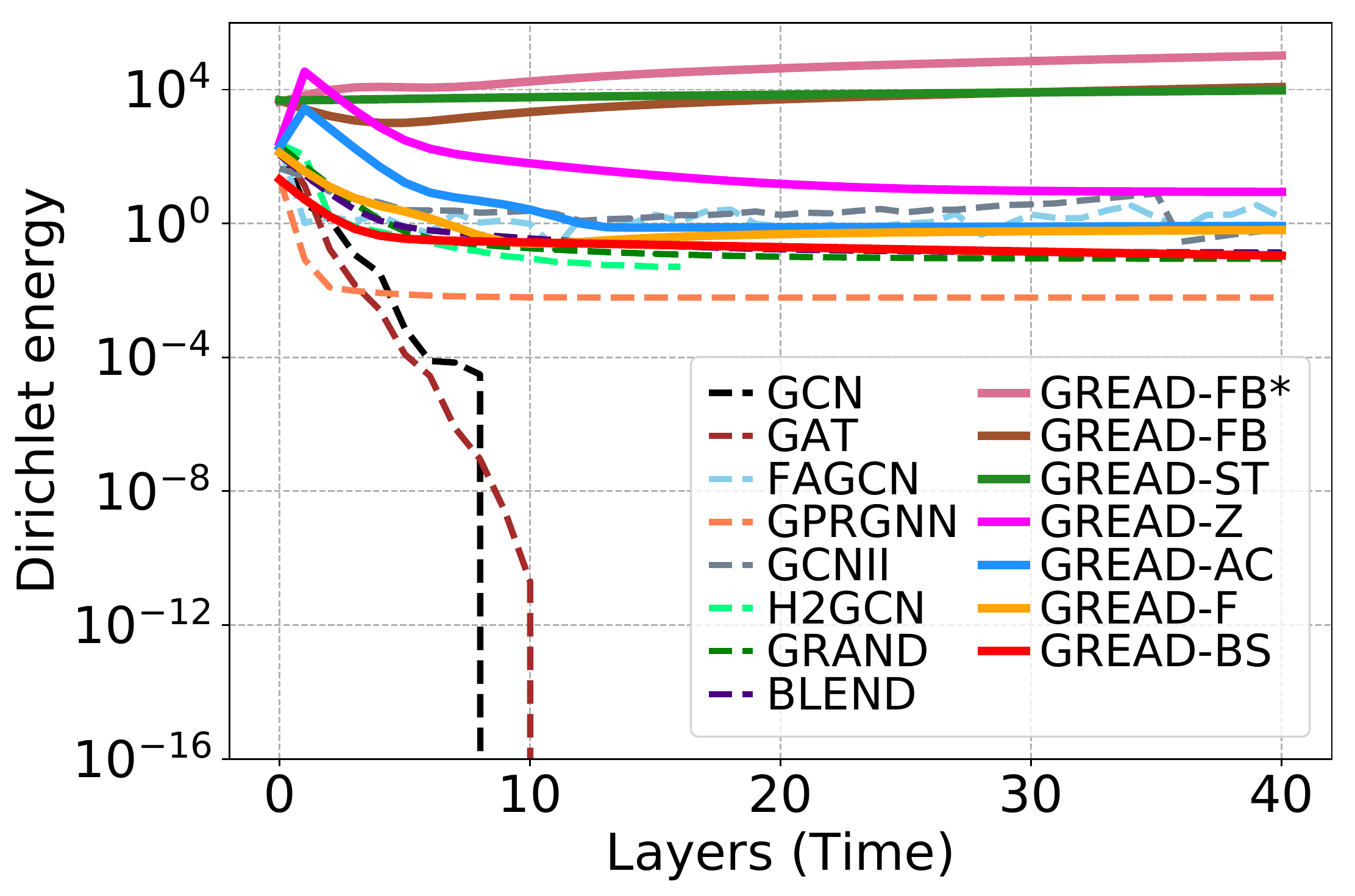}
    \caption{Evolution of the Dirichlet energy on the synthetic random graph. The Y-axis is the logarithmic Dirichlet energy in each layer’s output given a GNN of 40 layers.}
    \label{fig:energy}
\end{figure}

\paragraph{Experimental Environments} We use the synthetic dataset, called cSBMs~\cite{Deshpande2018cSBM}, to demonstrate the mitigation of oversmoothing. This synthetic data is an undirected graph representing 100 nodes in a two-dimensional space with two classes randomly connected with a probability of $p=0.9$. We report the layer-wise Dirichlet energy given a GNN of 40 layers.

\paragraph{Experimental Results} Fig.~\ref{fig:energy} demonstrates traditional GNNs, such as GCN, and GAT, suffer from oversmoothing because the Dirichlet energy decays exponentially to zero in the first five layers. Converging to zero indicates that the node features become constant, while GREAD has no such behaviors. The Dirichlet energy of GREAD can be bounded in time thanks to the reaction term. GRAND only has a diffusion term with learned diffusivity, so that it can delay the oversmoothing. In the case of H2GCN, it is impossible to report on deeper layers due to memory limitations.

\begin{figure}[t]
    \centering
    \subfigure[GREAD-BS vs. baselines]{\includegraphics[width=0.49\columnwidth]{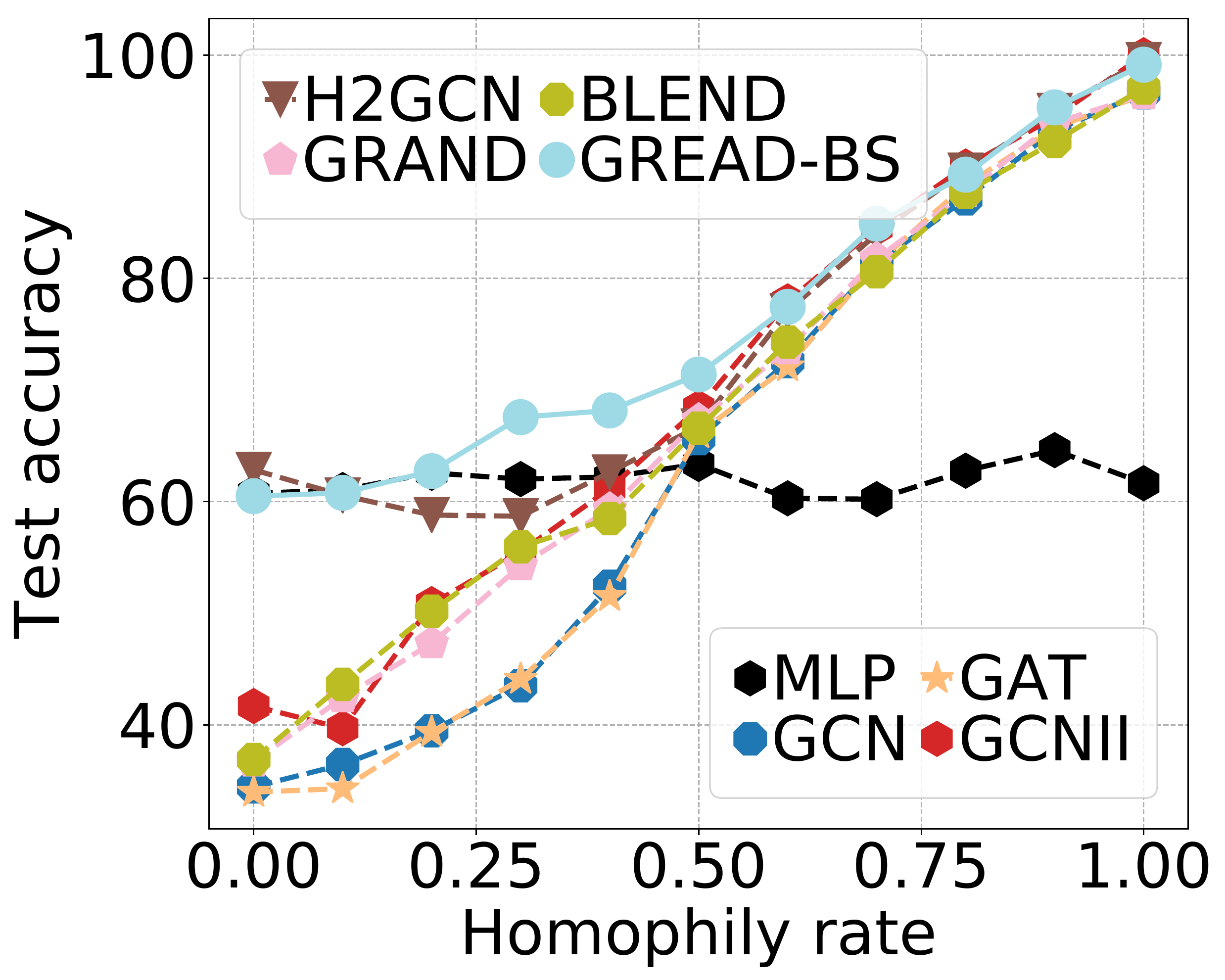}}
    \subfigure[GREAD]{\includegraphics[width=0.49\columnwidth]{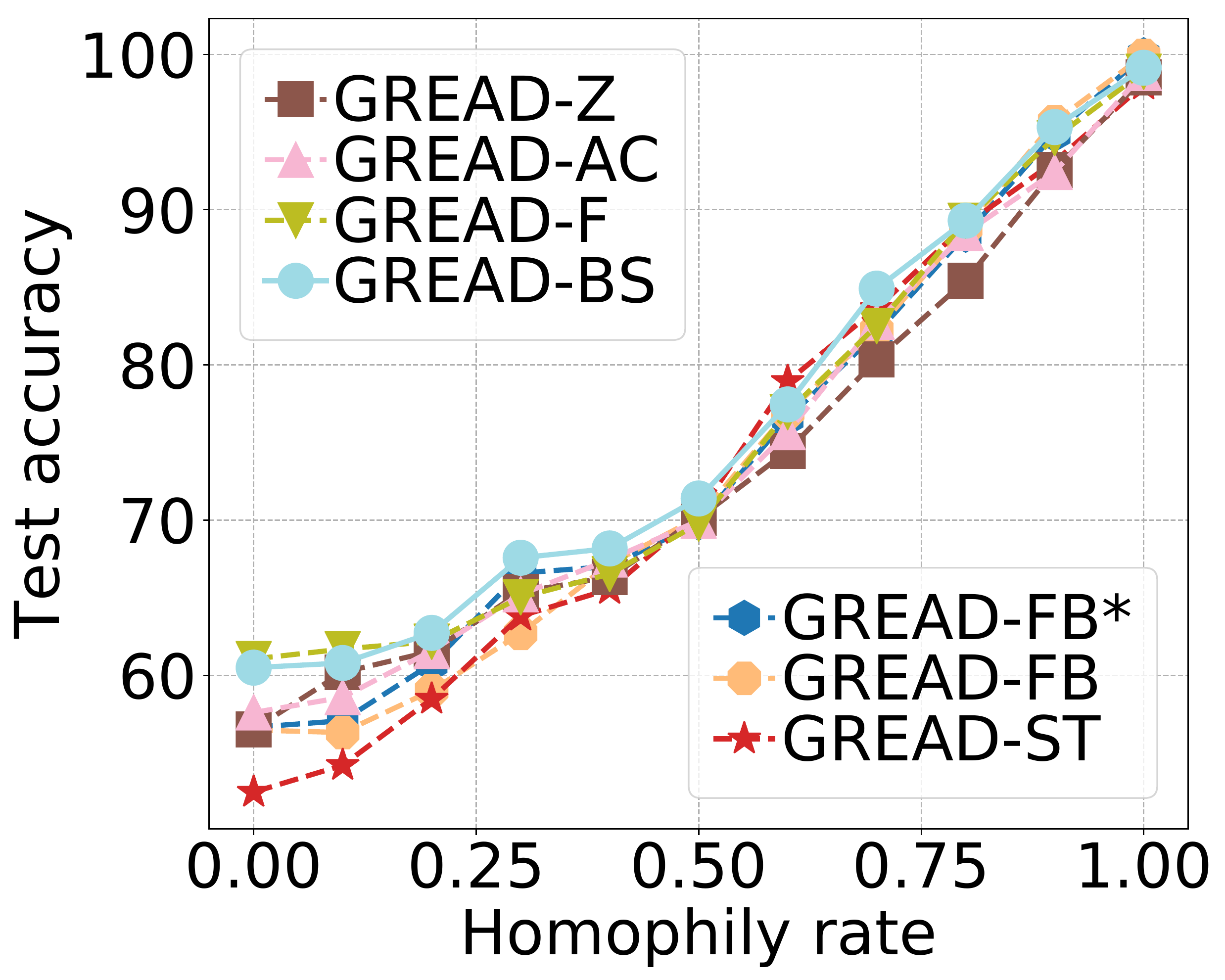}}
    \caption{Experiments on the synthetic Cora with controlled homophily rates.}
    \label{fig:syn_cora}
\end{figure}

\subsection{Different Homophily Levels}
\paragraph{Experimental Environments} In order to test the classification capability of GNNs, we use the synthetic Cora generator~\cite{zhu2020h2gcn,li2021deeprobust}. We generate synthetic graphs with various homophily ratios and report the test accuracy.

\paragraph{Experimental Results} Fig.~\ref{fig:syn_cora} shows the mean test accuracy on all random splits of the synthetic Cora datasets. MLP, which does not consider the connectivity of nodes, maintains its test accuracy for all homophily rates, which is obvious. GCN, GAT, and GRAND, which consider only diffusion, perform poorly at low homophily settings. H2GCN shows reasonable performance on low homophily rates, but its accuracy suddenly decreases at some homophily settings. All GREAD models have the best trend overall without sudden drops. The reaction terms of GREAD contribute to their stable accuracy for both homophily and heterophily settings compared with other models that rely on only diffusion processes, such as GCN and GRAND.

\section{Conclusions}
We presented the concept of graph neural reaction-diffusion equation, called GREAD. Our proposed GREAD is one of the most generalized architectures considering both the diffusion and reaction processes. We design a reaction-diffusion layer that has three types of reaction equations widely used in natural sciences. We also add four reaction terms, including one special reaction term called Blurring-sharpening (BS) designed by us for GNNs. Therefore, our reaction-diffusion layer has seven types. We consider a comprehensive set of 9 real-world datasets with various homophily difficulties and 28 baselines. GREAD marks the best accuracy in almost all cases. In our experiments with the two kinds of synthetic datasets, GREAD shows that it alleviates the oversmoothing problem and performs well on various homophily rates. This shows that our proposed model is a novel framework for constructing GNNs using the concept of the reaction-diffusion equation.

\section*{Acknowledgement}
Noseong Park is the corresponding author. This work was supported by an IITP grant funded by the Korean government (MSIT) (No.2020-0-01361, Artificial Intelligence Graduate School Program (Yonsei University), 10\%) and an ETRI grant funded by the Korean government (23ZS1100, Core Technology Research for Self-Improving Integrated Artificial Intelligence System, 90\%).

\bibliography{reference}
\bibliographystyle{icml2023}

\clearpage
\onecolumn
\appendix

\section{Full Ranking of Table~\ref{tab:summary}}\label{a:full_rank}
We show the average ranking/accuracy and the Olympic ranking of all methods in Table~\ref{tab:summary_full}. Our methods occupy all the top-4 positions.

\begin{table}[ht]
    \centering
    \caption{The average ranking/accuracy and the Olympic ranking in the 9 real-world datasets. Methods are sorted by Avg. Ranking.}
    \begin{tabular}{c cc ccc} \toprule
        \multirow{2}{*}{Method} & \multicolumn{2}{c}{Average} 
                                                           & \multicolumn{3}{c}{Olympic Ranking}\\  \cmidrule(lr){2-3}\cmidrule(lr){4-6}
                 & Ranking & Accuracy                      & Gold & Silver & Bronze\\ \midrule
        \textbf{GREAD-BS} 
                 & 1.56   & 76.64 & 5 & 4 & 0\\
        \textbf{GREAD-FB*}
                 & 6.72   & 74.51 & 0 & 0 & 3\\
        \textbf{GREAD-F}  
                 & 7.50   & 74.13 & 1 & 1 & 1\\
        \textbf{GREAD-FB} 
                 & 7.89   & 74.48 & 0 & 0 & 1\\
        GloGNN   & 8.17   & 74.99 & 0 & 0 & 1\\
        \textbf{GREAD-AC} 
                 & 8.50   & 73.71 & 1 & 0 & 0\\ 
        ACM-GCN  & 8.67   & 74.92 & 0 & 1 & 0\\
        GGCN     & 9.50   & 75.05 & 0 & 1 & 0\\
        \textbf{GREAD-Z}  
                 & 9.56   & 73.45 & 0 & 0 & 0\\
        \textbf{GREAD-ST} 
                 & 9.83   & 72.93 & 0 & 1 & 2\\
        Sheaf	 & 10.33  & 75.06 & 1 & 1 & 0\\
        GRAFF	 & 11.28  & 74.92 & 0 & 0 & 2\\
        WRGAT	 & 13.61  & 73.02 & 0 & 0 & 0\\
        GCNII	 & 14.56  & 70.16 & 0 & 0 & 0\\
        H2GCN	 & 16.00  & 71.33 & 0 & 0 & 0\\
        BLEND	 & 16.11  & 71.79 & 0 & 0 & 0\\
        ACMP     & 19.44  & 71.98 & 0 & 0 & 0\\
        GCON-GCN & 20.17  & 69.46 & 0 & 0 & 1\\
        LINKX	 & 21.44  & 71.11 & 1 & 0 & 0\\
        FA-GCN	 & 21.89  & 69.86 & 0 & 0 & 0\\
        GRAND    & 22.28  & 68.94 & 0 & 0 & 0\\
        GCON-GAT & 22.89  & 68.96 & 0 & 0 & 0\\
        GPR-GNN	 & 23.17  & 67.45 & 0 & 0 & 0\\
        GraphSAGE& 23.56  & 69.50 & 0 & 0 & 0\\
        Geom-GCN & 23.83  & 63.87 & 0 & 0 & 0\\
        Mixhop	 & 24.50  & 68.10 & 0 & 0 & 0\\
        GCN      & 24.61  & 62.77 & 0 & 0 & 0\\
        GDE      & 24.67  & 67.40 & 0 & 0 & 0\\
        MLP      & 25.61  & 66.26 & 0 & 0 & 0\\
        CGNN     & 26.11  & 63.96 & 0 & 0 & 0\\
        ChebNet  & 26.56  & 67.98 & 0 & 0 & 0\\
        GAT      & 28.22  & 60.23 & 0 & 0 & 0\\
        PairNorm & 28.56  & 61.68 & 0 & 0 & 0\\
        JKNet    & 30.06  & 60.77 & 0 & 0 & 0\\
        SGC      & 32.56  & 58.34 & 0 & 0 & 0\\
        \bottomrule
    \end{tabular}
    \label{tab:summary_full}
\end{table}

\clearpage

\section{Full Result of Table~\ref{tab:result}}\label{a:full}
We reported only the selected highly-performing 12 baselines in Table~\ref{tab:result}. We now report all the detailed results of the tested 28 baselines in Table~\ref{tab:result_full}.

\begin{table}[ht]
    \small
    \centering
    \setlength{\tabcolsep}{2pt}
    \caption{Results on real-world datasets: mean $\pm$ std. dev. accuracy for 10 different data splits. We show the best three methods in \BEST{red} (first), \SECOND{blue} (second), and \THIRD{purple} (third).}
    \begin{tabular}{c ccccccccc}\toprule
        Dataset     & Texas      & Wisconsin  & Cornell    & Film       & Squirrel   & Chameleon  & Cora       & Citeseer   & PubMed\\ \midrule
        MLP	        & 80.81\std{±4.75} & 85.29\std{±3.31} & 81.89\std{±6.40} & 36.53\std{±0.70} & 28.77\std{±1.56} & 46.21\std{±2.99} & 87.16\std{±0.37} & 74.02\std{±1.90} & 75.69\std{±2.00}\\
        \midrule
        GCN	        & 55.14\std{±5.16} & 51.76\std{±3.06} & 60.54\std{±5.30} & 27.32\std{±1.10} & 53.43\std{±2.01} & 64.82\std{±2.24} & 86.98\std{±1.27} & 76.50\std{±1.36} & 88.42\std{±0.50}\\
        ChebNet     & 78.37\std{±6.04} & 79.02\std{±3.18} & 75.68\std{±6.94} & 34.13\std{±1.09}	& 36.43\std{±1.17} & 58.64\std{±1.64} & 85.45\std{±1.58} & 75.07\std{±1.25}	& 89.00\std{±0.46}\\
        GAT	        & 52.16\std{±6.63} & 49.41\std{±4.09} & 61.89\std{±5.05} & 27.44\std{±0.89} & 40.72\std{±1.55} & 60.26\std{±2.50} & 86.33\std{±0.48} & 76.55\std{±1.23} & 87.30\std{±1.10}\\
        GraphSAGE	& 82.43\std{±6.14} & 81.18\std{±5.56} & 75.95\std{±5.01} & 34.23\std{±0.99} & 41.61\std{±0.74} & 58.73\std{±1.68} & 86.90\std{±1.04} & 76.04\std{±1.30} & 88.45\std{±0.50}\\
        SGC         & 58.10\std{±4.20} & 55.29\std{±4.28} & 60.00\std{±3.59} & 27.20\std{±1.52} & 33.00\std{±1.97} & 42.45\std{±3.82} & 86.12\std{±1.44} & 76.01\std{±1.31} & 86.90\std{±1.32}\\
        \midrule
        MixHop      & 77.84\std{±7.73} & 75.88\std{±4.90} & 73.51\std{±6.34} & 32.22\std{±2.34} & 43.80\std{±1.48} & 60.50\std{±2.53} & 87.61\std{±0.85} & 76.26\std{±1.33}	& 85.31\std{±0.61}\\
        Geom-GCN	& 66.76\std{±2.72} & 64.51\std{±3.66} & 60.54\std{±3.67} & 31.59\std{±1.15} & 38.15\std{±0.92} & 60.00\std{±2.81} & 85.35\std{±1.57} & \BEST{78.02\std{±1.15}} & 89.95\std{±0.47}\\
        FA-GCN      & 82.43\std{±6.89} & 82.94\std{±7.95} & 79.19\std{±9.79} & 34.87\std{±1.25} & 42.59\std{±0.79} & 55.22\std{±3.19} & 87.21\std{±1.43} & 76.87\std{±1.56}	& 87.45\std{±0.61}\\
        GPR-GNN	    & 78.38\std{±4.36} & 82.94\std{±4.21} & 80.27\std{±8.11} & 34.63\std{±1.22} & 31.61\std{±1.24} & 46.58\std{±1.84} & 87.95\std{±1.18} & 77.13\std{±1.67} & 87.54\std{±0.38}\\
        H2GCN	    & 84.86\std{±7.23} & 87.65\std{±4.98} & 82.70\std{±5.28} & 35.70\std{±1.00} & 36.48\std{±1.86} & 60.11\std{±2.15} & 87.87\std{±1.20} & 77.11\std{±1.57} & 89.49\std{±0.38}\\
        WRGAT       & 83.62\std{±5.50} & {86.98\std{±3.78}} & 81.62\std{±3.90} & 36.53\std{±0.77} & 48.85\std{±0.78} & 65.24\std{±0.87} & 88.20\std{±2.26} & 76.81\std{±1.89} & 89.29\std{±0.38}\\
        GGCN        & 84.86\std{±4.55} & 86.86\std{±3.29} & 85.68\std{±6.63} & 37.54\std{±1.56} & 55.17\std{±1.58} & \SECOND{71.14\std{±1.84}} & 87.95\std{±1.05} & 77.14\std{±1.45} & 89.15\std{±0.37}\\
        LINKX       & 74.60\std{±8.37} & 75.49\std{±5.72} & 77.84\std{±5.81} & 36.10\std{±1.55} & \BEST{61.81\std{±1.80}} & 68.42\std{±1.38} & 84.64\std{±1.13} & 73.19\std{±0.99} & 87.86\std{±0.77}\\
        GloGNN      & 84.32\std{±4.15} & 87.06\std{±3.53} & 83.51\std{±4.26} & 37.35\std{±1.30} & \THIRD{57.54\std{±1.39}} & 69.78\std{±2.42} & 88.31\std{±1.13} & 77.41\std{±1.65} & 89.62\std{±0.35}\\
        ACM-GCN	    & 87.84\std{±4.40} & \SECOND{88.43\std{±3.22}} & 85.14\std{±6.07} & 36.28\std{±1.09} & 54.40\std{±1.88} & 66.93\std{±1.85} & 87.91\std{±0.95} & 77.32\std{±1.70} & 90.00\std{±0.52}\\
        \midrule
        PairNorm	& 60.27\std{±4.34} & 48.43\std{±6.14} & 58.92\std{±3.15} & 27.40\std{±1.24} & 50.44\std{±2.04} & 62.74\std{±2.82} & 85.79\std{±1.01} & 73.59\std{±1.47} & 87.53\std{±0.44}\\
        JKNet   	& 62.70\std{±8.34} & 53.14\std{±5.22} & 59.72\std{±4.60} & 29.25\std{±1.37} & 39.78\std{±1.72} & 52.63\std{±3.90} & 86.48\std{±1.04} & 75.99\std{±1.28} & 87.23\std{±0.55}\\
        GCNII	    & 77.57\std{±3.83} & 80.39\std{±3.40} & 77.86\std{±3.79} & 37.44\std{±1.30} & 38.47\std{±1.58} & 63.86\std{±3.04} & \THIRD{88.37\std{±1.25}} & 77.33\std{±1.48} & \SECOND{90.15\std{±0.43}}\\
        GCON-GCN    & 85.40\std{±4.20} & 87.80\std{±3.30} & 84.30\std{±4.80} & 34.65\std{±0.61} & 33.30\std{±1.57} & 48.08\std{±2.16} & 87.40\std{±1.82} & 76.46\std{±1.70} & 87.71\std{±0.35}\\
        GCON-GAT    & 82.20\std{±4.70} & 85.70\std{±3.60} & 83.20\std{±7.00} & 35.85\std{±0.84} & 34.45\std{±1.08} & 48.31\std{±1.53} & 86.96\std{±1.73} & 76.20±\std{2.12} & 87.73\std{±0.41}\\
        \midrule
        CGNN	    & 71.35\std{±4.05} & 74.31\std{±7.26} & 66.22\std{±7.69} & 35.95\std{±0.86} & 29.24\std{±1.09} & 46.89\std{±1.66} & 87.10\std{±1.35} & 76.91\std{±1.81} & 87.70\std{±0.49}\\
        GDE 	    & 74.05\std{±6.96} & 79.80\std{±5.62} & 82.43\std{±7.07} & 35.36\std{±1.31} & 35.94\std{±1.91} & 47.76\std{±2.08} & 87.22\std{±1.41} & 76.21\std{±2.11} & 87.80\std{±0.38}\\
        GRAND       & 75.68\std{±7.25} & 79.41\std{±3.64} & 82.16\std{±7.09} & 35.62\std{±1.01} & 40.05\std{±1.50} & 54.67\std{±2.54} & 87.36\std{±0.96} & 76.46\std{±1.77} & 89.02\std{±0.51}\\
        BLEND       & 83.24\std{±4.65} & 84.12\std{±3.56} & 85.95\std{±6.82} & 35.63\std{±1.01} & 43.06\std{±1.39} & 60.11\std{±2.09} & 88.09\std{±1.22} & 76.63\std{±1.60} & 89.24\std{±0.42}\\
        ACMP        & 86.20\std{±0.30} & 86.10\std{±0.40} & 85.40\std{±0.70} & 34.44\std{±4.44} & 52.65\std{±2.23} & 52.63\std{±2.28} & 86.38\std{±3.79} & 76.52±\std{1.84} & 87.54\std{±0.57}\\
        Sheaf       & 85.05\std{±5.51} & \BEST{89.41\std{±4.74}} & 84.86\std{±4.71} & \SECOND{37.81\std{±1.15}} & 56.34\std{±1.32} & 68.04\std{±1.58} & 86.90\std{±1.13} & 76.70±\std{1.57} & 89.49\std{±0.40}\\
        GRAFF       & \THIRD{88.38\std{±4.53}} & 87.45\std{±2.94} & 83.24\std{±6.49} & 36.09\std{±0.81} & 54.52\std{±1.37} & \THIRD{71.08\std{±1.75}} & 87.61\std{±0.97} & 76.92±\std{1.70} & 88.95\std{±0.52}\\
        \midrule
        \textbf{GREAD-BS}
                    & \SECOND{88.92\std{±3.72}} & \BEST{89.41\std{±3.30}} & \SECOND{86.49\std{±7.15}} 
                    & \BEST{37.90\std{±1.17}} & \SECOND{59.22\std{±1.44}} & \BEST{71.38\std{±1.53}} 
                    & \BEST{88.57\std{±0.66}} & \SECOND{77.60\std{±1.81}} & \BEST{90.23\std{±0.55}}\\
        \textbf{GREAD-F} 
                    & \BEST{89.73\std{±4.49}} & 86.47\std{±4.84} & \SECOND{86.49\std{±5.13}}
                    & 36.72\std{±0.66}	& 46.16\std{±1.44} & 65.20\std{±1.40} 
                    & 88.39\std{±0.91} & 77.40\std{±1.54}	& 90.09\std{±0.31}\\
        \textbf{GREAD-AC}
                    & 85.95\std{±2.65} & 86.08\std{±3.56} & \BEST{87.03\std{±4.95}} 
                    & 37.21\std{±1.10} & 45.10\std{±2.11} & 65.09\std{±1.08} 
                    & 88.29\std{±0.67} & 77.38\std{±1.53} & 90.10\std{±0.36}\\
        \textbf{GREAD-Z} 
                    & 87.30\std{±5.68} & 86.29\std{±4.32} & 85.68\std{±5.41} 
                    & 37.01\std{±1.11} & 46.25\std{±1.72} & 62.70\std{±2.30} 
                    & 88.31\std{±1.10} & 77.39\std{±1.90} & 90.11\std{±0.27}\\
        \textbf{GREAD-ST} 
                        & 81.08\std{±5.67} & 86.67\std{±3.01} & \THIRD{86.22\std{±5.98}}
                        & 37.66\std{±0.90} & 45.83\std{±1.40} & 63.03\std{±1.32} 
                        & \SECOND{88.47\std{±1.19}} & 77.25\std{±1.47} & \THIRD{90.13\std{±0.36}}\\
        \textbf{GREAD-FB}   
                        & 86.76\std{±5.05} & 87.65\std{±3.17} & \THIRD{86.22\std{±5.85}} 
                        & 37.40\std{±0.55} & 50.83\std{±2.27} & 66.05\std{±1.21} 
                        & 88.01\std{±1.34} & 77.28\std{±1.73} & 90.07\std{±0.45}\\
        \textbf{GREAD-FB*}   
                        & 87.03\std{±3.97} & \THIRD{88.04\std{±1.63}} & 85.95\std{±5.64} 
                        & \THIRD{37.70\std{±0.51}} & 50.57\std{±1.52} & 65.83\std{±1.10} 
                        & 88.01\std{±0.80} & \THIRD{77.42\std{±1.93}} &  90.08\std{±0.46}\\
        \bottomrule
    \end{tabular}
    \label{tab:result_full}
\end{table}

\pagebreak

\section{Full Derivation of Eq.~\eqref{eq:bs}}\label{a:bs}
We show the omitted intermediate derivation steps of Eq.~\eqref{eq:bs}.

\begin{align*}\begin{split}
    \mathbf{H}(t+h) &= \mathbf{B}(t+h) + \tilde{\mathbf{L}}(\mathbf{B}(t+h)),\\
        &\Rightarrow \tilde{\mathbf{A}}\mathbf{H}(t) + \tilde{\mathbf{L}}(\tilde{\mathbf{A}}\mathbf{H}(t)),\\
            &\Rightarrow \tilde{\mathbf{A}}\mathbf{H}(t) + (\mathbf{I}-\tilde{\mathbf{A}})\tilde{\mathbf{A}}\mathbf{H}(t),\\
            &\Rightarrow 2\tilde{\mathbf{A}}\mathbf{H}(t) - \tilde{\mathbf{A}}^2\mathbf{H}(t),\\
            &\Rightarrow (2\mathbf{I}-\tilde{\mathbf{A}})\tilde{\mathbf{A}}\mathbf{H}(t),\\
            &\Rightarrow (\mathbf{I}+\tilde{\mathbf{L}})(\mathbf{I}-\tilde{\mathbf{L}})\mathbf{H}(t),\\
            &\Rightarrow \mathbf{H}(t) - \tilde{\mathbf{L}}^2\mathbf{H}(t),\\
            &\Rightarrow \mathbf{H}(t) - (\mathbf{I}-\tilde{\mathbf{A}})^2\mathbf{H}(t),\\
            &\Rightarrow \mathbf{H}(t) - (\mathbf{I}-\tilde{\mathbf{A}})\mathbf{H}(t) + (\tilde{\mathbf{A}}-\tilde{\mathbf{A}}^2)\mathbf{H}(t),\\
            &\Rightarrow \mathbf{H}(t) - \tilde{\mathbf{L}}\mathbf{H}(t) + (\tilde{\mathbf{A}}-\tilde{\mathbf{A}}^2)\mathbf{H}(t).
\end{split}\end{align*}

\section{Computational Complexity}\label{a:comp}
The space complexity of GREAD is dominated by evaluating the soft adjacency matrix in Eq.~\eqref{eq:soft}, which is $\mathcal{O}(|\mathcal{E}|\text{dim}(\mathbf{H}))$, where $|\mathcal{E}|$ is the number of edges and $\text{dim}(\mathbf{H})$ is the size of hidden dimension.

We also analyze the time complexity of the reaction-diffusion layer in Eq.~\eqref{eq:rd}. Our proposed model has different complexity depending on the reaction term $r$ in Eq.~\eqref{eq:reac}.

If we set the adjacency matrix and $\beta$ to OA and SC respectively, the time complexity of the one-step GREAD-BS computation becomes $\mathcal{O}(n_{\tau}(|\mathcal{E}|+|\mathcal{E}_2|)\text{dim}(\mathbf{H}) + |\mathcal{E}|d_{\text{max}})$, where $n_{\tau}$, $|\mathcal{V}|$, and $d_{\text{max}}$ are the number of steps in $[0,T]$, the number of nodes, and the maximum degree of all nodes, respectively. Given that $\mathbf{A}$ is sparse, we can calculate $\mathbf{A}^2$ in $\mathcal{O}(|\mathcal{E}|d_{\text{max}})$ because $d_{\text{max}}$ is equal to the maximum number of non-zeroes in any row of $\mathbf{A}$. The sparse matrix multiplication of $\mathbf{A}^2 \mathbf{H}(t)$ takes $\mathcal{O}(|\mathcal{E}_2|d_{\text{max}})$, where $|\mathcal{E}_2|=\frac{1}{2}\sum_{v\in\mathcal{V}}|\mathcal{N}_2|(v)$.
The computational complexity of the one-step GREAD-F computation is $\mathcal{O}(n_{\tau}(|\mathcal{E}|+\text{dim}(\mathbf{H})^k))\text{dim}(\mathbf{H}))$, where $k=1$. In the case of GREAD-AC and GREAD-Z, their $k$ values are 2 and 3, respectively. The computational complexities of the one-step GREAD-ST, GREAD-FB, and GREAD-FB* are $\mathcal{O}(n_{\tau}|\mathcal{E}|\text{dim}(\mathbf{H}) + |\mathcal{E}|d_{\text{max}})$. 

\section{Additional Details for Experiments}\label{a:details}
\subsection{Details of Datasets}

\paragraph{Real-world Datasets}
For the experiment with real-world datasets in Table~\ref{tab:result}, we consider both the heterophilic and homophilic datasets. They can be distinguished based on the homophily level. We employ the homophily ratio, defined by~\cite{pei2020geomGCN}, to distinguish high or low homophily/heterophily graphs:
\begin{align}
    \text{Homophily ratio}=\frac{1}{|\mathcal{V}|}\sum_{v\in\mathcal{V}}\frac{\sum_{u \in \mathcal{N}_v}(y_u=y_v)}{|\mathcal{N}_v|}.
\end{align}

A high homophily ratio means that neighbors tend to be in an identical class. Some dataset statistics are given in Table~\ref{tab:data}. The 9 real-world datasets we consider are as follows:
\begin{itemize}
    \item Chameleon and Squirrel are subgraphs of web pages in Wikipedia~\cite{benedek2021musae}. The node in Wikipedia graphs represent web pages, the edge mean mutual links between pages, and the node feature corresponds to several informative nouns in the Wikipedia page. All nodes are classified into 5 categories based on the average monthly traffic.
    \item Film is a subgraph of the film-director-actor-writer network~\cite{tang2009social}. Each node corresponds to an actor, an edge between two nodes denotes the co-occurrence relationship in a Wikipedia page, and the node feature corresponds to some keywords in the Wikipedia page. All nodes are classified into 5 categories according to the type of actors.
    \item Cornell, Texas, and Wisconsin are three subsets of the WebKB dataset collected by CMU, having many links between web pages of the universities. In these networks, nodes represent web pages, edges are hyperlinks between them, and node features are the bag-of-words representation of web pages. All nodes are classified into 5 categories: student, project, course, staff, and faculty.
    \item Cora~\cite{mccallum2000automating}, Citeseer~\cite{sen2008collective}, and Pubmed~\cite{yang2016revisiting} are among the most widely used benchmark datasets for the semi-supervised node classification. These are citation networks, where nodes, edges, features, and labels respectively correspond to papers, undirected paper citations, the bag-of-words representations of papers, and the academic topics of papers.
\end{itemize}

\paragraph{The Synthetic Cora Network}
The synthetic Cora dataset is provided by ~\cite{zhu2020h2gcn}. They generate graphs for a target homophily level using a modified preferential attachment process. Nodes, edges, and features are sampled from Cora to create a synthetic graph with a desired homogeneity and feature/label distribution. In Table~\ref{tab:syn_cora}, we summarize the properties of the synthetic Cora networks we used. 

\begin{table}[ht]
    \centering
    \caption{The detailed information of the synthetic Cora. All levels of homophily have the same number of features (1,433), nodes (1,480), edges (5,936), and classes (5).}
    \begin{tabular}{c ccc}\toprule
        Homophily & Avg. Degree & Max. Degree & Min. Degree\\ \midrule
        0.0 & 3.98 & 84.33 & 1.67 \\
        0.1 & 3.98 & 71.33 & 2.00 \\
        0.2 & 3.98 & 73.33 & 1.67 \\
        0.3 & 3.98 & 70.00 & 2.00 \\
        0.4 & 3.98 & 77.67 & 2.00 \\
        0.5 & 3.98 & 76.33 & 2.00 \\
        0.6 & 3.98 & 76.00 & 1.67 \\
        0.7 & 3.98 & 67.67 & 2.00 \\
        0.8 & 3.98 & 58.00 & 1.67 \\
        0.9 & 3.98 & 58.00 & 1.67 \\
        1.0 & 3.98 & 51.00 & 2.00 \\
        \bottomrule
    \end{tabular}
    \label{tab:syn_cora}
\end{table}

\paragraph{The cSBM Synthetic Network}
For Fig.~\ref{fig:energy}, we use cSBM~\cite{Deshpande2018cSBM} to generate synthetic networks. cSBM generates Gaussian random vectors as node features on top of the classical SBM. The synthetic graph has 100 nodes with 2 classes and two-dimensional features sampled from a normal distribution with $\sigma=2$, $\mu_1=-0.5$, and $\mu_2=0.5$. The nodes are randomly connected with a probability of $p=0.9$ if they are in the same class and $p=0.1$ otherwise.

\subsection{Details of Experimental Settings}\label{a:setting}
\paragraph{Evolution of the Dirichlet Energy}
We use the random graphs generated by cSBM to show the capability of GREAD to alleviate oversmoothing. In the case of GREAD, we run without any hyperparamerter search but list the full hyperparameter list we used in Table~\ref{tab:hyperparams_energy}.

\paragraph{Comparison with Various Homophily Rate}
To compare the performance in various homophily rates, we use the synthetic Cora network. We run the experiment with 3 fixed train/valid/test splits and report the mean and the standard deviation of accuracy accordingly. In Table~\ref{tab:search_homophily}, we list the hyperparameter range we consider.

\paragraph{Node Classification on Real-world Datasets}
The following software and hardware environments were used for all experiments: \textsc{Ubuntu} 18.04 LTS, \textsc{Python} 3.9.12,  \textsc{PyTorch} 1.11.0, \textsc{PyTorch Geometric} 2.0.4, \textsc{torchdiffeq} 0.2.3, \textsc{Numpy} 1.22.4, \textsc{Scipy} 1.8.1, \textsc{Matplotlib} 2.2.3, \textsc{CUDA} 11.3, and \textsc{NVIDIA} Driver 465.19, and i9 CPU, and \textsc{NVIDIA RTX 3090}.  We performed 10 repetitions on the train/valid/test splits taken from~\cite{pei2020geomGCN} and strictly followed their evaluation protocol. For all data sets, we used the largest connectivity component (LCC) except for Citeseer. We use the dropout only in the encoder network and the output layer. We refer to the dropout in the encoder as `input dropout' and  the dropout in the output layer as `dropout'. 

We fine-tune our model within the hyperparameter search space in Table~\ref{tab:search_real}. Our hyperparameter search used the method of W\&B Sweeps~\cite{wandb2020} with a standard random search with 500 counts. We introduce the best hyperparameter configuration in Tables~\ref{tab:best_BS} to~\ref{tab:best_Z}.

\begin{table}[ht!]
    \centering
    \caption{Hyperparameter for the cSBM synthetic network}
    \begin{tabular}{cc} \toprule
        Hyperparameters  &  Value \\ \midrule
        epochs           & 100 \\
        adjacency matrix & OA \\
        $\alpha$         & SC \\
        $\beta$          & VC \\
        learning rate    & 0.001\\
        weight decay     & \num{5e-4}\\
        dropout          & 0.0\\
        input dropout    & 0.5\\
        dim($\mathbf{H}$)& 2\\
        step size $\tau$ & 1.0\\
        time $T$         & 40 \\
        ODE solver       & Euler\\
        \bottomrule
    \end{tabular}
    \label{tab:hyperparams_energy}
\end{table}

\begin{table}[ht!]
    \centering
    \caption{Hyperparameter search space for the synthetic Cora network}
    \label{tab:search_homophily}
    \begin{tabular}{cc} \toprule
        Hyperparameters  &  Search Space \\ \midrule
        epochs           & 100 \\
        adjacency matrix & \{OA, SA\}\\
        $\alpha$         & \{SC, VC\}\\\
        $\beta$          & \{SC, VC\}\\
        learning rate    & \{0.001, 0.002, 0.0025, 0.005, 0.01\}\\
        weight decay     & \{0.01, 0.001, 0.0005, 0.0001\}\\
        dropout          & 0.35\\
        input dropout    & 0.5\\
        dim($\mathbf{H}$)& 64\\
        step size $\tau$ & \{0.1, 0.5, 1.0\}\\
        time $T$         & \{1, 2, 3, 4\}\\
        ODE solver       & Euler\\
        \bottomrule
    \end{tabular}
\end{table}

\begin{table}[ht!]
    \centering
    \caption{Hyperparameter search space for real-world datasets}
    \label{tab:search_real}
    \begin{tabular}{cc} \toprule
        Hyperparameters  &  Search Space \\ \midrule
        epochs           & 200\\
        adjacency matrix & \{OA, SA\}\\
        $\alpha$         & \{SC, VC\}\\
        $\beta$          & \{SC, VC\}\\
        learning rate    & [\num{1e-3}, \num{6e-2}]\\
        weight decay     & [\num{0}, \num{3e-2}]\\
        dropout          & [\num{0}, \num{0.6}]\\
        input dropout    & [\num{0}, \num{0.6}]\\
        dim($\mathbf{H}$)& \{32, 64, 128, 256\}\\
        step size $\tau$ & [0.1, 1.5]\\
        time $T$         & [0.1, 6.0]\\
        ODE solver       & \{Euler, RK4, DOPRI5\}\\
        \bottomrule
    \end{tabular}
\end{table}

\begin{table*}[ht!]
    \centering
    \small
    \caption{Best hyperparameters of GREAD-BS}
    \label{tab:best_BS}
    \begin{tabular}{c ccccccccc} \toprule
        Hyperparameters  & Texas  & Wisconsin 
                                           & Cornell& Film   & Squirrel 
                                                                      & Chameleon
                                                                               & Cora   & Citeseer 
                                                                                                 & PubMed\\ \midrule
        adjacency matrix & OA     & SA     & SA     & SA     & OA     & SA     & SA     & SA     & SA    \\
        $\alpha$         & SC     & SC     & VC     & SC     & VC     & VC     & VC     & SC     & VC    \\
        $\beta$          & VC     & VC     & VC     & VC     & VC     & VC     & SC     & SC     & SC    \\
        learning rate    & 0.0100 & 0.0154 & 0.0082 & 0.0079 & 0.0171 & 0.0068 & 0.0105 & 0.0024 & 0.0108\\
        weight decay     & 0.0247 & 0.0090 & 0.0280 & 0.0014 & 0.0000 & 0.0000 & 0.0060 & 0.0146 & 0.0005\\
        input dropout    & 0.47   & 0.54   & 0.49   & 0.42   & 0.52   & 0.68   & 0.53   & 0.50   & 0.36  \\
        dropout          & 0.48   & 0.48   & 0.32   & 0.65   & 0.09   & 0.05   & 0.45   & 0.47   & 0.26  \\
        dim($\mathbf{H}$)& 128    & 256    & 128    & 64     & 256    & 256    & 64     & 128    & 64    \\
        step size $\tau$ & 1.0    & 0.25   & 0.2    & 0.1    & 0.75   & 1.5    & 0.25   & 0.5    & 0.8   \\
        time $T$         & 1.46   & 0.75   & 0.12   & 0.31   & 5.70   & 1.71   & 3.49   & 2.35   & 1.74  \\
        ODE solver       & Euler  & RK4    & RK4    & RK4    & Euler  & Euler  & RK4    & RK4    & RK4  \\
        \bottomrule
    \end{tabular}
\end{table*}

\begin{table*}[ht!]
    \centering
    \small
    \caption{Best hyperparameters of GREAD-F}
    \label{tab:best_F}
    \begin{tabular}{c ccccccccc} \toprule
        Hyperparameters  & Texas  & Wisconsin 
                                           & Cornell& Film   & Squirrel 
                                                                      & Chameleon
                                                                               & Cora   & Citeseer 
                                                                                                 & PubMed\\ \midrule
        adjacency matrix & SA     & SA     & SA     & SA     & SA     & SA     & SA     & SA     & SA    \\
        $\alpha$         & VC     & SC     & VC     & SC     & VC     & SC     & SC     & SC     & VC    \\
        $\beta$          & VC     & VC     & VC     & VC     & VC     & VC     & SC     & VC     & VC    \\
        learning rate    & 0.0113 & 0.0094 & 0.0092 & 0.0068 & 0.0054 & 0.0101 & 0.0048 & 0.0013 & 0.0120\\
        weight decay     & 0.0079 & 0.0057 & 0.0263 & 0.0006 & 0.0011 & 0.0015 & 0.0370 & 0.0041 & 0.0003\\
        input dropout    & 0.46   & 0.41   & 0.46   & 0.48   & 0.48   & 0.50   & 0.50   & 0.50   & 0.36\\
        dropout          & 0.38   & 0.05   & 0.31   & 0.48   & 0.36   & 0.24   & 0.35   & 0.51   & 0.25\\
        dim($\mathbf{H}$)& 256    & 64     & 256    & 128    & 128    & 256    & 32     & 256    & 128\\
        step size $\tau$ & 1.0    & 0.1    & 1.0    & 0.75   & 1.0    & 1.0    & 0.2    & 0.9    & 1\\
        time $T$         & 1.26   & 0.12   & 1.0    & 1.14   & 2.23   & 1.0    & 2.27   & 1.86   & 1.44\\
        ODE solver       & Euler  & RK4  & Euler  & RK4    & Euler  & RK4    & Euler    & RK4    & RK4  \\
        \bottomrule
    \end{tabular}
\end{table*}

\begin{table*}[ht!]
    \centering
    \small
    \caption{Best hyperparameters of GREAD-AC}
    \label{tab:best_AC}
    \begin{tabular}{c ccccccccc} \toprule
        Hyperparameters  & Texas  & Wisconsin 
                                           & Cornell& Film   & Squirrel 
                                                                      & Chameleon
                                                                               & Cora   & Citeseer 
                                                                                                 & PubMed\\ \midrule
        adjacency matrix & SA     & SA     & SA     & SA     & SA     & SA     & SA     & SA     & SA    \\
        $\alpha$         & VC     & VC     & SC     & SC     & SC     & SC     & SC     & SC     & VC    \\
        $\beta$          & VC     & VC     & VC     & SC     & VC     & VC     & VC     & VC     & VC    \\
        learning rate    & 0.0070 & 0.0083 & 0.0084 & 0.0027 & 0.0025 & 0.0038 & 0.0039 & 0.0029 & 0.0124\\
        weight decay     & 0.0136 & 0.0081 & 0.0311 & 0.0001 & 0.0020 & 0.0007 & 0.0469 & 0.0140 & 0.0006\\
        input dropout    & 0.40   & 0.45   & 0.49   & 0.46   & 0.52   & 0.52   & 0.40   & 0.47   & 0.30\\
        dropout          & 0.30   & 0.20   & 0.29   & 0.48   & 0.28   & 0.35   & 0.40   & 0.49   & 0.26\\
        dim($\mathbf{H}$)& 256    & 128    & 128    & 128    & 128    & 256    & 128    & 64     & 128\\
        step size $\tau$ & 1.0    & 0.5    & 0.75   & 1.0    & 1.0    & 1.0    & 0.1    & 0.9    & 1.0\\
        time $T$         & 1.36   & 0.20   & 0.18   & 1.06   & 1.98   & 2.0    & 3.52   & 2.78   & 1.65\\
        ODE solver       & Euler  & RK4    & RK4    & Euler  & Euler  & RK4    & Euler    & RK4    & RK4  \\
        \bottomrule
    \end{tabular}
\end{table*}

\begin{table*}[ht!]
    \centering
    \small
    \caption{Best hyperparameters of GREAD-Z}
    \label{tab:best_Z}
    \begin{tabular}{c ccccccccc} \toprule
        Hyperparameters  & Texas  & Wisconsin 
                                          & Cornell& Film   & Squirrel 
                                                                      & Chameleon
                                                                               & Cora   & Citeseer 
                                                                                                 & PubMed\\ \midrule
        adjacency matrix & OA     & SA     & SA     & SA     & SA     & SA     & SA     & SA     & SA    \\
        $\alpha$         & VC     & SC     & SC     & SC     & VC     & VC     & VC     & VC     & VC    \\
        $\beta$          & SC     & VC     & SC     & VC     & VC     & VC     & SC     & VC     & VC    \\
        learning rate    & 0.0088 & 0.0046 & 0.0048 & 0.0023 & 0.0099 & 0.0111 & 0.0045 & 0.0027 & 0.0091\\
        weight decay     & 0.0462 & 0.0086 & 0.0435 & 0.0011 & 0.0007 & 0.0012 & 0.0050 & 0.0145 & 0.0004\\
        input dropout    & 0.48   & 0.45   & 0.4272 & 0.48   & 0.53   & 0.45   & 0.4    & 0.50   & 0.37\\
        dropout          & 0.46   & 0.18   & 0.29   & 0.48   & 0.44   & 0.31   & 0.2    & 0.49   & 0.22\\
        dim($\mathbf{H}$)& 256    & 128    & 256    & 64     & 128    & 256    & 64     & 64     & 64\\
        step size $\tau$ & 1.2    & 0.4    & 0.2    & 0.2    & 1.0    & 1.0    & 0.1    & 0.8    & 0.8\\
        time $T$         & 1.2    & 0.11   & 0.13   & 0.75   & 2.71   & 1.0    & 3.55   & 2.01   & 1.12\\
        ODE solver       & RK4  & RK4  & RK4  & RK4    & RK4  & RK4    & RK4    & RK4    & RK4  \\
        \bottomrule
    \end{tabular}
\end{table*}

\begin{table*}[ht!]
    \centering
    \small
    \caption{Best hyperparameters of GREAD-ST}
    \label{tab:best_ST}
    \begin{tabular}{c ccccccccc} \toprule
        Hyperparameters  & Texas  & Wisconsin 
                                          & Cornell& Film   & Squirrel 
                                                                      & Chameleon
                                                                               & Cora   & Citeseer 
                                                                                                 & PubMed\\ \midrule
        adjacency matrix & OA     & SA     & SA     & SA     & SA     & SA     & SA     & SA     & SA    \\
        $\alpha$         & SC     & SC     & SC     & SC     & VC     & VC     & SC     & SC     & SC    \\
        $\beta$          & SC     & VC     & VC     & SC     & VC     & VC     & VC     & SC     & SC    \\
        learning rate    & 0.0200 & 0.0180 & 0.0050 & 0.0081 & 0.0538 & 0.0077 & 0.0074 & 0.0038 & 0.0108\\
        weight decay     & 0.0295 & 0.0082 & 0.0275 & 0.0013 & 0.0000 & 0.0000 & 0.0086 & 0.0042 & 0.0004\\
        input dropout    & 0.46   & 0.54   & 0.47   & 0.42   & 0.61   & 0.65   & 0.37   & 0.49   & 0.36  \\
        dropout          & 0.50   & 0.50   & 0.25   & 0.56   & 0.95   & 0.09   & 0.41   & 0.54   & 0.22  \\
        dim($\mathbf{H}$)& 126    & 256    & 256    & 64     & 256    & 256    & 128    & 64     & 64    \\
        step size $\tau$ & 0.5    & 0.5    & 0.25   & 0.7    & 1.0    & 1.0    & 0.1    & 0.6    & 0.9   \\
        time $T$         & 1.02   & 0.1    & 0.20   & 0.15   & 3.54   & 1.0    & 3.04   & 2.37   & 1.28  \\
        ODE solver       & Euler  & RK4    & RK4    & RK4    & Euler  & Euler  & RK4    & RK4    & RK4   \\
        \bottomrule
    \end{tabular}
\end{table*}

\begin{table*}[ht!]
    \centering
    \small
    \caption{Best hyperparameters of GREAD-FB}
    \label{tab:best_FB}
    \begin{tabular}{c ccccccccc} \toprule
        Hyperparameters  & Texas  & Wisconsin 
                                          & Cornell& Film   & Squirrel 
                                                                      & Chameleon
                                                                               & Cora   & Citeseer 
                                                                                                 & PubMed\\ \midrule
        adjacency matrix & OA     & SA     & SA     & SA     & SA     & SA     & SA     & OA     & SA    \\
        $\alpha$         & VC     & SC     & SC     & SC     & VC     & VC     & SC     & VC     & VC    \\
        $\beta$          & SC     & VC     & VC     & VC     & VC     & VC     & VC     & SC     & VC    \\
        learning rate    & 0.0016 & 0.0185 & 0.0050 & 0.0133 & 0.0090 & 0.0010 & 0.0064 & 0.0012 & 0.0102\\
        weight decay     & 0.0055 & 0.0113 & 0.0283 & 0.0014 & 0.0000 & 0.0000 & 0.0091 & 0.0042 & 0.0004\\
        input dropout    & 0.52   & 0.50   & 0.36   & 0.51   & 0.62   & 0.64   & 0.47   & 0.45   & 0.35\\
        dropout          & 0.48   & 0.53   & 0.23   & 0.60   & 0.06   & 0.05   & 0.50   & 0.54   & 0.21\\
        dim($\mathbf{H}$)& 64     & 64     & 256    & 64     & 128    & 128    & 256    & 128    & 64\\
        step size $\tau$ & 1.5    & 0.7    & 1.0    & 0.6    & 0.25   & 0.25   & 0.5    & 0.6    & 0.2\\
        time $T$         & 1.4    & 1.0    & 0.1    & 1.3    & 2.4    & 1.8    & 3.1    & 1.5    & 1.0\\
        ODE solver       & Euler  & RK4    & RK4    & RK4    & Euler  & Euler  & RK4    & RK4    & Euler  \\
        \bottomrule
    \end{tabular}
\end{table*}
\begin{table*}[ht!]
    \centering
    \small
    \caption{Best hyperparameters of GREAD-FB*}
    \label{tab:best_FB3}
    \begin{tabular}{c ccccccccc} \toprule
        Hyperparameters  & Texas  & Wisconsin 
                                          & Cornell& Film   & Squirrel 
                                                                      & Chameleon
                                                                               & Cora   & Citeseer 
                                                                                                 & PubMed\\ \midrule
        adjacency matrix & OA     & SA     & SA     & SA     & SA     & SA     & SA     & SA     & SA    \\
        $\alpha$         & SC     & SC     & SC     & SC     & VC     & VC     & VC     & VC     & SC    \\
        $\beta$          & VC     & VC     & SC     & VC     & VC     & VC     & VC     & SC     & VC    \\
        learning rate    & 0.0194 & 0.0195 & 0.0072 & 0.0144 & 0.0055 & 0.0095 & 0.0097 & 0.0020 & 0.0166\\
        weight decay     & 0.0113 & 0.0142 & 0.0169 & 0.0010 & 0.0000 & 0.0000 & 0.0090 & 0.0048 & 0.0005\\
        input dropout    & 0.45   & 0.51   & 0.36   & 0.52   & 0.63   & 0.65   & 0.50   & 0.57   & 0.35  \\
        dropout          & 0.52   & 0.48   & 0.19   & 0.59   & 0.05   & 0.14   & 0.39   & 0.39   & 0.22  \\
        dim($\mathbf{H}$)& 64     & 64     & 128    & 64     & 128    & 128    & 64     & 64     & 128   \\
        step size $\tau$ & 1.5    & 1.0    & 0.1    & 0.8    & 0.1    & 0.2    & 0.1    & 0.9    & 1.0   \\
        time $T$         & 1.4    & 1.0    & 0.2    & 1.9    & 2.0    & 1.5    & 3.3    & 1.7    & 1.4  \\
        ODE solver       & Euler  & Euler  & Euler  & RK4    & Euler  & Euler  & Euler  & RK4    & RK4   \\
        \bottomrule
    \end{tabular}
\end{table*}

\clearpage

\section{Additional Experimental Results on Real-world Datasets}\label{a:add_exp}

\subsection{Ablation Studies}\label{a:ablation}
Tables~\ref{tab:soft_appendix} to~\ref{tab:beta_appendix} show the results of our additional ablation studies in the remaining datasets that are not reported in our main paper. In Table~\ref{tab:soft_appendix}, SA outperforms OA in all the datasets except for Texas and Wisconsin. In the case of Texas and Wisconsin, SA performs worse than OA from time to time. In Table~\ref{tab:beta_appendix}, we compare two types of $\beta$. $\beta$ can be either a scalar parameter (SC) or a learnable vector parameter (VC). In almost cases, it shows better performance when the type of $\beta$ is VC.

 \begin{table}[ht!]
     \small
     \centering
     \caption{Ablation study on soft adjacency matrix}
     \label{tab:soft_appendix}
     \begin{tabular}{cc cccc ccc}\toprule
        Dataset & $\mathbf{A}$ &  GREAD-BS & GREAD-F & GREAD-AC & GREAD-Z & GREAD-ST & GREAD-FB & GREAD-FB*\\ 
        \midrule \multirow{2}{*}{Texas}
        & OA   &  \textbf{88.92\std{±3.72}} & 86.49\std{±4.69} & 85.41\std{±5.16} & \textbf{87.30\std{±5.68}} & \textbf{81.08\std{±5.67}} & \textbf{86.76\std{±5.05}} & \textbf{87.03\std{±3.97}}\\
        & SA   &  85.41\std{±2.76} & \textbf{89.73\std{±4.49}} & \textbf{85.95\std{±2.65}} & 86.49\std{±3.20} & 80.00\std{±6.23} & 84.41\std{±4.22} & 85.14\std{±5.57}\\
        \midrule \multirow{2}{*}{Wisconsin}
        & OA   &  87.45\std{±3.53} & \textbf{86.47\std{±4.16}} & \textbf{87.26\std{±3.87}} & 86.28\std{±3.62} & 85.88\std{±3.26} & 85.13\std{±4.13} & 85.42\std{±4.51}\\
        & SA   &  \textbf{89.41\std{±3.30}} & \textbf{86.47\std{±4.84}} & 85.69\std{±5.04} & \textbf{86.29\std{±4.32}} & \textbf{86.67\std{±3.01}} & \textbf{87.65\std{±3.17}} & \textbf{88.04\std{±1.63}}\\
        \midrule \multirow{2}{*}{Squirrel}
        & OA   &  47.03\std{±1.31} & 37.85\std{±1.11} & 38.07\std{±1.71} & 38.43\std{±1.37} & 41.56\std{±1.74} & 49.88\std{±1.44} & 49.21\std{±1.95}\\
        & SA   &  \textbf{59.22\std{±1.44}} & \textbf{46.16\std{±1.44}} & \textbf{45.10\std{±2.11}} & \textbf{46.25\std{±1.72}} & \textbf{45.83\std{±1.40}} & \textbf{50.83\std{±2.27}} & \textbf{50.57\std{±1.52}}\\
        \midrule \multirow{2}{*}{Chameleon}
        & OA   &  67.79\std{±1.91} & 59.80\std{±1.54} & 58.93\std{±1.92} & 54.45\std{±2.29} & 56.05\std{±1.28} & 62.41\std{±1.99} & 62.63\std{±1.64}\\
        & SA   &  \textbf{71.38\std{±1.31}} & \textbf{65.20\std{±1.65}} & \textbf{65.09\std{±1.08}} & \textbf{62.70\std{±2.30}} & \textbf{62.30\std{±1.99}} & \textbf{66.05\std{±1.21}} & \textbf{65.83\std{±1.10}}\\
        \midrule \multirow{2}{*}{Cora}
        & OA   &  87.34\std{±1.34} & 86.72\std{±1.17} & 86.88\std{±1.09} & 86.90\std{±1.02} & 87.77\std{±1.35} & 88.01\std{±1.34} & 87.67\std{±1.14}\\
        & SA   &  \textbf{88.57\std{±0.66}} & \textbf{88.39\std{±0.91}} & \textbf{88.29\std{±0.67}} & \textbf{88.31\std{±1.10}} & \textbf{88.47\std{±1.19}} & \textbf{88.03\std{±0.78}} & \textbf{88.01\std{±0.80}}\\
        \midrule \multirow{2}{*}{Citeseer}
        & OA   &  77.33\std{±1.74} & 76.29\std{±1.74} & 76.73\std{±1.52} & 76.69\std{±1.97} & 76.57\std{±1.29} & \textbf{77.28\std{±1.73}} & 77.38\std{±1.79}\\
        & SA   &  \textbf{77.60\std{±1.81}} & \textbf{77.40\std{±1.54}} & \textbf{77.38\std{±1.53}} & \textbf{77.39\std{±1.73}} & \textbf{77.25\std{±1.47}} & 77.09\std{±1.73} & \textbf{77.42\std{±1.93}}\\
        \midrule \multirow{2}{*}{Pubmed}
        & OA   &  89.98\std{±0.38} & 87.99\std{±0.41} & 88.31\std{±0.44} & 88.83\std{±0.37} & 87.87\std{±0.33} & 89.96\std{±0.33} & 89.58\std{±0.39}\\
        & SA   &  \textbf{90.23\std{±0.55}} & \textbf{90.09\std{±0.31}} & \textbf{90.10\std{±0.36}} & \textbf{90.11\std{±0.27}} & \textbf{90.13\std{±0.36}} & \textbf{90.07\std{±0.45}} & \textbf{90.08\std{±0.46}}\\
        \bottomrule
     \end{tabular}
 \end{table}

\begin{table}[h!]
     \small
     \centering
     \caption{Ablation study on $\beta$}
     \label{tab:beta_appendix}
     \begin{tabular}{cc cccc ccc}\toprule
        Dataset & $\beta$ &  GREAD-BS & GREAD-F & GREAD-AC & GREAD-Z & GREAD-ST & GREAD-FB & GREAD-FB*\\
        \midrule \multirow{2}{*}{Cornell}
        & SC    &  85.14\std{±5.57} & 85.41\std{±6.75} & 85.41\std{±6.96} & 84.60\std{±6.17} & 85.95\std{±6.60} & 85.65\std{±6.21} & 84.16\std{±6.02}\\
        & VC    &  \textbf{86.49\std{±7.15}} & \textbf{86.49\std{±5.13}} & \textbf{87.03\std{±4.95}} & \textbf{85.68\std{±5.41}} & \textbf{86.22\std{±5.98}} & \textbf{86.22\std{±5.85}} & \textbf{85.95\std{±5.64}}\\
        \midrule \multirow{2}{*}{Film}
        & SC    &  37.09\std{±1.15} & 36.53\std{±1.04} & \textbf{37.21\std{±1.10}} & \textbf{37.01\std{±1.11}} & \textbf{37.66\std{±0.90}} & 35.07\std{±0.92} & 34.24\std{±1.21}\\
        & VC    &  \textbf{37.90\std{±1.17}} & \textbf{37.20\std{±1.26}} & 36.76\std{±0.99} & 36.70\std{±0.69} & 37.33\std{±1.35} & \textbf{37.40\std{±0.55}} & \textbf{37.70\std{±0.51}}\\
        \midrule \multirow{2}{*}{Squirrel}
        & SC    &  42.74\std{±1.34} & 44.88\std{±1.62} & 39.61\std{±1.69} & 40.33\std{±2.06} & 43.41\std{±1.61} & 40.59\std{±1.14} & 40.15\std{±1.66}\\
        & VC    &  \textbf{59.22\std{±1.44}} & \textbf{46.16\std{±1.44}} & \textbf{45.10\std{±2.11}} & \textbf{46.25\std{±1.72}} & \textbf{45.83\std{±1.40}} & \textbf{50.83\std{±2.27}} & \textbf{50.57\std{±1.52}}\\
        \midrule \multirow{2}{*}{Chameleon}
        & SC    & 62.02\std{±1.86} & 61.80\std{±1.80} & 56.56\std{±2.28} & 59.17\std{±1.26} & 60.70\std{±1.40} & 57.57\std{±1.83} & 57.70\std{±2.11} \\
        & VC    & \textbf{71.38\std{±1.31}} & \textbf{65.20\std{±1.65}} & \textbf{65.09\std{±1.08}} & \textbf{62.70\std{±2.30}} & \textbf{62.30\std{±1.99}} & \textbf{66.05\std{±1.21}} & \textbf{65.83\std{±1.10}} \\
        \midrule \multirow{2}{*}{Cora}
        & SC    &  87.45\std{±1.08} & 88.07\std{±0.96} & 88.01\std{±0.85} & 88.13\std{±0.40} & 88.35\std{±1.32} & 87.75\std{±1.24} & 86.68\std{±0.88} \\
        & VC    &  \textbf{88.57\std{±0.66}} & \textbf{88.39\std{±0.91}} & \textbf{88.29\std{±0.67}} & \textbf{88.31\std{±1.10}} & \textbf{88.47\std{±1.19}} & \textbf{88.03\std{±0.78}} & \textbf{88.01\std{±0.80}} \\
        \midrule \multirow{2}{*}{Citeseer}
        & SC    &  76.73\std{±1.73} & 76.70\std{±1.75} & 75.83\std{±1.36} & 76.83\std{±1.16} & \textbf{77.25\std{±1.47}} & \textbf{77.28\std{±1.73}} & \textbf{77.42\std{±1.93}} \\
        & VC    &  \textbf{77.60\std{±1.81}} & \textbf{77.40\std{±1.54}} & \textbf{77.38\std{±1.53}} & \textbf{77.39\std{±1.73}} & 77.13\std{±2.20} & 77.22\std{±2.13} & 77.23\std{±1.89} \\ 
        \midrule \multirow{2}{*}{Pubmed}
        & SC    &  89.96\std{±0.42} & 87.51\std{±0.44} & 88.76\std{±0.45} & 90.04\std{±0.26} & \textbf{90.13\std{±0.36}} & 89.90\std{±0.47} & 89.99\std{±0.24} \\ 
        & VC    &  \textbf{90.23\std{±0.55}} & \textbf{90.09\std{±0.31}} & \textbf{90.10\std{±0.36}} & \textbf{90.11\std{±0.27}} & 90.10\std{±0.41} & \textbf{90.07\std{±0.45}} & \textbf{90.08\std{±0.46}} \\
        \bottomrule
     \end{tabular}
 \end{table}

\clearpage
 
\subsection{Sensitivity Analyses}\label{a:sensitivity}
\begin{figure}[ht!]
    \centering
    \includegraphics[width=\textwidth]{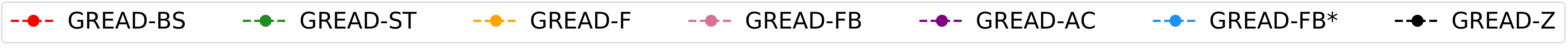}\vspace{0.3in}
    \subfigure[Texas]{\includegraphics[width=0.32\textwidth]{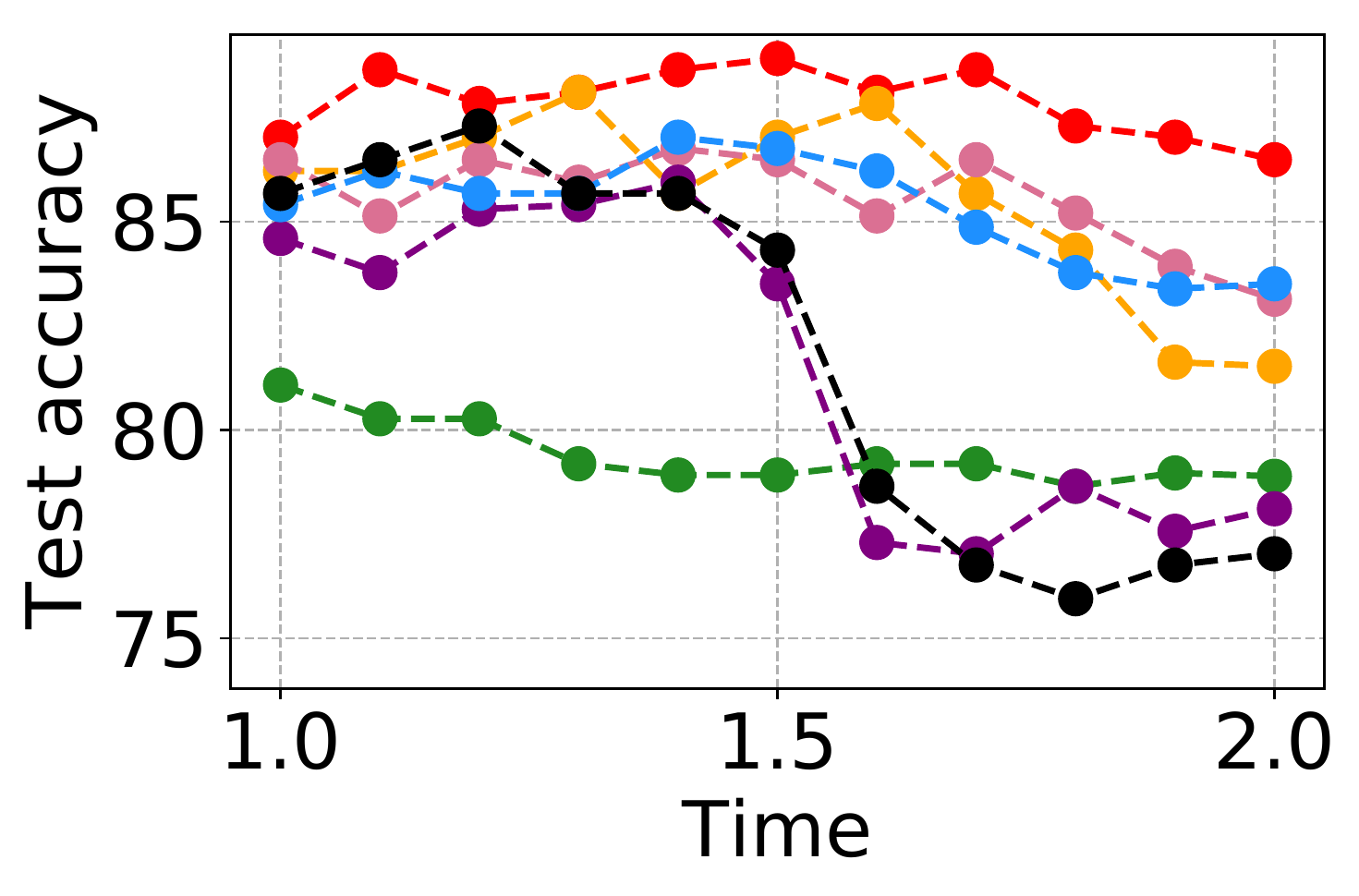}}
    \subfigure[Wisconsin]{\includegraphics[width=0.32\textwidth]{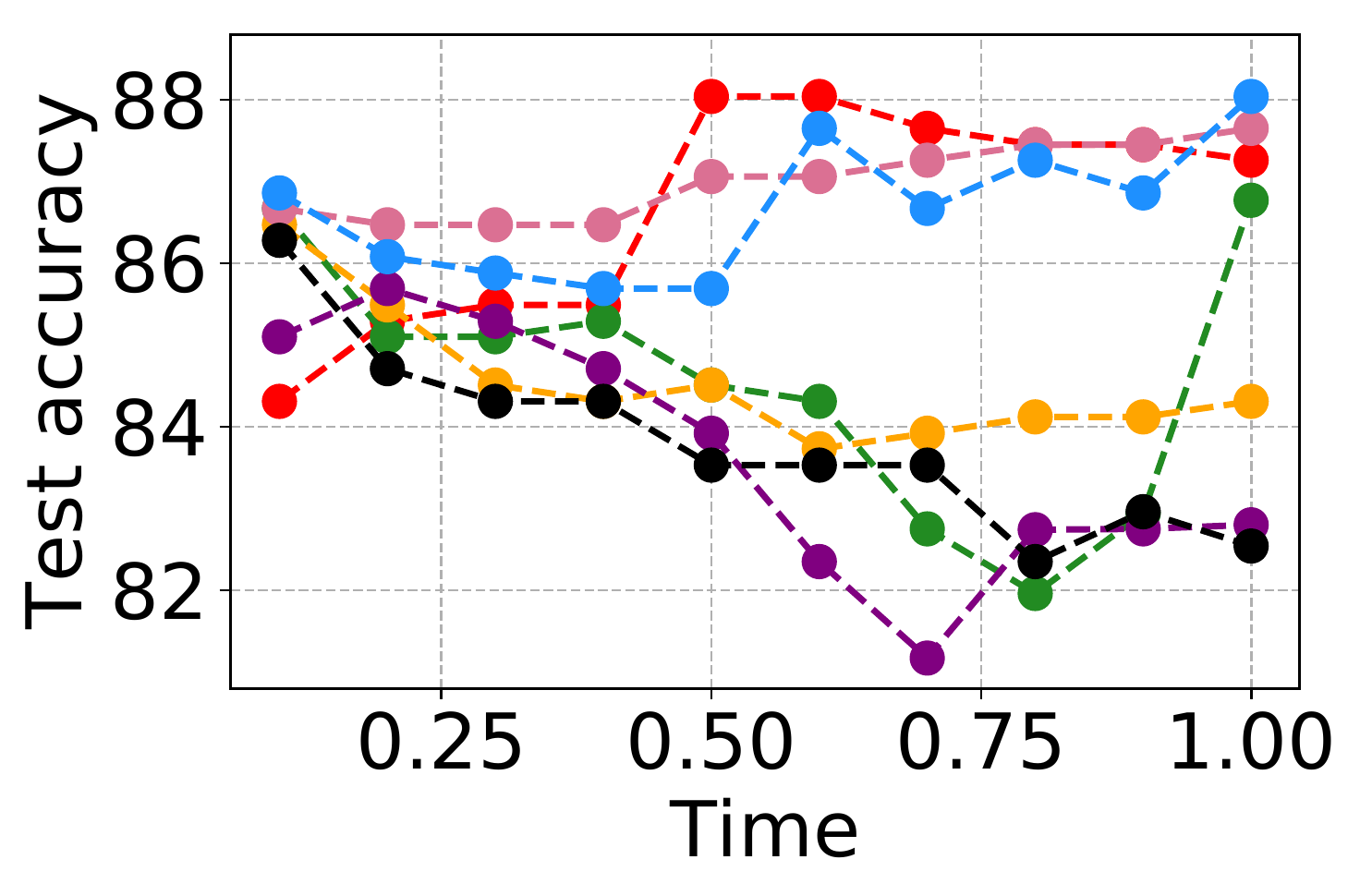}}
    \subfigure[Cornell]{\includegraphics[width=0.32\textwidth]{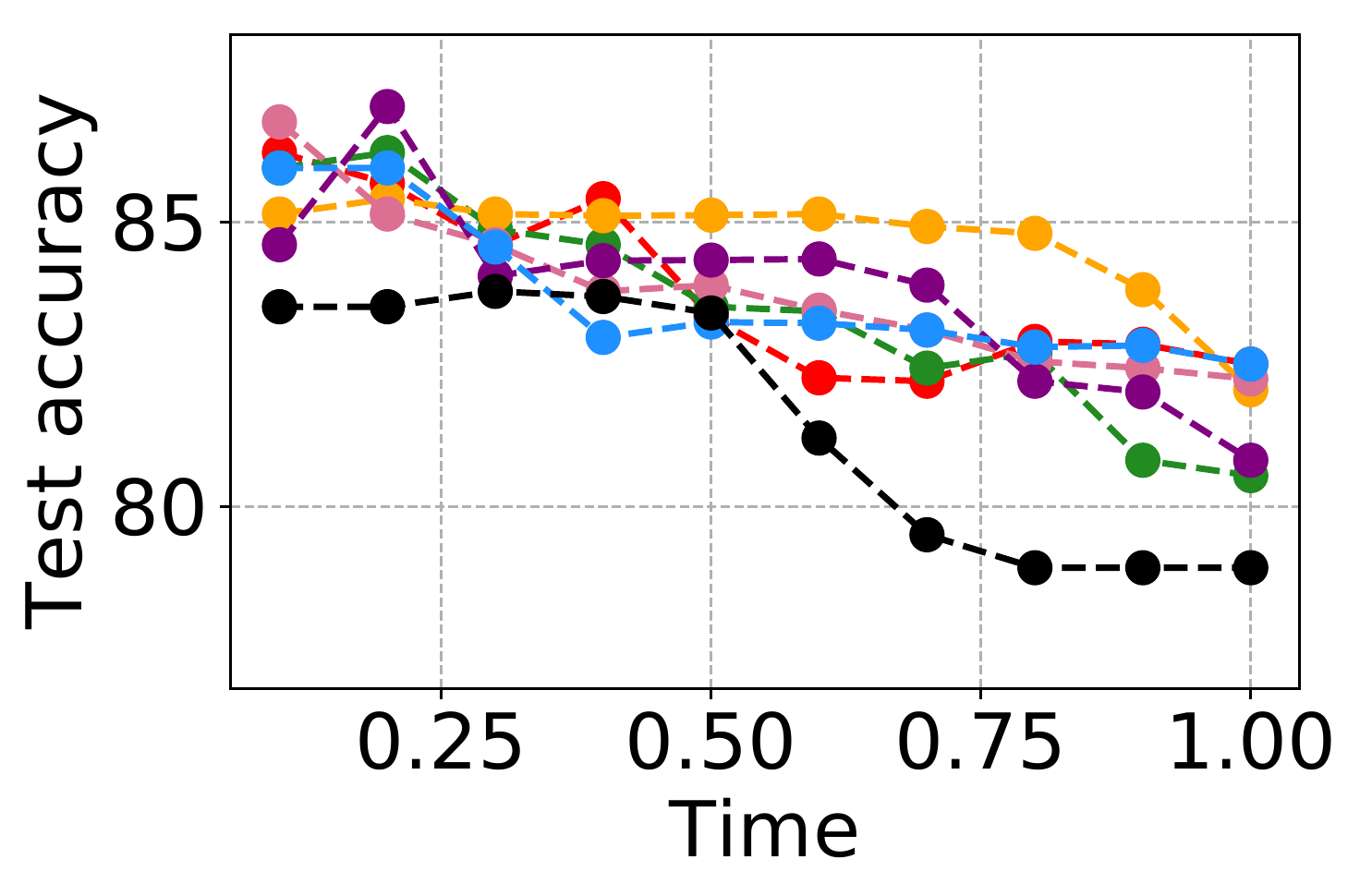}}
    \subfigure[Film]{\includegraphics[width=0.32\textwidth]{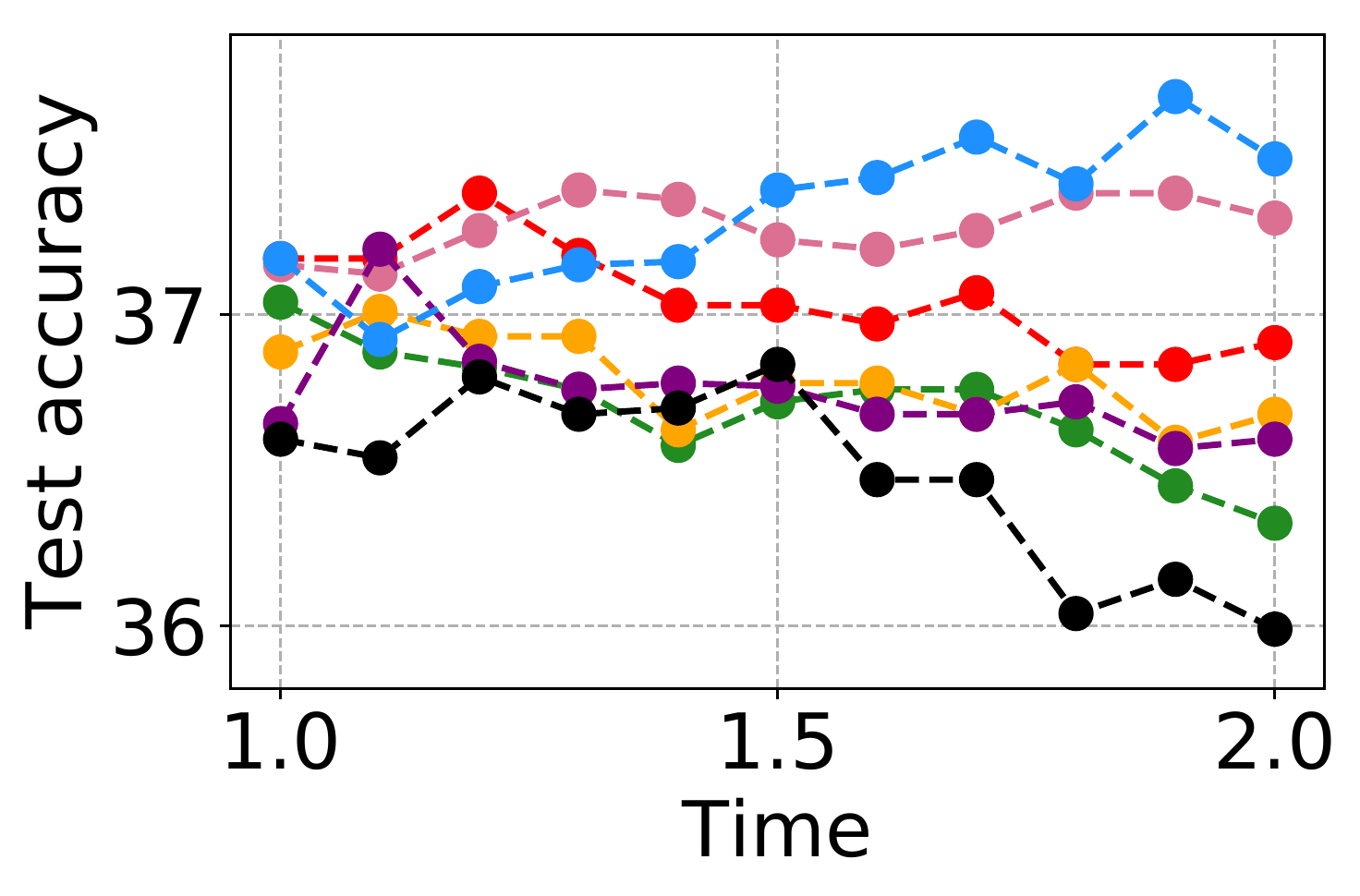}}
    \subfigure[Squirrel]{\includegraphics[width=0.32\textwidth]{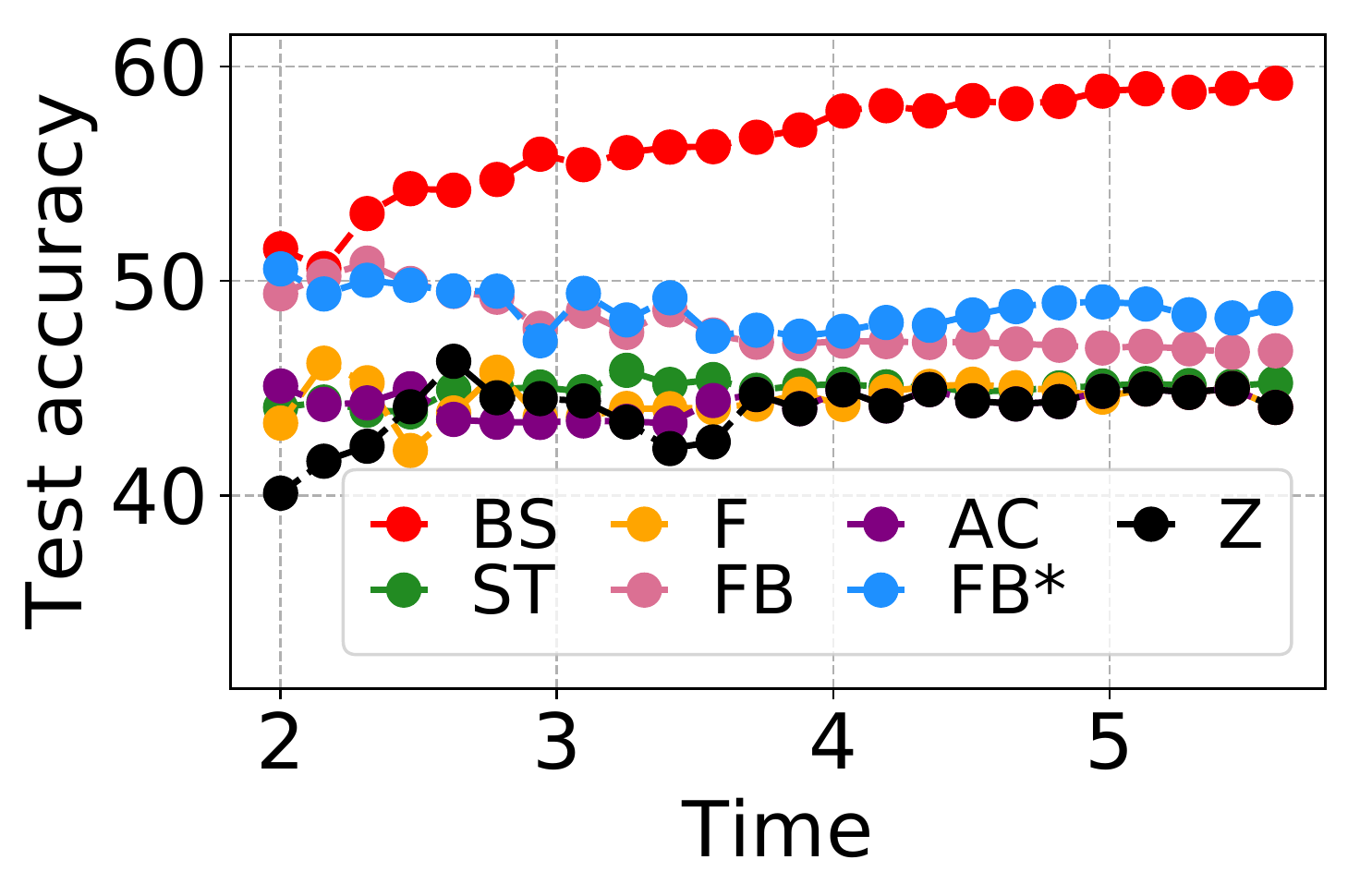}}
    \subfigure[Chameleon]{\includegraphics[width=0.32\textwidth]{img/ablation-cm/chameleon_T.pdf}}
    \subfigure[Cora]{\includegraphics[width=0.32\textwidth]{img/ablation-cm/cora_T.pdf}}
    \subfigure[Citeseer]{\includegraphics[width=0.32\textwidth]{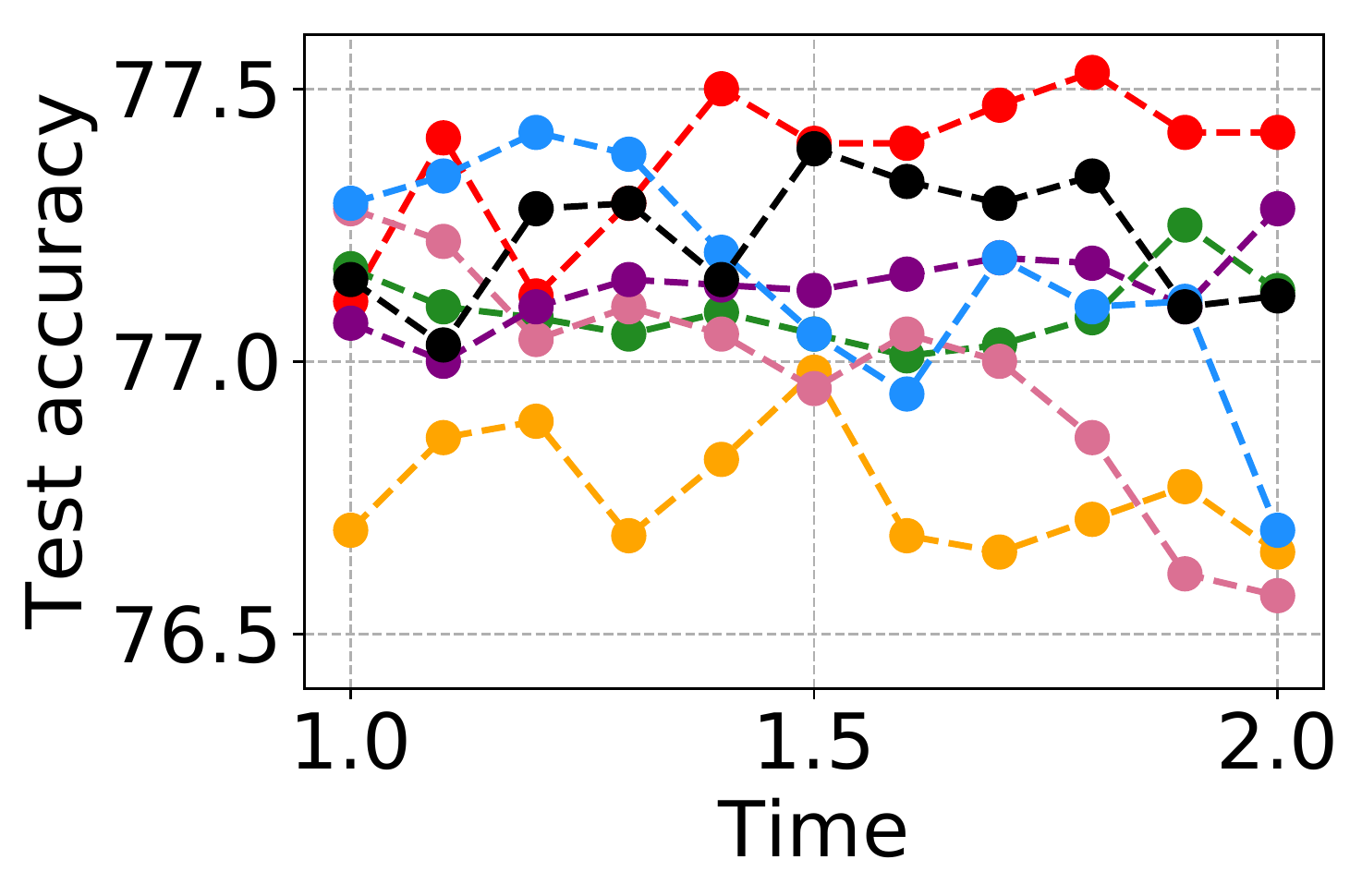}}
    \subfigure[Pubmed]{\includegraphics[width=0.32\textwidth]{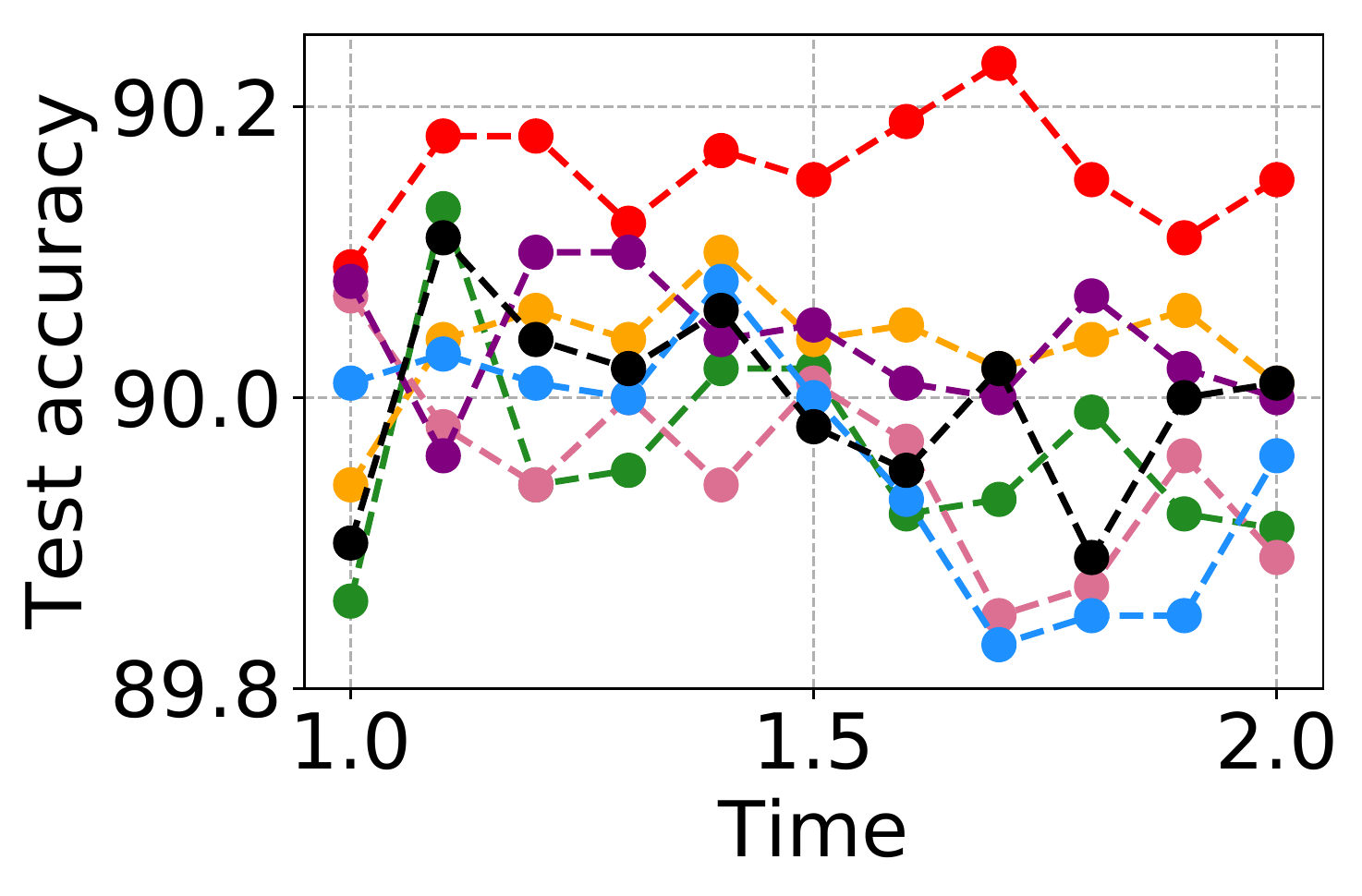}}
    \caption{Sensitivity to $T$}
    \label{fig:sens_T_appendix}
\end{figure}

\begin{figure}[ht!]
    \centering
    \includegraphics[width=\textwidth]{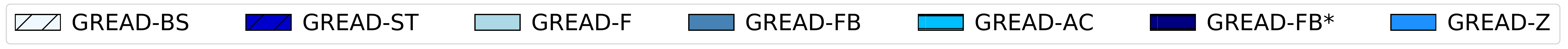}\vspace{0.2in}
    \subfigure[Texas]{\includegraphics[width=0.32\textwidth]{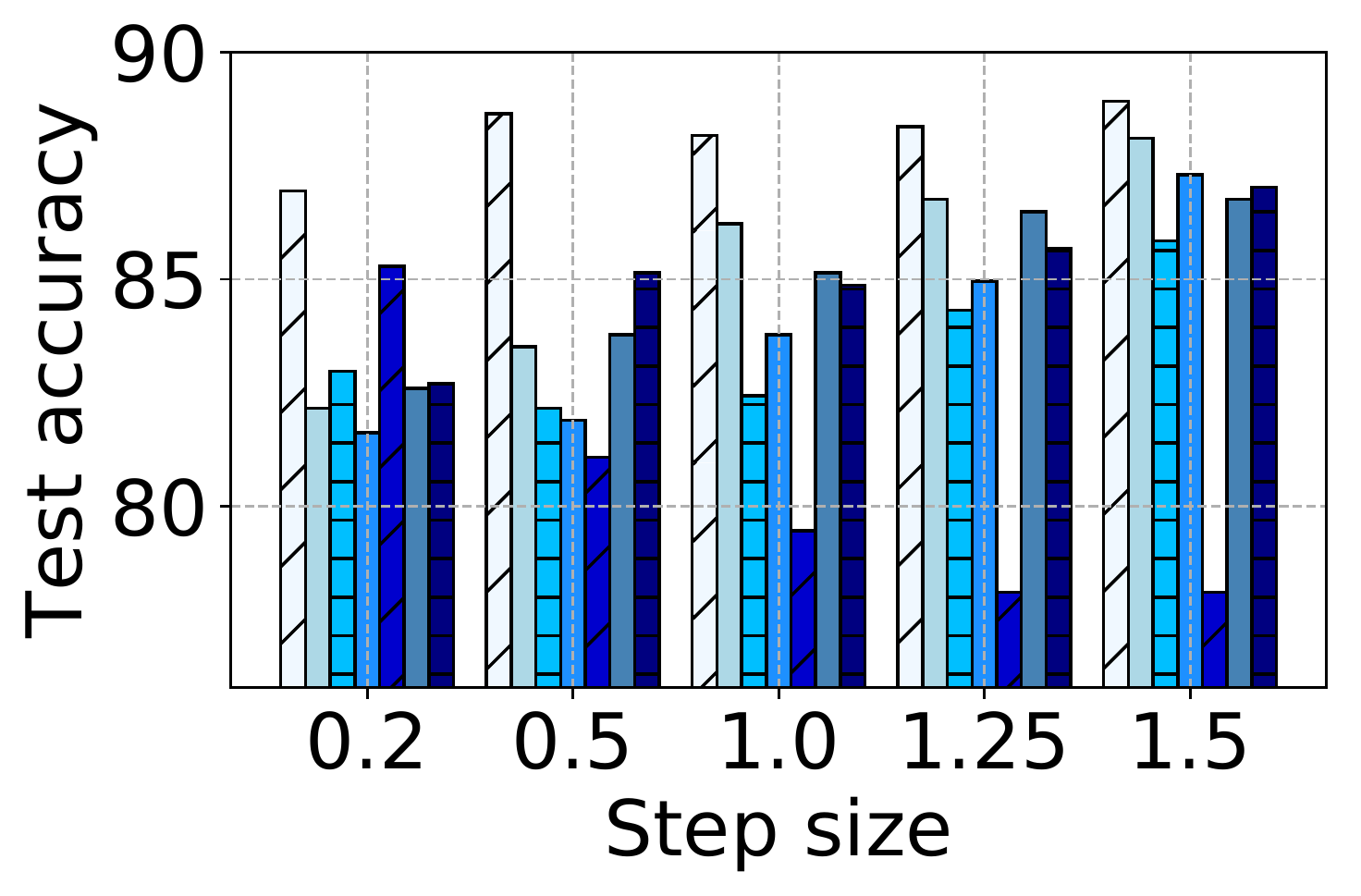}}
    \subfigure[Wisconsin]{\includegraphics[width=0.32\textwidth]{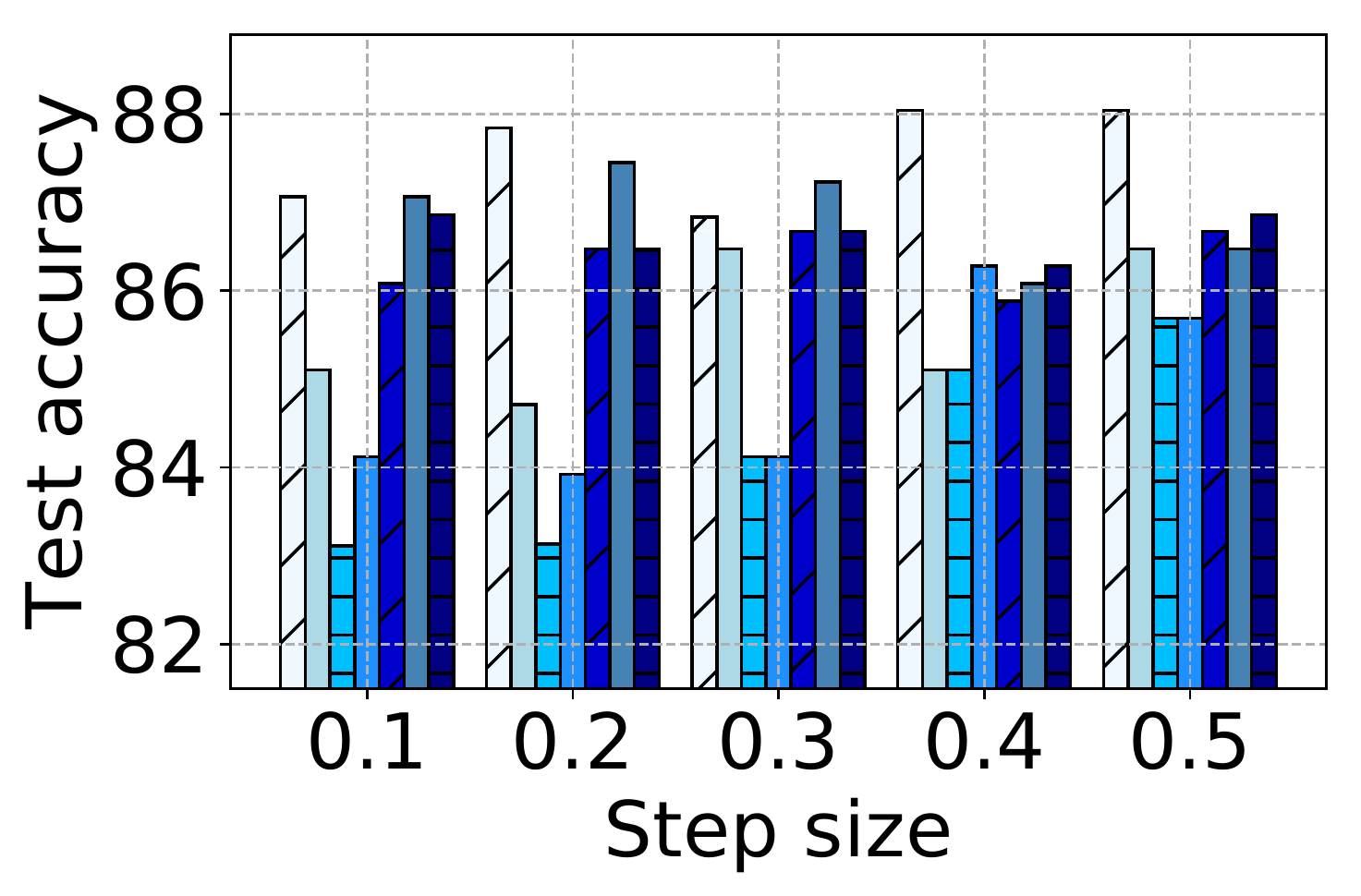}}
    \subfigure[Cornell]{\includegraphics[width=0.32\textwidth]{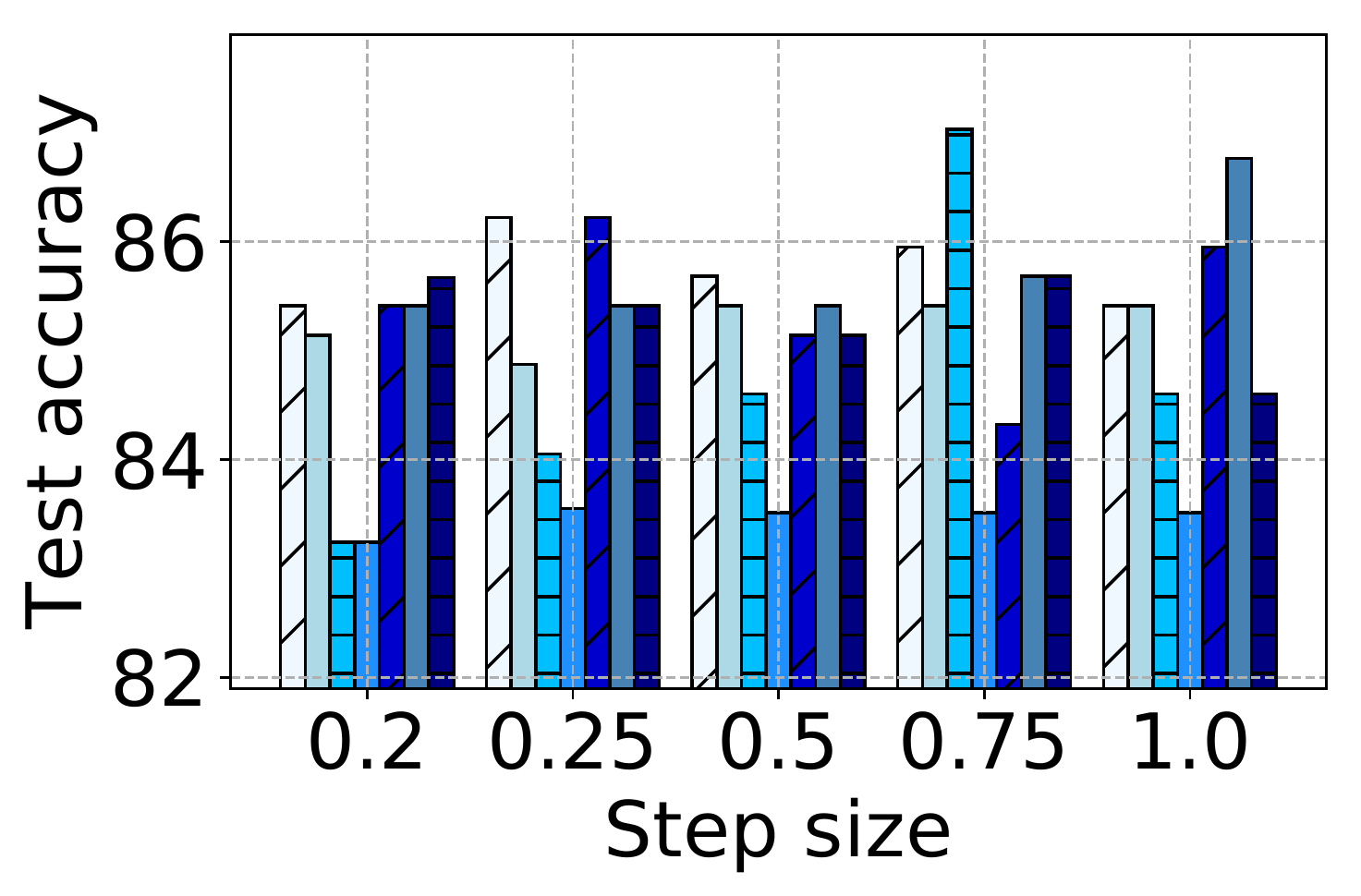}}
    \subfigure[Film]{\includegraphics[width=0.32\textwidth]{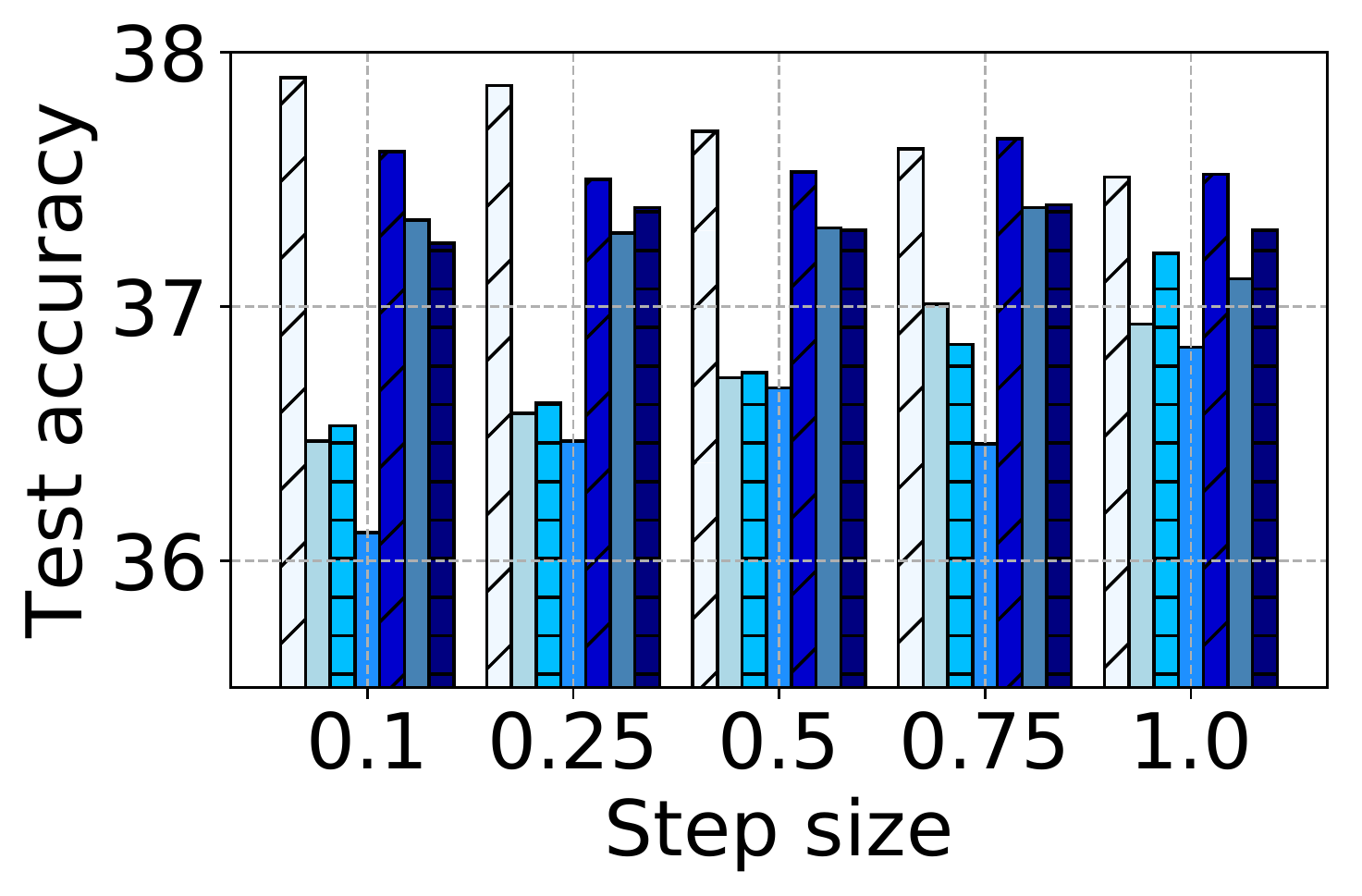}}
    \subfigure[Squirrel]{\includegraphics[width=0.32\textwidth]{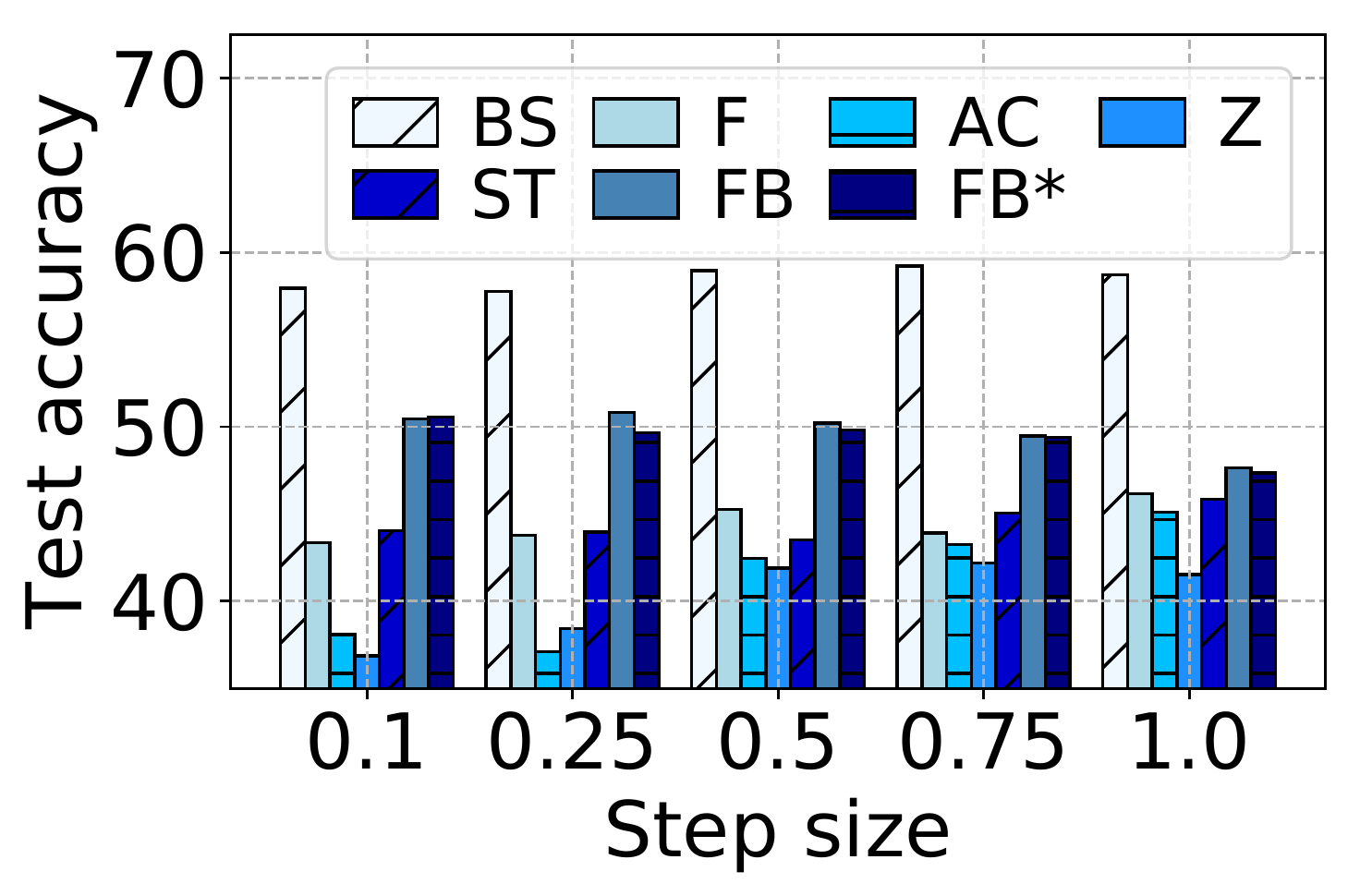}}
    \subfigure[Chameleon]{\includegraphics[width=0.32\textwidth]{img/ablation-cm/chameleon_step.pdf}}
    \subfigure[Cora]{\includegraphics[width=0.32\textwidth]{img/ablation-cm/cora_step.pdf}}
    \subfigure[Citeseer]{\includegraphics[width=0.32\textwidth]{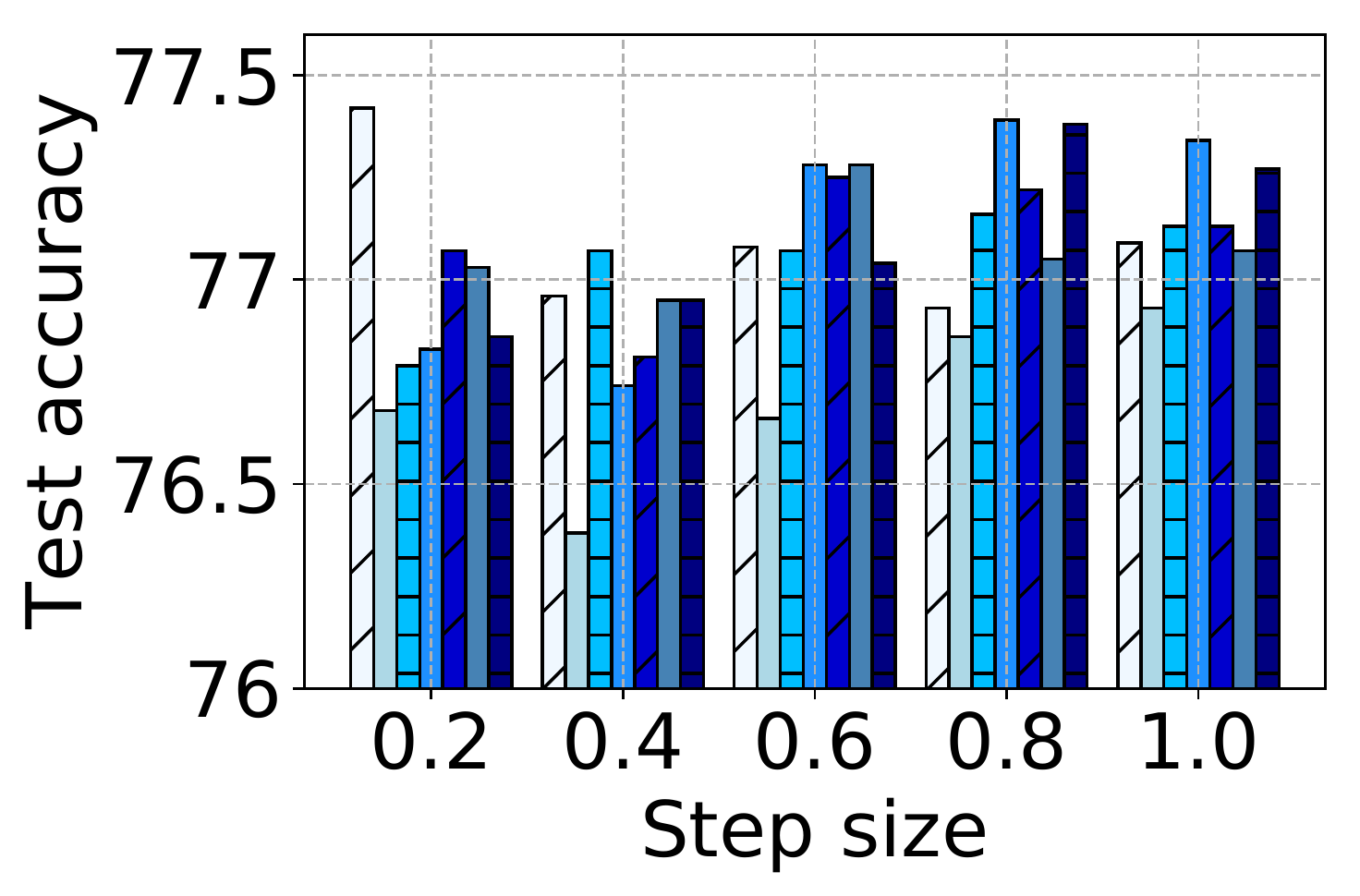}}
    \subfigure[Pubmed]{\includegraphics[width=0.32\textwidth]{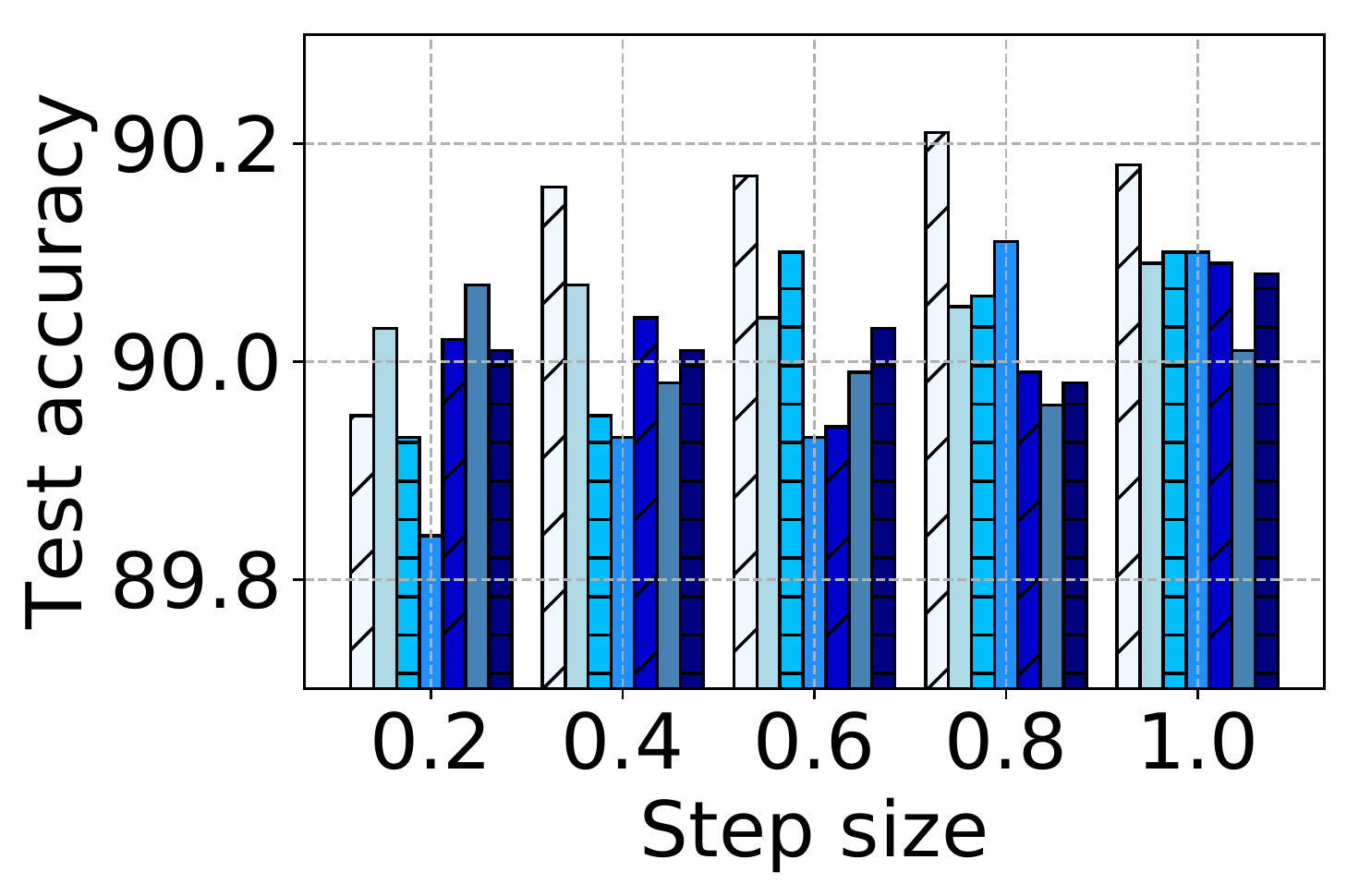}}
    \caption{Sensitivity to step size}
    \label{fig:sens_step_appendix}
\end{figure}

In Figs.~\ref{fig:sens_T_appendix} and~\ref{fig:sens_step_appendix}, we show the findings of our sensitivity studies in the remaining datasets that are not disclosed in our main manuscript. GREAD-BS maintains performance even when $T$ is increased, but GREAD-Z tends to show low performance in Texas, Cornell, and Film.

\clearpage

\subsection{Training Time}\label{a:time}
\begin{figure}[ht!]
    \centering
    \subfigure[Euler]{\includegraphics[height=0.15\textheight]{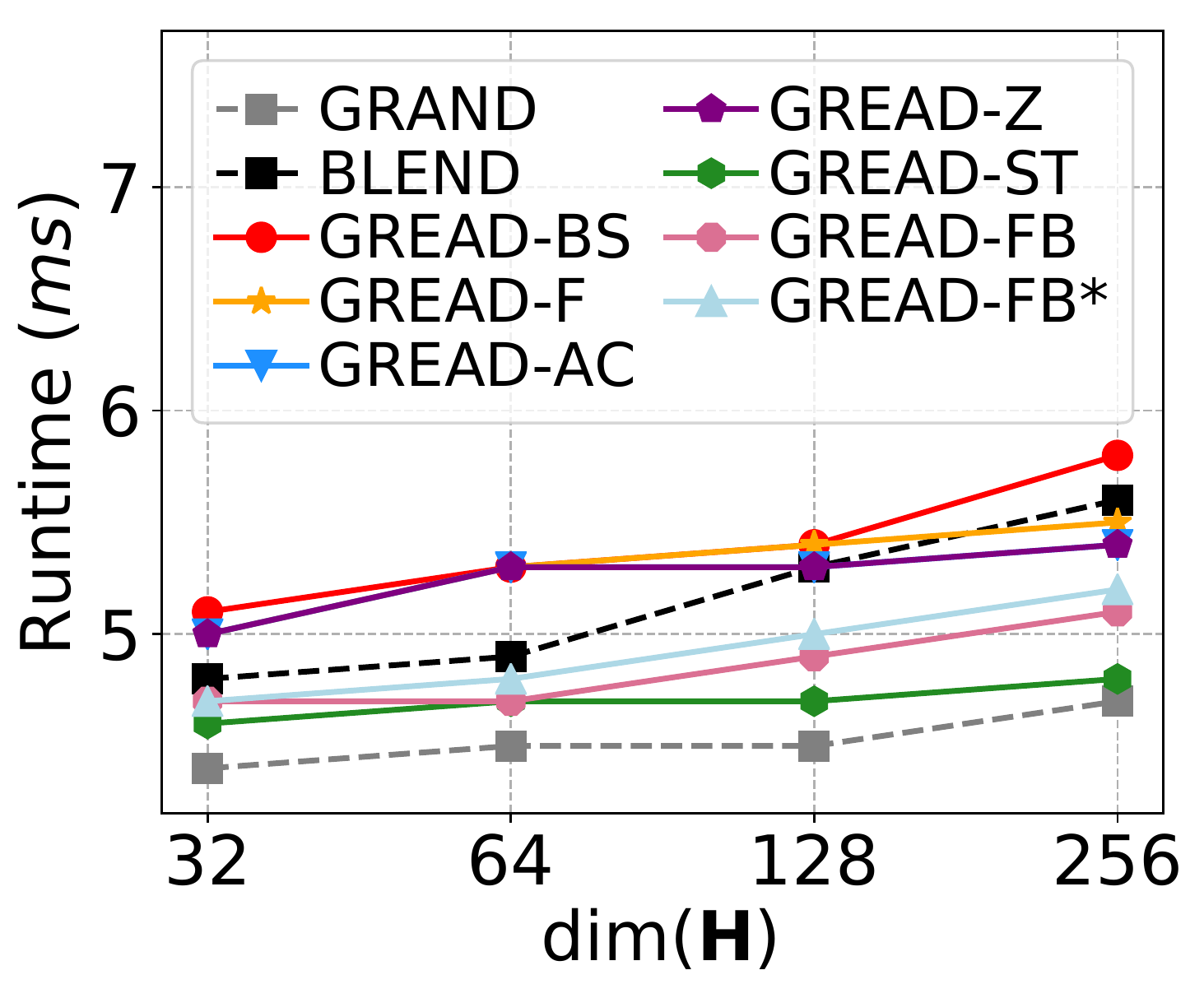}}
    \subfigure[RK4]{\includegraphics[height=0.15\textheight]{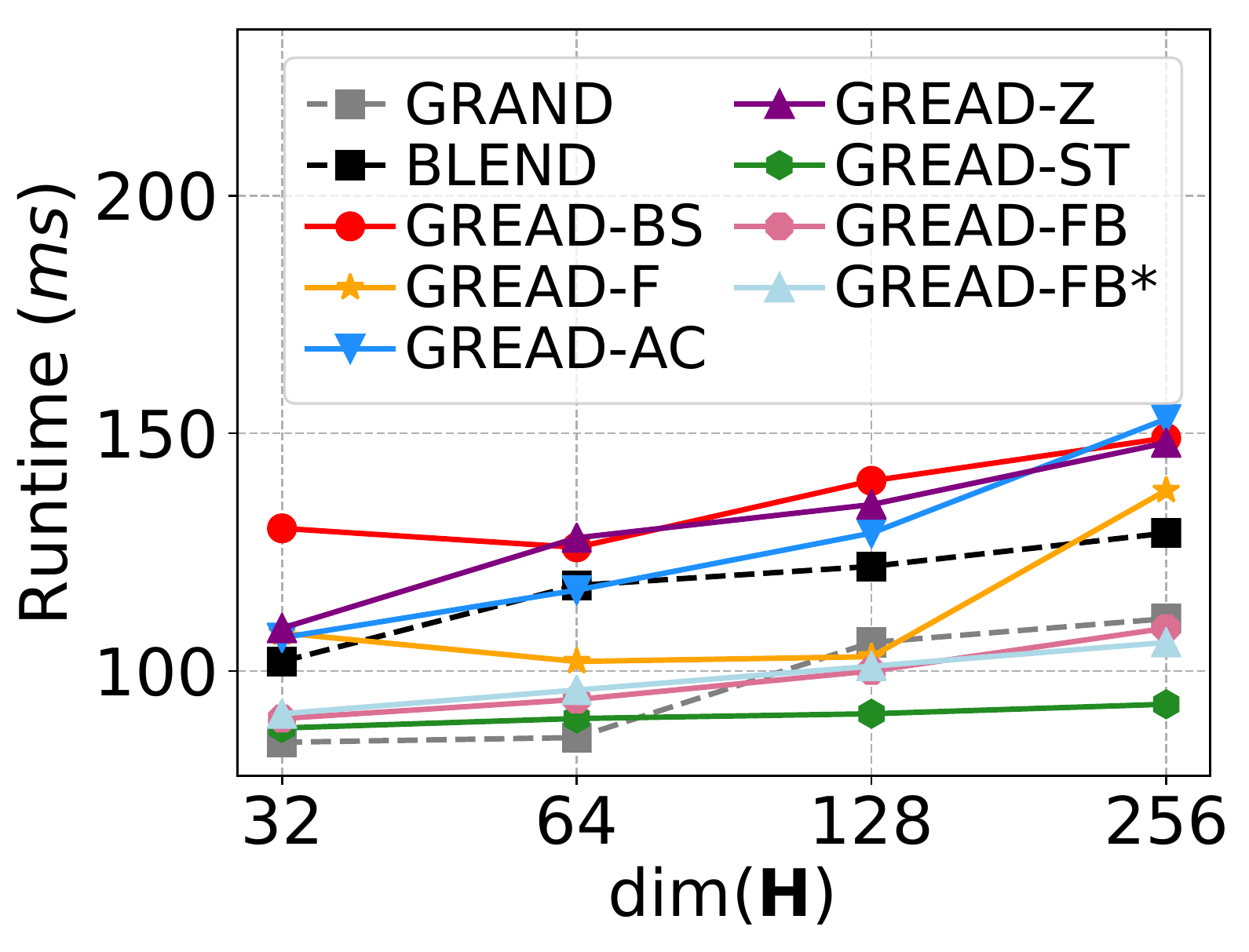}}
    \caption{Average running time per epoch (ms) on Cora dataset when $T=3$, $\tau=1.0$, SC, and OA.}
    \label{fig:runtime}
\end{figure}

We present the training time of GREAD and some selected baselines in Fig.~\ref{fig:runtime}. In general, our method's training time is little larger than those of the existing baselines because GREAD has an additional operation in its reaction term. 

\subsection{Visualizations}\label{a:visualization}
\begin{figure}[ht!]
    \centering
    \subfigure[Texas]{\includegraphics[width=0.3\textwidth]{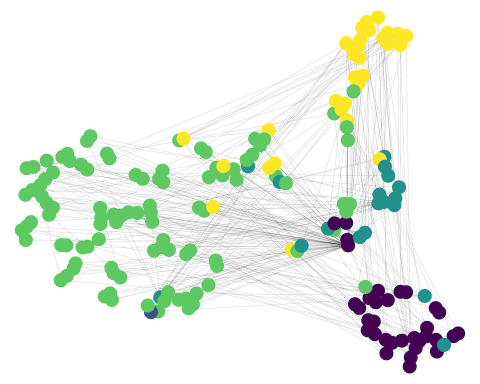}}
    \subfigure[Wisconsin]{\includegraphics[width=0.3\textwidth]{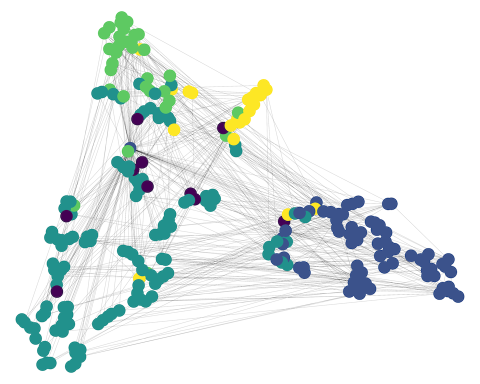}}
    \subfigure[Cornell]{\includegraphics[width=0.3\textwidth]{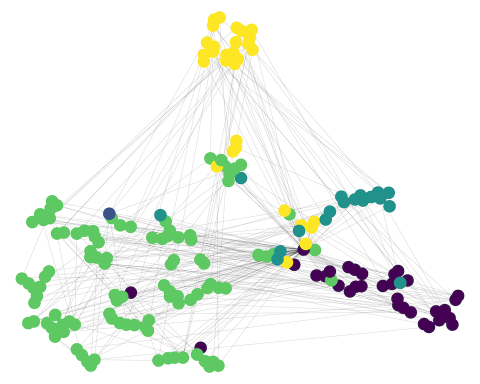}}\\
    \subfigure[Film]{\includegraphics[width=0.3\textwidth]{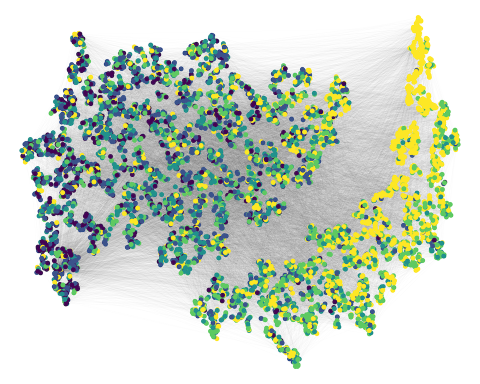}}
    \subfigure[Squirrel]{\includegraphics[width=0.3\textwidth]{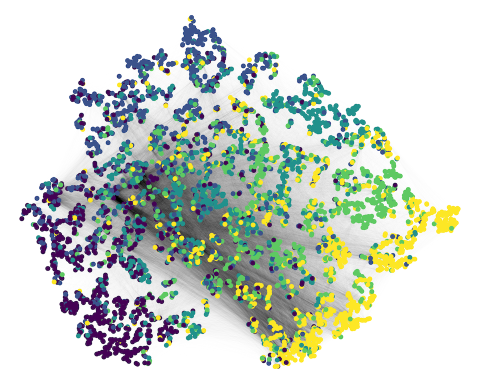}}
    \subfigure[Chameleon]{\includegraphics[width=0.3\textwidth]{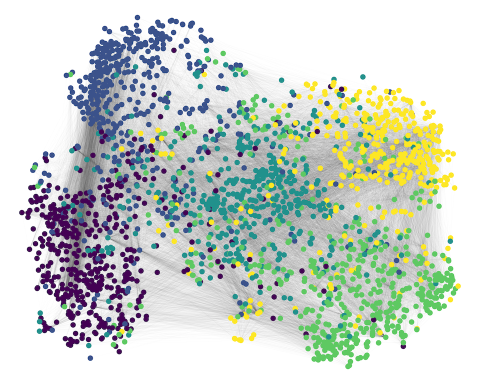}}\\
    \subfigure[Cora]{\includegraphics[width=0.3\textwidth]{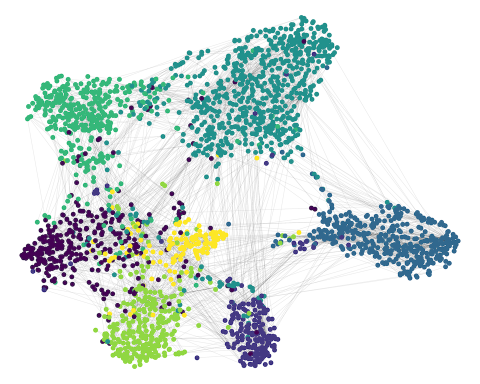}}
    \subfigure[Citeseer]{\includegraphics[width=0.3\textwidth]{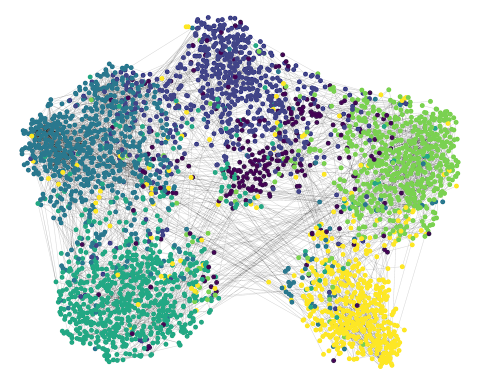}}
    \subfigure[Pubmed]{\includegraphics[width=0.3\textwidth]{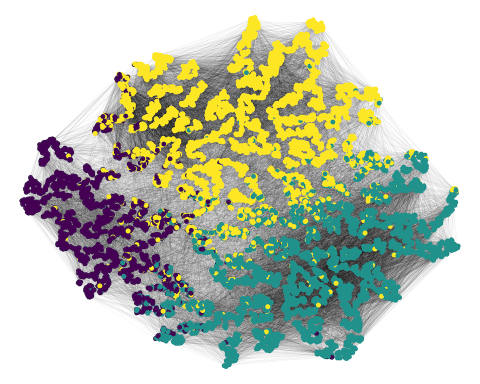}}
    \caption{Visualization of networks}
    \label{fig:networks}
 \end{figure}

In order to show the effectiveness of our proposed model more intuitively, we further conduct visualization tasks for all datasets. We extract the output vector in the final layer of GREAD and visualize those vectors using t-SNE. Fig.~\ref{fig:networks} shows the visualization results on each dataset. Different colors mean different ground-truth classes.

\clearpage

\section{Additional Experimental Results on Synthetic Datasets}
\subsection{Ablation Studies on $\beta$}
\begin{figure}[ht!]
    \centering
    \subfigure[GREAD-BS]{\includegraphics[width=0.24\textwidth]{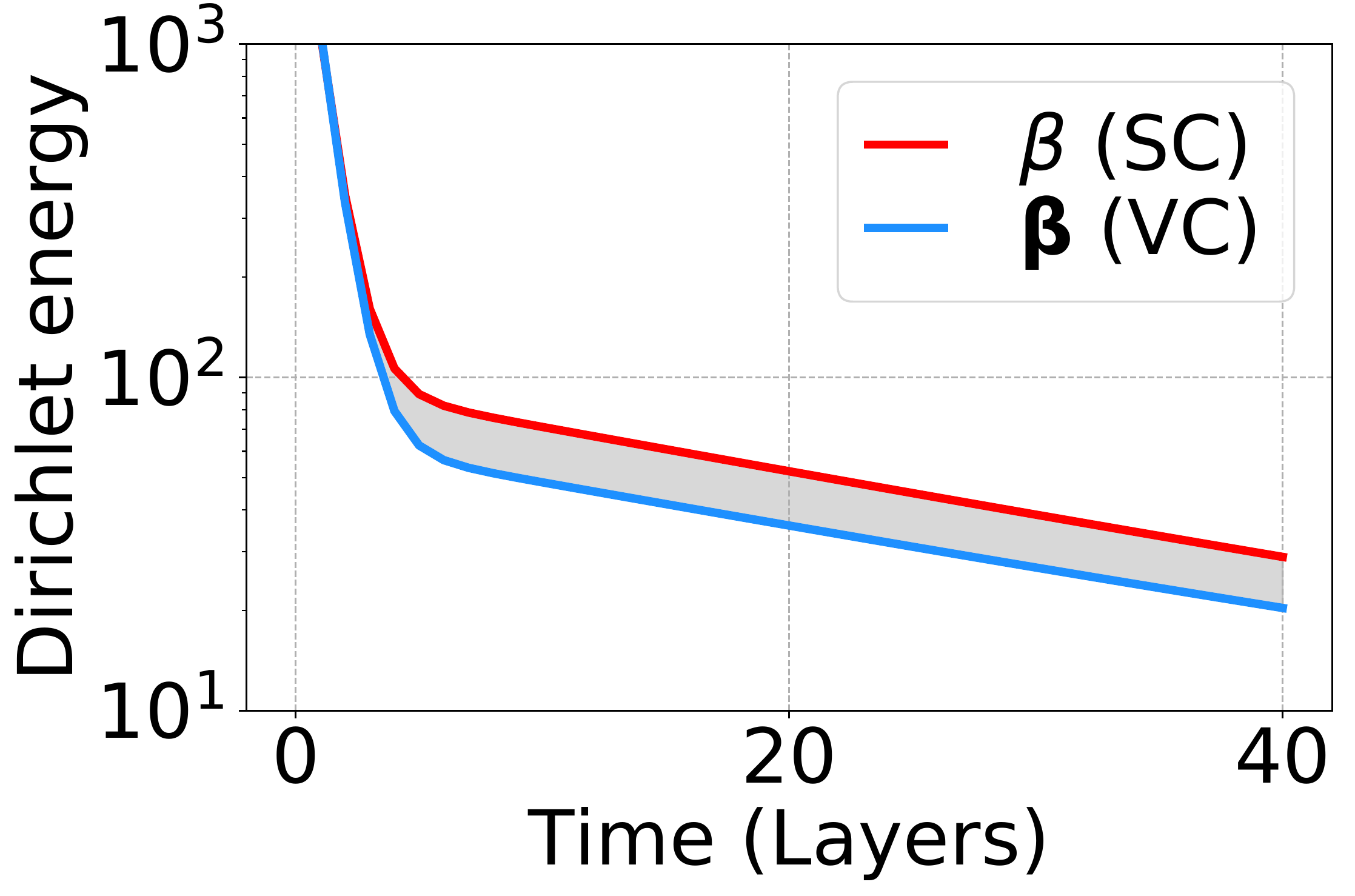}}
    \subfigure[GREAD-F]{\includegraphics[width=0.24\textwidth]{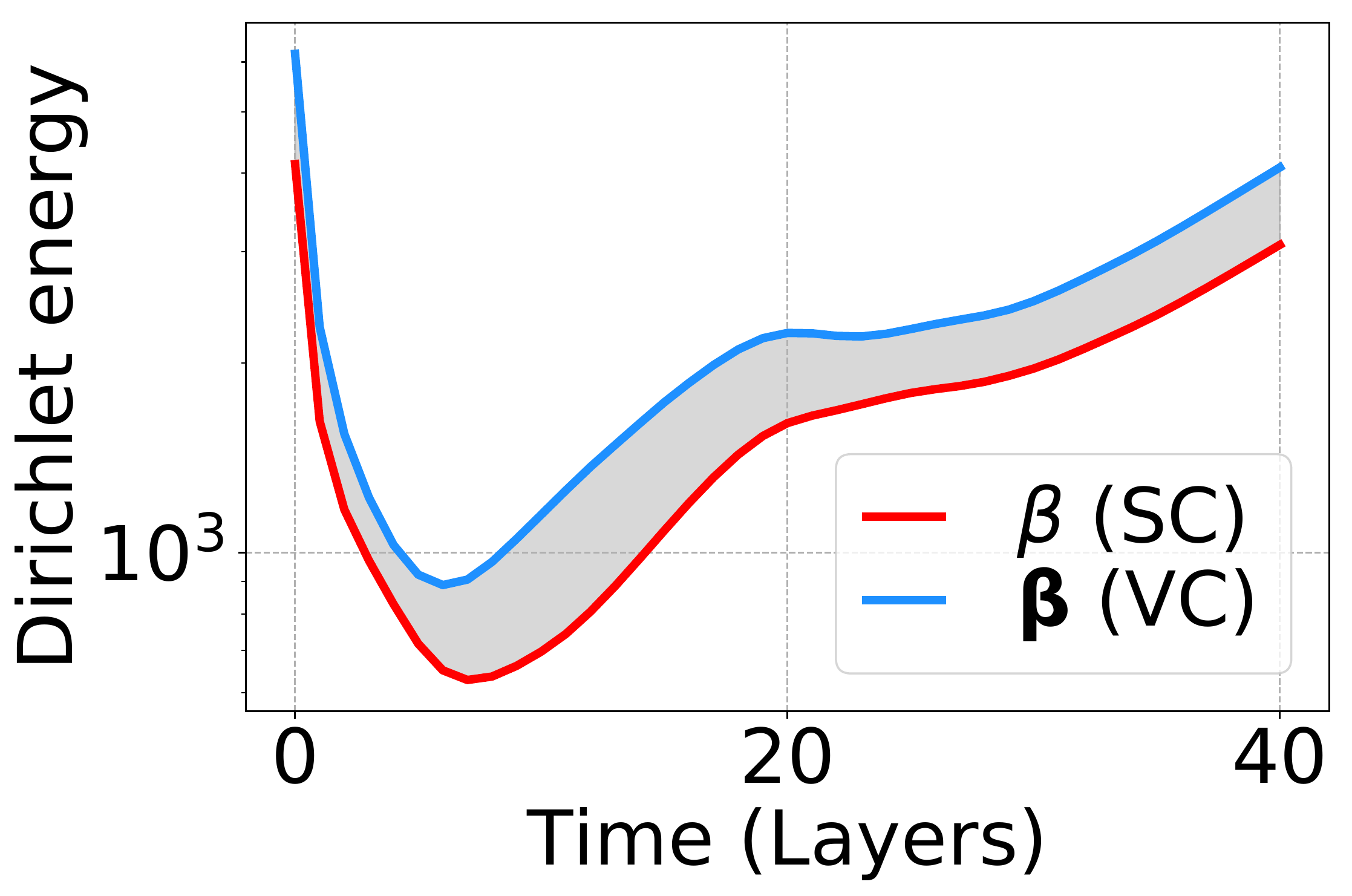}}
    \subfigure[GREAD-AC]{\includegraphics[width=0.24\textwidth]{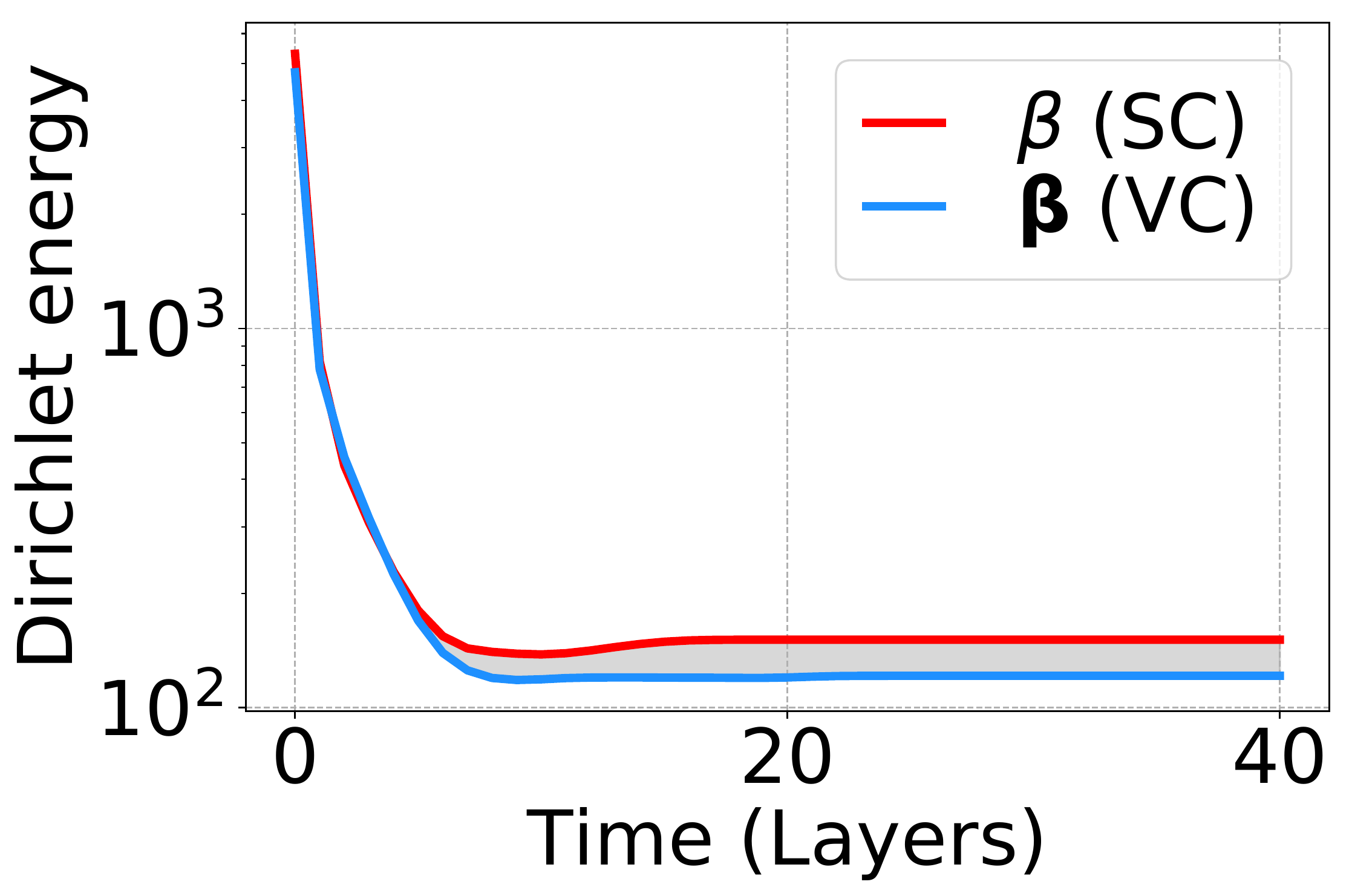}}
    \subfigure[GREAD-Z]{\includegraphics[width=0.24\textwidth]{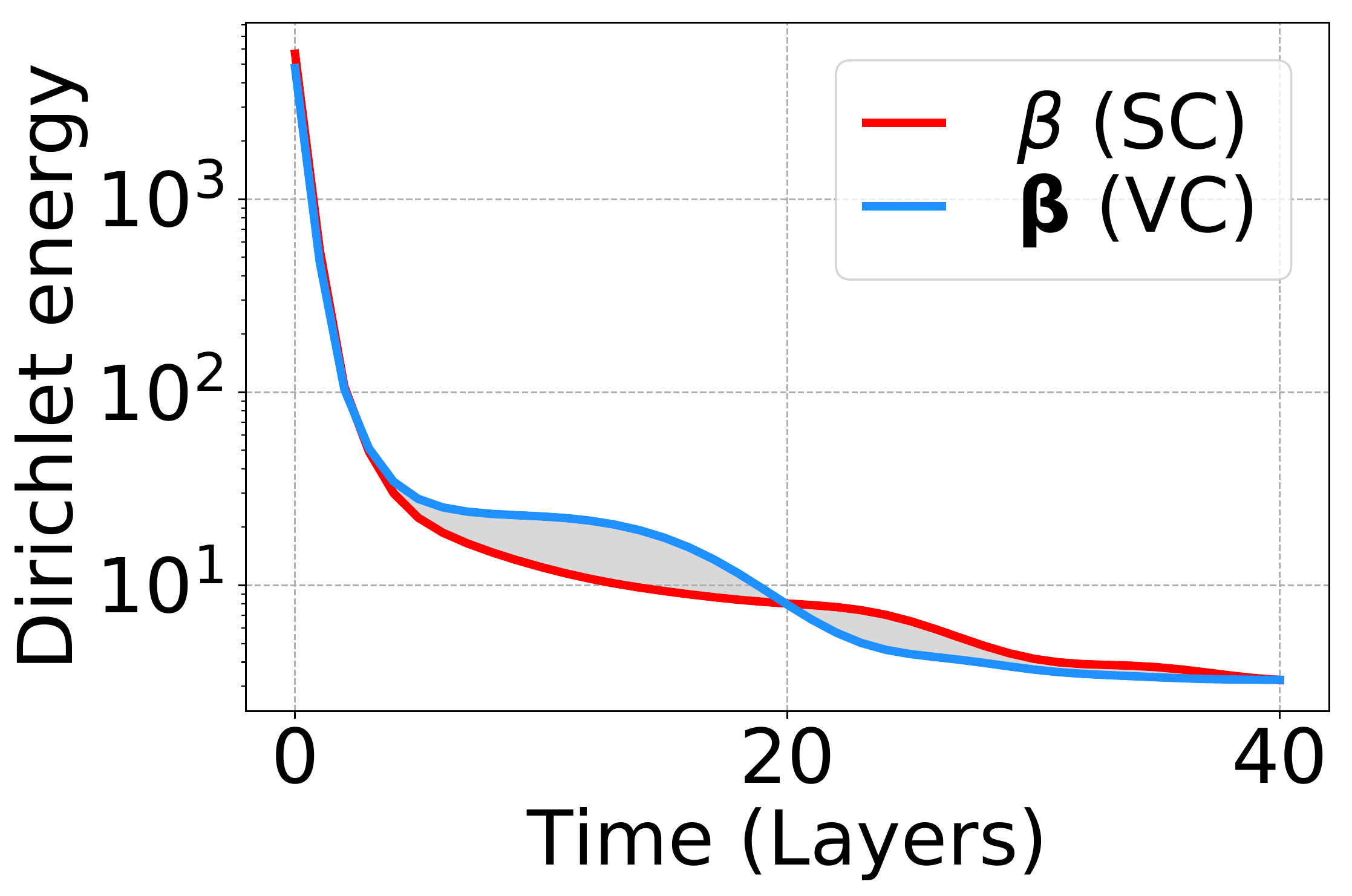}}
    \subfigure[GREAD-ST]{\includegraphics[width=0.24\textwidth]{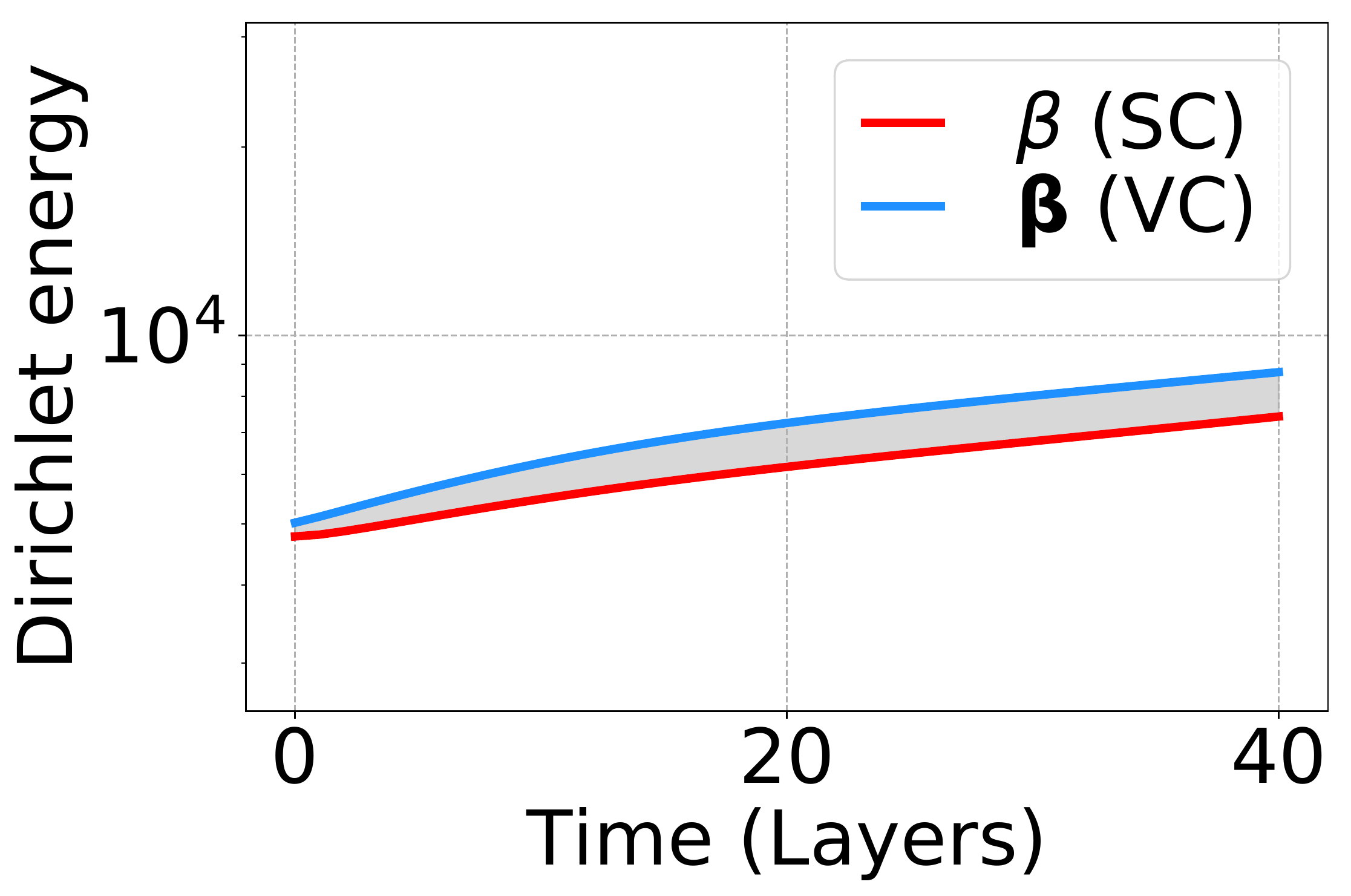}}
    \subfigure[GREAD-FB]{\includegraphics[width=0.24\textwidth]{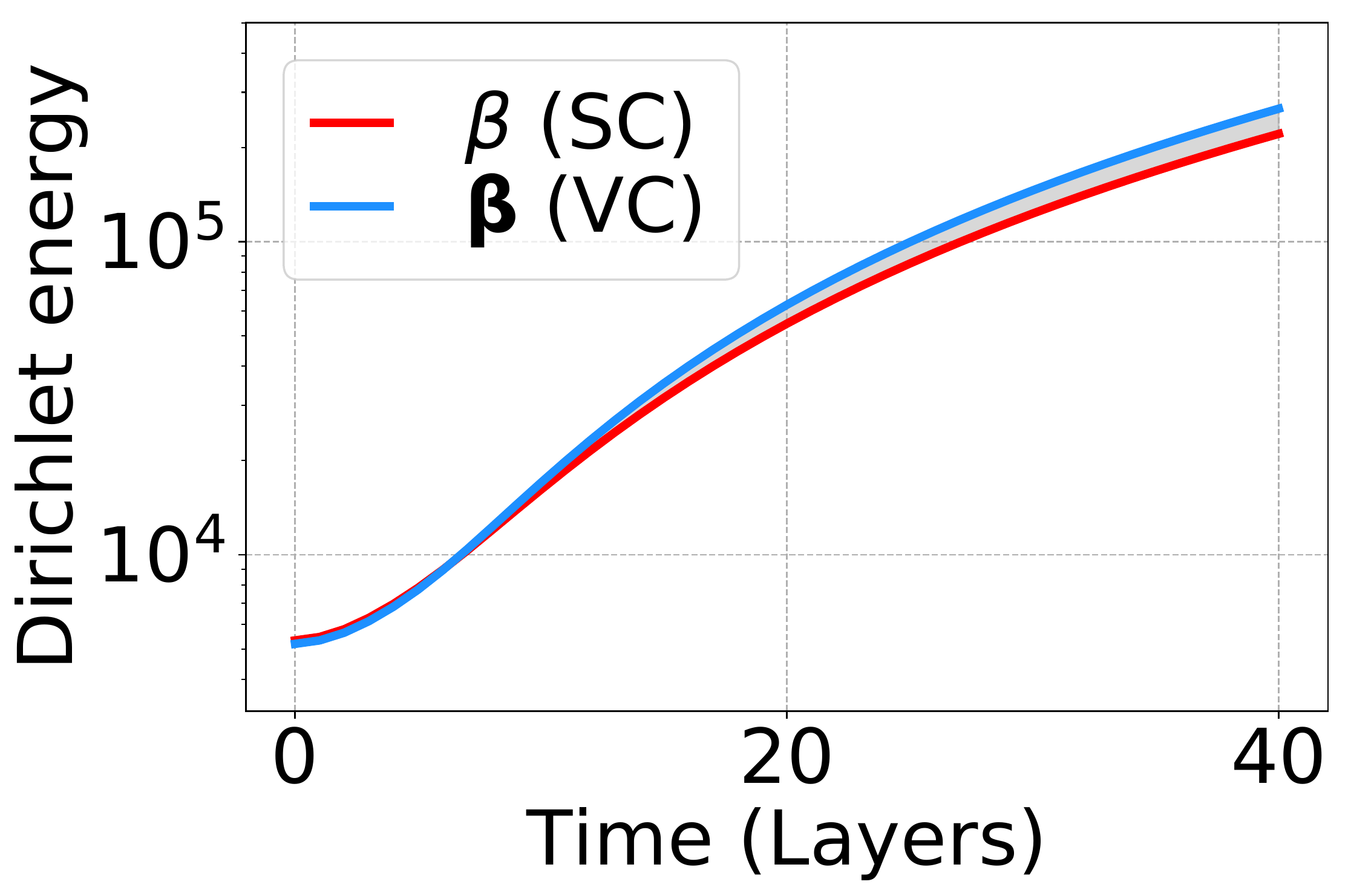}}
    \subfigure[GREAD-FB*]{\includegraphics[width=0.24\textwidth]{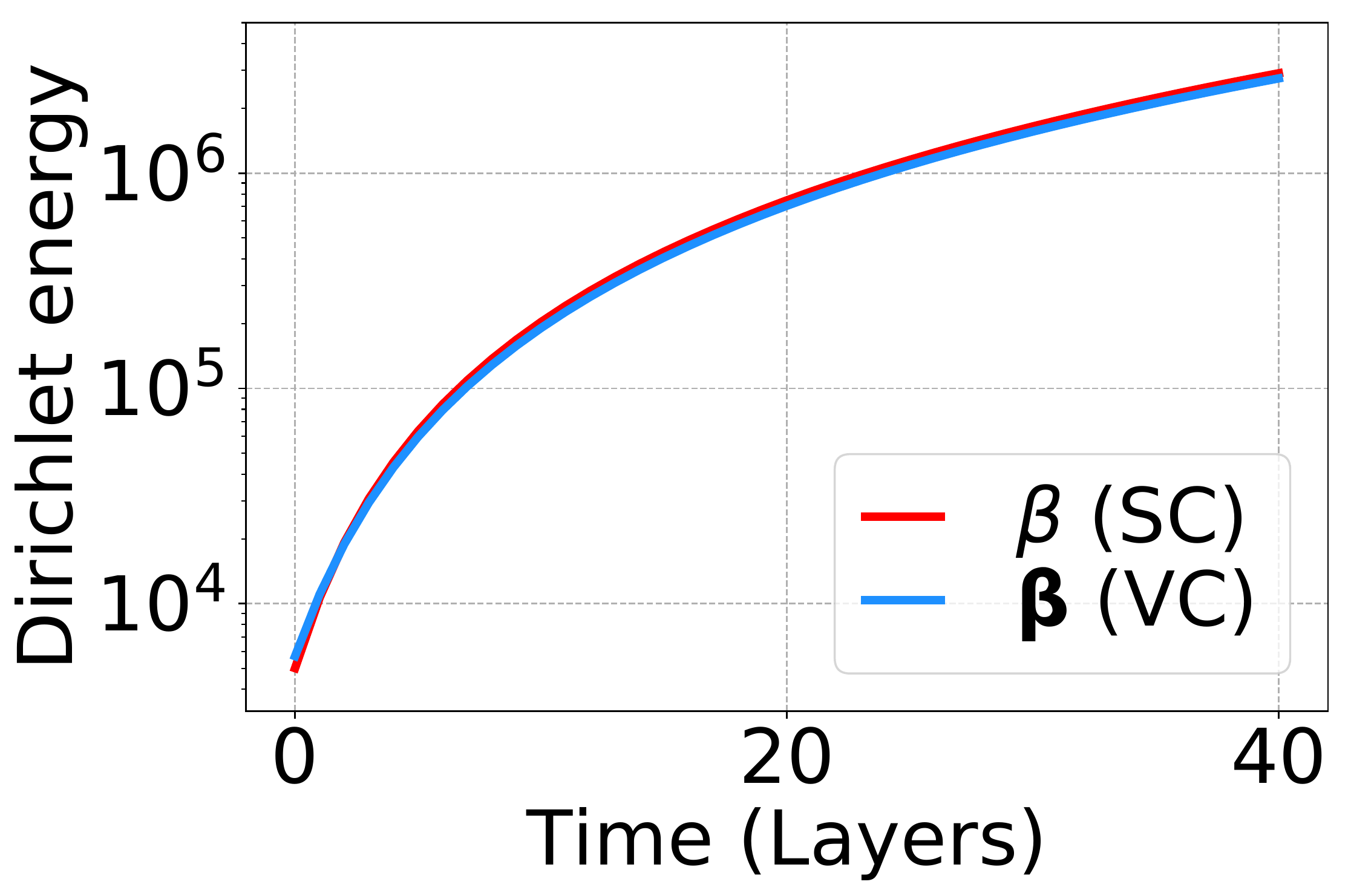}}
    \caption{Evolution of the Dirichlet energy on the synthetic random graph. The Y-axis is the logarithmic Dirichlet energy in each layer’s output given a GNN of 40 layers. The gray area is the Dirichlet energy difference between SC and VC.}
    \label{fig:energy_beta}
\end{figure}

We perform the ablation study on $\beta$ from the perspective of the Dirichlet energy. $\beta$ can be either a scalar parameter (SC) or a learnable vector parameter (VC). In Fig.~\ref{fig:energy_beta}, we show the evolution of the Dirichlet energy on the synthetic random graph created from cSBM~\cite{Deshpande2018cSBM}, and compare SC and VC for our proposed method. In the case of GREAD-F, GREAD-AC, GREAD-ST, and GREAD-FB, VC conserves more energy than SC, so the reaction term multiplied with $\beta$ successfully mitigates the oversmoothing problem.

\clearpage

\section{Comparison with GRAND++ and GREAD-ST}
The structures of GREAD-ST and GRAND++ are similar to each other since they commonly use a source term to alleviate the oversmoothing problem. However, there exists a key difference. First, GRAND++ is written as follows:
\begin{align}
    f(\mathbf{H}(t))_{\text{GRAND++}} := -\mathbf{L}\mathbf{H}(t) + \mathbf{C}(0),
\end{align}
where $\mathbf{C}(0)$ is a subset of $\mathbf{H}(0)$, which consists of only “trustworthy” nodes.

On the other hand, the formula of GREAD-ST is written as follows:
\begin{align}
    f(\mathbf{H}(t))_{\text{GREAD-ST}} := \alpha (-\mathbf{L}\mathbf{H}(t)) + \beta(\mathbf{C}(0)).
\end{align}

In our GREAD-ST, the full source term $\mathbf{H}(0)$ is added to the reaction term. After that, $\beta$, a learnable parameter, determines how much the source term is added. In other words, GREAD-ST learns how to utilize $\mathbf{H}(0)$ in the reaction term.

We compare GRAND++~\cite{thorpe2022grands} and GREAD-ST through experiments using the benchmark dataset in Table~\ref{tab:gread-st}. GREAD-ST shows superiority in all datasets. In Table~\ref{tab:gread-st-depth}, we also show the performance by varying the number of layers for Cora. Our method is more robust to the oversmoothing problem than GRAND++.

\begin{table}[h!]
    \setlength{\tabcolsep}{2pt}
    \centering
    \caption{Comparison with GRAND++ and GREAD-ST}
    \begin{tabular}{c ccccccccc}\toprule
        Model    & Texas & Wisconsin & Cornell & Film & Squirrel & Chameleon & Cora & Citeseer & Pubmed\\\midrule
        GRAND++  & 77.57\std{±5.00} & 82.75\std{±4.19} & 81.89\std{±5.28} & 33.63\std{±0.48} & 40.06\std{±1.70} & 56.20\std{±2.15} & 88.15\std{±1.22} & 76.57\std{±1.46} & 88.50\std{±0.35}\\
        \textbf{GREAD-ST} & \textbf{81.08\std{±5.67}} & \textbf{86.67\std{±3.01}} & \textbf{86.22\std{±5.98}} & \textbf{37.66\std{±0.90}} & \textbf{45.83\std{±1.40}} & \textbf{63.03\std{±1.32}} & \textbf{88.47\std{±1.19}} & \textbf{77.25\std{±1.47}} & \textbf{90.13\std{±0.36}}\\\bottomrule
    \end{tabular}
    \label{tab:gread-st}
\end{table}

\begin{table}[h]
    \centering
    \caption{Classification accuracy of GRAND++ and GREAD-ST with different depths on Cora dataset}
    \begin{tabular}{cc cccccc}\toprule
        \multirow{2}{*}{Dataset} & \multirow{2}{*}{Model} & \multicolumn{6}{c}{Layer}\\\cmidrule{3-8} 
                &       & 2 & 4 & 8 & 16 & 32 & 64\\ \midrule
        \multirow{2}{*}{Cora}    
        & GRAND++ & 87.38\std{±2.01} & 88.15\std{±1.22} & 87.89\std{±1.13} & 87.73\std{±0.96} & 87.52\std{±1.28} & 87.73\std{±1.30}\\ 
        & \textbf{GREAD-ST} & \textbf{88.17\std{±0.38}} & \textbf{88.31\std{±0.39}} & \textbf{88.39\std{±0.83}} & \textbf{88.12\std{±0.53}} & \textbf{88.47\std{±1.19}} & \textbf{88.37\std{±0.98}}\\ \bottomrule
    \end{tabular}
    \label{tab:gread-st-depth}
\end{table}

\section{Well-posedness of GREAD}
The well-posedness\footnote{A well-posed problem means i) its solution uniquely exists, and ii) its solution continuously changes as input data changes.} of NODEs was already proved in \citet[Theorem 1.3]{lyons2004differential} under the mild condition of the Lipschitz continuity. Almost all activations, such as ReLU, Leaky ReLU, SoftPlus, Tanh, Sigmoid, ArcTan, and Softsign, have a Lipschitz constant of 1. Other common neural network layers, such as dropout, batch normalization, and other pooling methods, have explicit Lipschitz constant values. Therefore, the Lipschitz continuity of $\mathbf{f}$ can be fulfilled in some cases of GREAD, making the initial value problem in Eq.~\eqref{eq:rd} a well-posed problem. However, some other functions are locally Lipschitz continuous. For instance, GREAD with the soft adjacency matrix does not satisfy the globally Lipschitz continuous property. Nevertheless, our experimental results show that GREAD can be properly trained and outperforms many baselines.

\clearpage

\section{Statistical Testing on Cora Dataset}
For proper statistical testing, we experiment more with 10 different seeds per split and perform statistical tests using a total of 100 experimental results. The unpaired t-test of GRAND and GREAD-BS is shown in Table~\ref{tab:ttest}. As shown, GREAD-BS has a clear improvement in all datasets compared to GRAND.

\begin{table}[!ht]
    \centering
    \setlength{\tabcolsep}{2pt}
    \caption{Significance test between GRAND and GRAND-BS utilizing unpaired t-test}
    \begin{tabular}{cccccccccc}
    \toprule
        ~ & Texas & Wisconsin & Cornell & Film & Squirrel & Chameleon & Cora & Citeseer & Pubmed \\ \midrule
        GRAND & 76.10\std{±6.03} & 79.19\std{±5.26} & 82.17\std{±5.99} & 33.43\std{±1.29} & 38.09\std{±1.38} & 53.86\std{±2.04} & 87.12\std{±1.74} & 76.11\std{±1.24} & 88.82\std{±0.50} \\ 
        \textbf{GREAD-BS} & \textbf{88.71\std{±3.24}} & \textbf{89.10\std{±2.90}} & \textbf{86.44\std{±7.03}} & \textbf{39.90\std{±1.02}} & \textbf{58.89\std{±1.11}} & \textbf{70.04\std{±0.93}} & \textbf{88.43\std{±0.59}} & \textbf{77.48\std{±1.15}} & \textbf{90.03\std{±0.49}} \\ \midrule
        t-statistic & 18.42 & 16.50 & 4.62 & 39.34 & 117.45 & 72.17 & 7.13 & 8.10 & 17.28 \\ 
        p-value & $<$0.05 & $<$0.05 & $<$0.05 & $<$0.05 & $<$0.05 & $<$0.05 & $<$0.05 & $<$0.05 & $<$0.05 \\ \bottomrule
    \end{tabular}
    \label{tab:ttest}
\end{table}

\end{document}